%% file: main.tex
\documentclass[sglanonrev]{informs4-myManuscript}
\usepackage[T1]{fontenc}
\renewcommand{\theARTICLETOP}{}
\RRHSecondLine{}
\LRHSecondLine{}
\RequirePackage{tgtermes}
\RequirePackage{newtxtext}
\RequirePackage{newtxmath}
\RequirePackage{bm}
\RequirePackage{endnotes}

\OneAndAHalfSpacedXI 

\usepackage{graphicx}
\usepackage{graphbox} 
\usepackage{caption}
\DeclareCaptionLabelSeparator{sixteenpt}{\kern16pt}
\usepackage{subcaption} 
\usepackage{multirow}
\usepackage{listings}
\usepackage{hhline}
\usepackage[dvipsnames,svgnames,table]{xcolor}
\usepackage[most]{tcolorbox}
\tcbset{
    base/.style={
        boxrule=2pt,
        arc=0mm,
        breakable,
        colback=white!97!black,
        colframe=black!80,
        fonttitle=\small\bfseries,
        left=2mm,
        right=2mm,
        top=2mm,
        bottom=2mm
    }
}
\usepackage{lscape}
\usepackage{algorithm}
\usepackage{algorithmic}
\usepackage{float}

\usepackage{booktabs}
\usepackage{array}
\usepackage{threeparttable}

\usepackage{bbm}
\usepackage{anyfontsize}

\usepackage{appendix}

\usepackage{natbib}
 \bibpunct[, ]{(}{)}{;}{a}{,}{,}%
 \def\bibfont{\small}%
\usepackage{url}
\usepackage[letterpaper=true,colorlinks=true,pdfpagemode=none,citecolor=OliveGreen,linkcolor=BrickRed,urlcolor=BrickRed,pdfstartview=FitH]{hyperref}

\long\def\fignote#1{\par{\FigureNoteStyle
    \HD{10}{0}{\it\FigureNoteName}\enskip
    #1\HD{0}{3}\vspace*{-3pt}\endgraf}}%

\long\def\tabnote#1{\par\TableNoteStyle\parindent1em\raggedright\ignorespaces
\HD{11}{0}#1\endgraf}

\EquationsNumberedThrough    
\TheoremsNumberedThrough     
\ECRepeatTheorems  %

\begin{document}
\RUNAUTHOR{Chen et~al.} %
\RUNTITLE{Causal Prompt Optimization}
\TITLE{Optimizing Prompts for Large Language Models: A Causal Approach}

\ARTICLEAUTHORS{%
\AUTHOR{Wei Chen}
\AFF{School of Business, University of Connecticut, Stamford, CT, US, \EMAIL{weichen@uconn.edu}}
\AUTHOR{Yanbin Fang}
\AFF{Antai College of Economics and Management, Shanghai Jiao Tong University, Shanghai, China, \EMAIL{yibo\_fang@sjtu.edu.cn}}
\AUTHOR{Shuran Fu}
\AFF{School of Business, University of Connecticut, Storrs, CT, US, \EMAIL{mlp24004@uconn.edu}}
\AUTHOR{Fasheng Xu}
\AFF{School of Business, University of Connecticut, Stamford, CT, US, \EMAIL{fasheng.xu@uconn.edu}}
\AUTHOR{Xuan Wei}
\AFF{Antai College of Economics and Management, Shanghai Jiao Tong University, Shanghai, China, \EMAIL{weix@sjtu.edu.cn}} 
}

\ABSTRACT{
Large Language Models (LLMs) are increasingly embedded in enterprise workflows, yet their performance remains highly sensitive to prompt design. Automatic Prompt Optimization (APO) seeks to mitigate this instability, but existing approaches face two persistent challenges. First, commonly used prompt strategies rely on static instructions that perform well on average but fail to adapt to heterogeneous queries. Second, more dynamic approaches depend on offline reward models that are fundamentally correlational, confounding prompt effectiveness with query characteristics. 
We propose Causal Prompt Optimization (CPO), a framework that reframes prompt design as a problem of causal estimation. CPO operates in two stages. First, it learns an offline causal reward model by applying Double Machine Learning (DML) to semantic embeddings of prompts and queries, isolating the causal effect of prompt variations from confounding query attributes. 
Second, it utilizes this unbiased reward signal to guide a resource-efficient search for query-specific prompts without relying on costly online evaluation. We evaluate CPO across benchmarks in mathematical reasoning, visualization, and data analytics. CPO consistently outperforms human-engineered prompts and state-of-the-art automated optimizers. The gains are driven primarily by improved robustness on hard queries, where existing methods tend to deteriorate. Beyond performance, CPO fundamentally reshapes the economics of prompt optimization: by shifting evaluation from real-time model execution to an offline causal model, it enables high-precision, per-query customization at a fraction of the inference cost required by online methods. Together, these results establish causal inference as a scalable foundation for reliable and cost-efficient prompt optimization in enterprise LLM deployments.
}

\KEYWORDS{Large Language Models (LLMs), Causal Prompt Optimization (CPO), Automatic Prompt Optimization, Causal Inference, Query-Adaptive Prompting, Double Machine Learning (DML), AI System Design} 

\maketitle
\section{Introduction}
The rapid integration of Large Language Models (LLMs) to support enterprise workflows, such as financial reporting, customer support, and market analysis, has fundamentally altered the landscape of managerial decision-making~\citep[e.g.,][]{eloundou2024gpts,zhao2024revolutionizing, chen2024large, zhou2024generative, brynjolfsson2025generative}. In these high-stakes contexts, natural-language-based instructions to LLMs, i.e., the ``prompts,'' serve as the primary interface for organizations to specify objectives, impose constraints, and manage risk~\citep{brown2020language, radford2019language}. Yet LLMs routinely exhibit sensitivity to subtle variations in how prompts are phrased or structured, producing large and often unpredictable differences in output quality~\citep{li2021prefix, liu2023pre}. This brittleness forces practitioners to rely on extensive manual tuning to ensure performance. Consequently, the prevailing practice of prompt engineering remains labor-intensive, relying on ad-hoc heuristics and manual trial-and-error that are prohibitively costly to scale and difficult to standardize across heterogeneous business tasks~\citep{zhou2022large, reynolds2021prompt}.

To overcome these inefficiencies, the field has turned toward Automatic Prompt Optimization (APO), an approach that automatically refines prompts without accessing the LLM’s internal parameters\footnote{Given the prevalence of proprietary models accessed via APIs (e.g., OpenAI’s GPT, Anthropic’s Claude, or Google’s Gemini) in business settings, we focus specifically on \textit{black-box} prompt design—a setting where prompts are refined entirely in the natural language space without access to the LLM's internal gradients or weights.}~\citep{cheng2024black, pryzant2023automatic}. Most APO practices have primarily focused on task-level optimization, aiming to discover a single, \textit{static} prompt that performs well on average across a given dataset of a task category~\citep[e.g.,][]{zhou2022large, pryzant2023automatic}. However, this ``one-size-fits-all'' paradigm often fails to account for the diversity of individual queries in a task category, as a prompt suitable for one query may underperform on another. Consequently, recent advancements have shifted toward query-level optimization, where prompts are \textit{dynamically} tailored to specific input instances to maximize performance~\citep{kong2025query}. Crucially, the success of this granular optimization hinges on a reliable reward model, a mechanism to accurately estimate the quality of a prompt on a specific query without requiring actual LLM calls.

A critical limitation undermining current query-level optimization is that existing reward models rely fundamentally on \textit{correlational signals}. Whether trained on historical logs or offline datasets, these models typically infer the efficacy of a prompt by observing how well it performed on a specific set of queries within a task category. However, such signals unavoidably confound\footnote{Confounding factors here refer to variables such as intrinsic task difficulty or domain complexity that influence the outcome but are not caused by the prompt itself.} the intrinsic effect of the prompt with the underlying (often unaccounted) characteristics of the queries. For instance, prompts tested disproportionately on easier queries appear better than they are; those used on harder queries appear worse. As a result, when the optimizer explores novel prompts or encounters queries that diverge from the training distribution, these correlation-based rewards often break down, guiding the system toward prompts that appear optimal in historical data but fail to generalize in practice due to spurious correlations.

A causal framing offers a principled pathway out of this unreliability. At its core, reliable prompt evaluation requires answering a causal question: \emph{How would model performance change if the prompt were altered while holding the underlying query constant?} Answering this requires isolating the treatment effect of prompt variations from confounding task heterogeneity, such as intrinsic task complexity or domain-specific structure~\citep{imbens2015causal}. Without this causal separation, organizations cannot reliably discern whether a performance gain is attributable to a superior prompt or merely the result of an easier task instance, leading to suboptimal decision-making in automated workflows.

To operationalize this causal perspective, we propose \emph{Causal Prompt Optimization (CPO)}, a framework that estimates and exploits the causal effects of semantic variations in prompts. CPO begins by mapping queries and prompts into a structured semantic representation that captures meaningful linguistic and query-level features~\citep{ellickson2024using}. With this representation, the framework utilizes techniques from modern causal machine learning, specifically DML ~\citep{chernozhukov2018double,wager2018estimation,shi2025what}, to estimate the Conditional Average Treatment Effect (CATE) of changes in prompt semantic variations on model performance. By explicitly separating prompt semantics from query characteristics, CPO produces a \emph{causal reward model} that captures how interventions to the prompt causally influence outcomes, rather than merely how they correlate with historical usage patterns. A key implication of this causal formulation is that it induces a well-defined ranking over prompts based on estimated treatment effects. This, in turn, answers a core empirical question when properly evaluated: \emph{If the model predicts that prompt A will outperform prompt B based on estimated causal effects, does this prediction actually hold when evaluated on unseen queries and prompt instances?}

Building on the causal reward model, CPO uses it to guide a scalable search for improved prompt strategies. This optimization-driven use of causal estimates naturally answers our second question: \emph{Do prompts discovered through causal-guided optimization deliver systematic performance improvements relative to existing APO methods?} Instead of evaluating every candidate prompt by actually querying the LLM, CPO leverages the estimated causal reward model to score candidates during optimization. This enables broad and computationally efficient exploration of the prompt design space. The search process includes a structured mechanism for generating candidate improvements, drawing on the LLM’s own capacity for semantic refinement, and evaluates each candidate using the causal reward model. By anchoring optimization in causal estimates rather than raw correlations or expensive online evaluations, CPO identifies prompts that generalize more reliably across heterogeneous and previously unseen queries. 

We evaluate CPO across three heterogeneous and business-relevant benchmarks, mathematical reasoning (MATH), visualization (VisEval), and data analytics (DABench), to assess its effectiveness under diverse query characteristics (e.g., structures and difficulty profiles). Across all settings, CPO identifies prompts that outperform human-engineered baselines, conventional prompting methods such as Reflexion~\citep{shinn2023reflexion}, and state-of-the-art automated optimizers such as PromptBreeder~\citep{fernando2024promptbreeder} and TextGrad~\citep{yuksekgonul2025optimizing}. The framework achieves top overall accuracy on every benchmark and maintains particularly strong performance on the most challenging subsets, such as complex mathematical problems and high-difficulty visualization and data-analytics queries where existing correlational approaches typically deteriorate. Beyond the final optimization performance, the intermediate causal reward model underlying CPO demonstrates substantially stronger agreement with true prompt rankings than non-causal predictive models. When tested on unseen queries and prompts, the causal estimator consistently yields higher concordance with ground-truth outcomes, indicating that it captures stable causal relationships rather than spurious correlations in historical logs. These findings establish that CPO can effectively discover high-performing prompts, primarily relying on a reward signal that generalizes reliably across new and heterogeneous queries and prompts.

A series of further analyses further validates the mechanisms through which CPO achieves performance gains. Ablation studies show that replacing the causal reward with a non-causal predictive model, while holding all other components constant, substantially reduces optimization performance, especially on hard queries. This demonstrates that CPO’s improvements stem from its ability to separate prompt effects from query characteristics rather than from differences in model architecture or search procedure. Additional validation confirms that the latent treatment representation used in CPO is semantically meaningful: the principal components extracted from prompt embeddings align systematically with interpretable prompt-design patterns such as constraint strictness, structural framing, and guidance style. Finally, CPO exhibits favorable scaling behavior with accumulated offline data. The causal reward model shows consistent improvements in both ranking accuracy and optimization performance as data volume grows, while predictive baselines fluctuate or decline. This pattern suggests increasing returns to historical data—an important operational consideration for enterprises whose LLM systems continuously accumulate interaction logs.

These empirical findings support three contributions. \emph{First}, we conceptualize prompt engineering as a causal estimation problem, departing from the correlation-based paradigms prevalent in prior work. We identify that standard correlational reward signals inherently confound the intrinsic efficacy of a prompt with confounding factors such as difficulty. By formalizing this confounding bias, we establish the necessity of isolating the causal treatment effect of semantic variations to ensure reliable optimization.
\emph{Second}, we develop a methodological framework that implements this causal perspective through DML on semantic embeddings. By orthogonalizing prompt features against query characteristics, our approach constructs a causal reward model that is robust to spurious correlations. This design allows our approach to target the true value-added of a prompt rather than relying on historical associations that may not generalize.
\emph{Third}, we demonstrate the operational and economic value of \textit{dynamic}, query-level adaptation, compared with \textit{static}, task-level approaches. By enabling high-precision prompt customization at a fraction of the cost of online simulation, CPO resolves the scalability bottleneck that has historically limited enterprises to static, ``one-size-fits-all'' prompts. This dynamic capability not only improves robustness on heterogeneous and complex tasks but also converts accumulated interaction data into progressively stronger, adaptive prompt optimization approaches.

Taken together, these contributions seek to move prompt design for LLM-based operations from ad hoc experimentation toward a more systematic and data-efficient practice. Rather than offering yet another heuristic for writing prompts, CPO provides a framework for learning how prompt features affect performance across heterogeneous queries and for leveraging that understanding to design better prompt policies under realistic budget and deployment constraints. In summary, CPO establishes a principled and operationally grounded foundation for prompt optimization, providing the reliability and efficiency required for the next generation of AI-enabled organizational systems.

\section{Literature Review}
\label{sec:lit_review}

Our research is situated at the intersection of prompt engineering, automated optimization, and causal machine learning. As LLMs become increasingly central to organizational workflows, the information systems literature has emphasized the need for systematic frameworks to design reliable and robust AI systems \citep{rai2024pathways, susarla2025inventing, yoo2024next}. Within this broader agenda, we focus specifically on the challenge of prompt optimization---how to systematically control LLM behavior through natural language instructions. We review these streams to position CPO not merely as an algorithmic improvement, but as a strategic response to the fundamental operational and economic limitations of existing APO methods.

\subsection{Prompt Engineering: From Heuristic Art to Systematic Design}

Prompt engineering is the practice of crafting and refining input prompts to effectively guide LLMs to produce accurate, relevant, and high-quality responses. In managerial contexts, prompt engineering represents the primary mechanism for aligning LLM behavior with organizational goals without accessing internal model parameters.

Historically, this process has been predominantly manual and heuristic. Practitioners typically engage in an iterative cycle of trial-and-error, experimenting with diverse prompt formats, lexical choices, and structural constraints—such as varying the number of exemplars or adjusting the tone of instructions—to modulate model behavior~\citep{li2021prefix, reynolds2021prompt, zhou2022large}. As these manual practices accumulated, two major paradigms emerged to systematize this control: \textit{example-based prompting} and \textit{reasoning-oriented prompting}. Example-based prompting techniques, such as few-shot prompting~\citep{brown2020language}, utilized task-specific exemplars to stimulate in-context learning. Subsequently, Chain-of-Thought (CoT) reasoning~\citep{wei2022chain, wangself, kojima2022large} and its extensions like Tree-of-Thought~\citep{yao2023tree} demonstrated that enforcing structured logical steps could significantly enhance performance on complex tasks. Beyond structuring the reasoning process, researchers have developed mechanisms to enhance contextual grounding and domain specificity. Role-based prompting~\citep{kong2024better} explicitly instructs the model to adopt a specific professional persona (e.g., ``You are a senior risk analyst''), thereby aligning the output's tone and terminology with functional expectations. Retrieval-Augmented Generation (RAG)~\citep{lewis2020retrieval} goes further by dynamically injecting relevant external data into the prompt context, significantly bolstering the factual reliability.

Despite these advancements, manual prompt engineering faces a severe scalability barrier in enterprise deployment. It is highly labor-intensive, requiring domain-specific expertise to craft prompts that often prove ``brittle''—failing to transfer across different models or evolving task distributions~\citep{zhou2022large, sahoo2024systematic}. In a managerial context, this lack of reproducibility creates significant operational risk and cost, as critical business workflows may rely on unstable, hand-tuned instructions that degrade unpredictably when inputs vary. This bottleneck necessitates a shift from ad-hoc human crafting toward automated, data-driven optimization frameworks that can systematically guarantee performance.

\subsection{Automated Prompt Optimization}
To overcome the limitations of manual prompt engineering, the field has advanced toward APO. This framework recasts prompt design as a computational search problem within the discrete text space~\citep{pryzant2023automatic, cheng2024black}, aiming to systematically identify optimal prompts. However, navigating this unstructured text space presents unique challenges distinct from traditional numerical optimization. The search space is discrete and semantically non-linear; minor textual edits can precipitate disproportionate and unpredictable shifts in performance, lacking stable directional signals to guide improvement~\citep{liu2023pre}.
To manage this complexity, most APO systems follow an iterative \emph{generate-evaluate-filter} pipeline: they begin with seed prompts (manually specified or generated by an LLM), produce variants in textual space (via rewrite rules, mutation operators, or LLM proposals), and screen candidates using some feedback rewards. To navigate this complex search space, these systems rely on established search strategies ranging from greedy selection to evolutionary heuristics~\citep{ramnath2025systematic}. We categorize these methods through an economic and operational lens, categorizing them into \emph{Static} versus \emph{Dynamic} approaches.

\subsubsection{Static Approach: Task-Level Prompt Optimization}
The dominant paradigm, exemplified by methods like APE~\citep{zhou2022large}, OPRO~\citep{yang2023large}, and PromptBreeder~\citep{fernando2024promptbreeder},  aims to identify a single, universal prompt that maximizes average performance across an entire task or dataset.
They leverage LLMs to produce candidates—either by reverse-engineering demonstrations~\citep{zhou2022large} or iteratively mutating seed instructions~\citep{cui2024phaseevo, yang2023large}—and filter them using heuristic search algorithms such as beam search, Monte Carlo tree search or bandits~\citep{pryzant2023automatic, wangpromptagent,wu2024prompt}. Crucially, evaluating candidate prompts requires extensive validation on training data, making this evaluation process the primary cost driver and turning prompt optimization into a large fixed upfront investment. However, once the optimal prompt is locked, the marginal cost of deployment is low, as the same static prompt is reused for all future queries.

Despite its economic appeal for deployment, the static approach inherently assumes that a single prompt can suffice for diverse queries. In complex business environments characterized by heterogeneous or ``long-tail'' queries, this policy often yields suboptimal outcomes. By forcing a ``one-size-fits-all'' solution, static optimization inherently sacrifices performance on difficult edge cases to optimize for the average~\citep{sun2024querydependent}.

\subsubsection{Dynamic Approach: Query-Level Prompt Optimization}
To address query heterogeneity within a task category, recent approaches seek to implement dynamic approaches that tailor prompts to individual queries at a fine granularity. Unlike static methods that freeze a single prompt, dynamic approaches aim to learn a mapping function, often parameterized as a policy model, that generates context-aware prompts for specific queries.
The literature has achieved this adaptation by training dedicated optimization models via reinforcement learning (RL) or supervised fine-tuning. For instance, methods like TRPrompt~\citep{nica2025trprompt} fine-tune pre-trained models using textual feedback, while frameworks like Prompt-OIRL~\citep{sun2024querydependent} and QPO~\citep{kong2025query} employ offline inverse reinforcement learning. These methods first learn a reward function from historical logs and subsequently train a policy model to maximize this inferred reward, enabling personalized adaptation.

However, relying on learned policy models introduces significant challenges, primarily due to the flawed nature of the reward signals used to train them. To avoid the prohibitive latency of online simulation, dynamic methods typically rely on offline reward predictors. Yet, these standard predictors minimize prediction error based on correlational signals, failing to disentangle the intrinsic effectiveness of a prompt from underlying confounding factors, such as intrinsic query difficulty or domain complexity. For example, a prompt might receive a high score merely because it was historically applied to easy queries, leading the model to learn spurious correlations (such as favoring specific keywords or lengths) rather than true reasoning patterns~\citep{deng2022rlprompt}. This correlation-causation fallacy generates unstable signals that mislead the optimizer. When a policy model is trained on these confounded signals, it learns to exploit ``shortcuts'' rather than causal mechanisms, resulting in brittle policies that require continuous data recollection and re-training to adapt to novel or difficult queries. This limitation underscores the critical need for a framework capable of rigorous causal estimation to guide robust optimization.

\subsection{Causal Machine Learning for Robust Decision Making}

Our methodological foundation draws from causal machine learning, specifically the estimation of heterogeneous treatment effects (HTE) in high-dimensional settings. While traditional machine learning models are powerful tools for capturing complex relationships, they are typically optimized for prediction (i.e., exploiting correlations) rather than valid intervention (i.e., estimating unbiased effects)~\citep{wager2018estimation, athey2019generalized}. This distinction is critical in prompt optimization, where the goal is not merely to predict the score of a prompt, but to estimate the causal gain of changing the prompt.

To address this challenge, we leverage the DML framework~\citep{chernozhukov2018double}. The key innovation of DML is \textit{orthogonalization}: by utilizing flexible machine learning models to estimate nuisance parameters (e.g., the baseline difficulty of a query), DML effectively separates or partials out the effects of high-dimensional confounders from the treatment. This enables unbiased inference even when the underlying data-generating process exhibits complex confounding.

Due to its robustness, causal machine learning has proven significantly valuable in solving business problems~\citep{hunermund2022causal, ellickson2024using}. In particular, DML has been successfully applied to isolate the causal impact of strategic interventions from environmental noise. For instance, \cite{ellickson2023estimating} apply this framework to estimate the heterogeneous effects of targeted email promotions, specifically showing how variations in content, framing, and personalization causally influence consumer engagement and conversion across the marketing funnel. Similarly, \cite{more2023double} utilize it to estimate the causal impacts of hundreds of customer actions, achieving higher accuracy and computational efficiency while enabling large-scale causal inference in business applications. Beyond marketing contexts, causal machine learning has proven valuable for understanding heterogeneous treatment effects across diverse applications \citep{leng2024calibration}.

CPO represents a novel translation of this framework into the domain of Natural Language Processing (NLP). While there is a growing body of work at the intersection of causal inference and NLP, most existing research focuses on representation learning for interpretability or fairness, such as learning invariant representations to remove spurious correlations~\citep{shi2021invariant} or using causal embeddings to explain model behaviors~\citep{xu2025causal}. 
We depart from this interpretability-focused paradigm by operationalizing causal inference as an optimization engine. Specifically, we utilize Principal Component Analysis (PCA)-reduced semantic embeddings as continuous treatments to estimate the CATE of prompt variations. This allows us to mathematically isolate the value-add of a prompt from the intrinsic difficulty of a task instance, providing the rigorous statistical foundation necessary to automate reliable, query-adaptive prompting.

\section{Methodology}
\label{sec:methodology}
In this section, we present the CPO framework. We begin by formally defining query-level prompt optimization as a reward maximization problem (Section~\ref{subsec:problem_formulation}). We then introduce our core methodological innovation: treating prompt design as a causal inference challenge (Section~\ref{subsec:causal_formulation}). The framework proceeds in two distinct stages: \emph{Stage 1} employs DML on PCA-reduced semantic embeddings to learn a causal reward model (Section~\ref{subsec:stage1}), while \emph{Stage 2} utilizes this model to guide a resource-efficient, tree-based search for better prompts (Section~\ref{subsec:stage2}).

\subsection{Problem Formulation}
\label{subsec:problem_formulation}
We are given a representative task dataset
$\mathcal{D}_{\text{task}} = \{(x_i, l_i)\}_{i=1}^N,$
where each $x_i$ is a natural language query description (e.g., a math problem), and $l_i$ is the corresponding ground-truth label or solution.
Let $LLM_{\text{task}}$ be a task-execution language model, and let $\mathcal{E}(\cdot,\cdot)$ denote an evaluation function (e.g., accuracy) that maps a model-generated response and a reference answer to a scalar performance score. For a representative query $x \in \mathcal{X}$ and prompt $t \in \mathcal{T}$, where $\mathcal{X}$ and $\mathcal{T}$ denote the spaces of queries and candidate prompts, the language model generates a response $LLM_{\text{task}}(x, t)$ conditioned on both the query and the prompt. This response is evaluated against the reference $l$ associated with $x$ to produce a score 
$y = \mathcal{E}\left(LLM_{\text{task}}(x, t), l\right),$ where a higher score represents closer agreement between the model output and the ground-truth label, and thus better task performance.

\begin{figure*}[!ht]
    \begin{center}
    \includegraphics[width=0.9\textwidth]{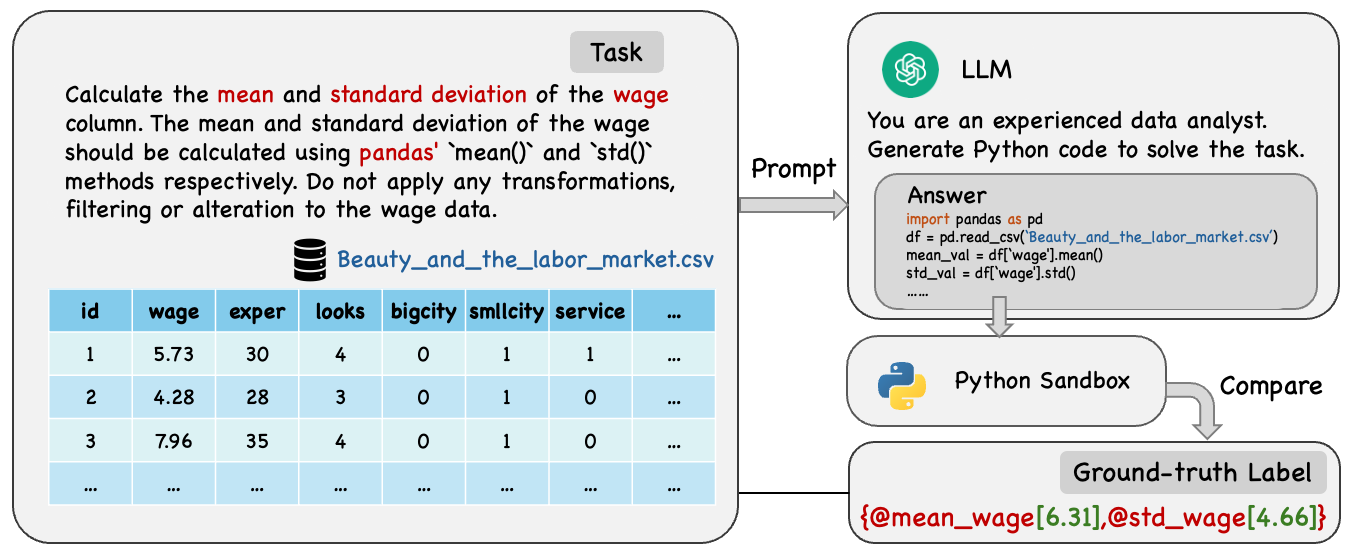}
    \end{center}
    \caption{An Example of Analytical Task}
    \label{fig:dabench_example}
\end{figure*}

Figure~\ref{fig:dabench_example} provides an illustrative example drawn from DABench~\citep{hu2024infiagent}, one of our experimental datasets on data analytics. In this example, the input $x$ is a natural language query that asks the model to compute the mean and standard deviation of a specified column in a given dataset. The corresponding label $l$ is the ground-truth numerical values of the mean and standard deviation computed from the data. Given a prompt $t \in \mathcal{T}$ that specifies the role assumed by the model and outlines additional information, such as a general task-solving strategy, the task-execution model $LLM_{\text{task}}$ produces the code that is executed in a sandbox environment to generate an output, which is subsequently evaluated against the corresponding label $l$. The evaluation function $\mathcal{E}\left(LLM_{\text{task}}(x, t), l\right)$ assigns a score of 1 if both the mean and the standard deviation are correctly computed, a score of 0 if neither value is correct, and a score of 0.5 if exactly one of the two values is correct.

Our goal is to identify the prompt that provides the maximum score for a specific query $x$. Formally, we seek a prompt $t^* \in \mathcal{T}$ that maximizes the evaluation score:
\begin{equation}
t^* = \arg\max_{t \in \mathcal{T}} \; \mathcal{E}\left(LLM_{\text{task}}(x, t), l\right).
\label{eq:instance_opt}
\end{equation}
This formulation reflects a query-level optimization goal, where the optimal prompting strategy adapts to the specific characteristics of the query $x$. For convenience, a summary of major notation is provided in Appendix~\ref{sec:notation}.

\subsection{Prompt Optimization as Causal Inference}
\label{subsec:causal_formulation}

We formulate prompt optimization not merely as a search problem, but as a problem of \emph{causal inference}. Standard supervised learning often conflates correlation with causation; for instance, ``hard'' tasks may inherently lead to lower scores regardless of the prompt, while simultaneously inducing users to attempt more complex prompts. A correlational model might erroneously learn that complex prompts \emph{cause} lower scores. To resolve this, we view each text prompt as a \emph{treatment} and the model’s performance as the \emph{outcome}, aiming to isolate the causal effect of the prompt from task-specific confounders.

Let $X \in \mathcal{X}$ denote the query features, and $Y(t) \in [0,1]$ the evaluation score. Following the potential outcomes framework~\citep{imbens2015causal}, the \emph{potential outcome} under prompt $t\in \mathcal{T}$ is defined as:
\begin{equation}
\mu(x, t) := \mathbb{E}[Y(t) \mid X = x],
\label{eq:mu}
\end{equation}
where $Y(t)$ is the score obtained from evaluating the response of query $x$ under prompt $t$.

To measure the true effectiveness of a prompt, we define the CATE of a candidate prompt $t$ relative to a baseline prompt $t_0$:
\begin{equation}
\tau(x, t) := \mu(x, t) - \mu(x, t_0)
           = \mathbb{E}[Y(t) - Y(t_0) \mid X = x].
\label{eq:cate}
\end{equation}
This quantity $\tau(x, t)$ represents the \emph{counterfactual gain}: how much better the model performs with prompt $t$ compared to the baseline $t_0$, strictly due to the prompt itself. The objective of query-level optimization is to select the prompt that maximizes this causal effect:
\begin{equation}
t^*(x) = \arg\max_{t \in \mathcal{T}} \tau(x, t),
\label{eq:policy}
\end{equation}
which connects to the empirical prompt-search objective in \eqref{eq:instance_opt} by replacing the realized evaluation score for a single execution with its counterfactual, feature-conditional expectation and by explicitly accounting for confounding query characteristics. Note that estimating either $\mu(x,t)$ or $\tau(x,t)$ yields the same optimal prompt, since for a fixed baseline prompt $t_0$, $\arg\max_t \tau(x,t) = \arg\max_t \mu(x,t)$. We adopt the CATE formulation primarily for interpretability: it directly captures the \emph{causal improvement} of a candidate prompt over a control prompt, while in practice, its estimation complexity and accuracy are essentially identical to those of estimating $\mu(x,t)$ directly.

\begin{figure*}[h]
    \begin{center}
        \includegraphics[width=\textwidth]{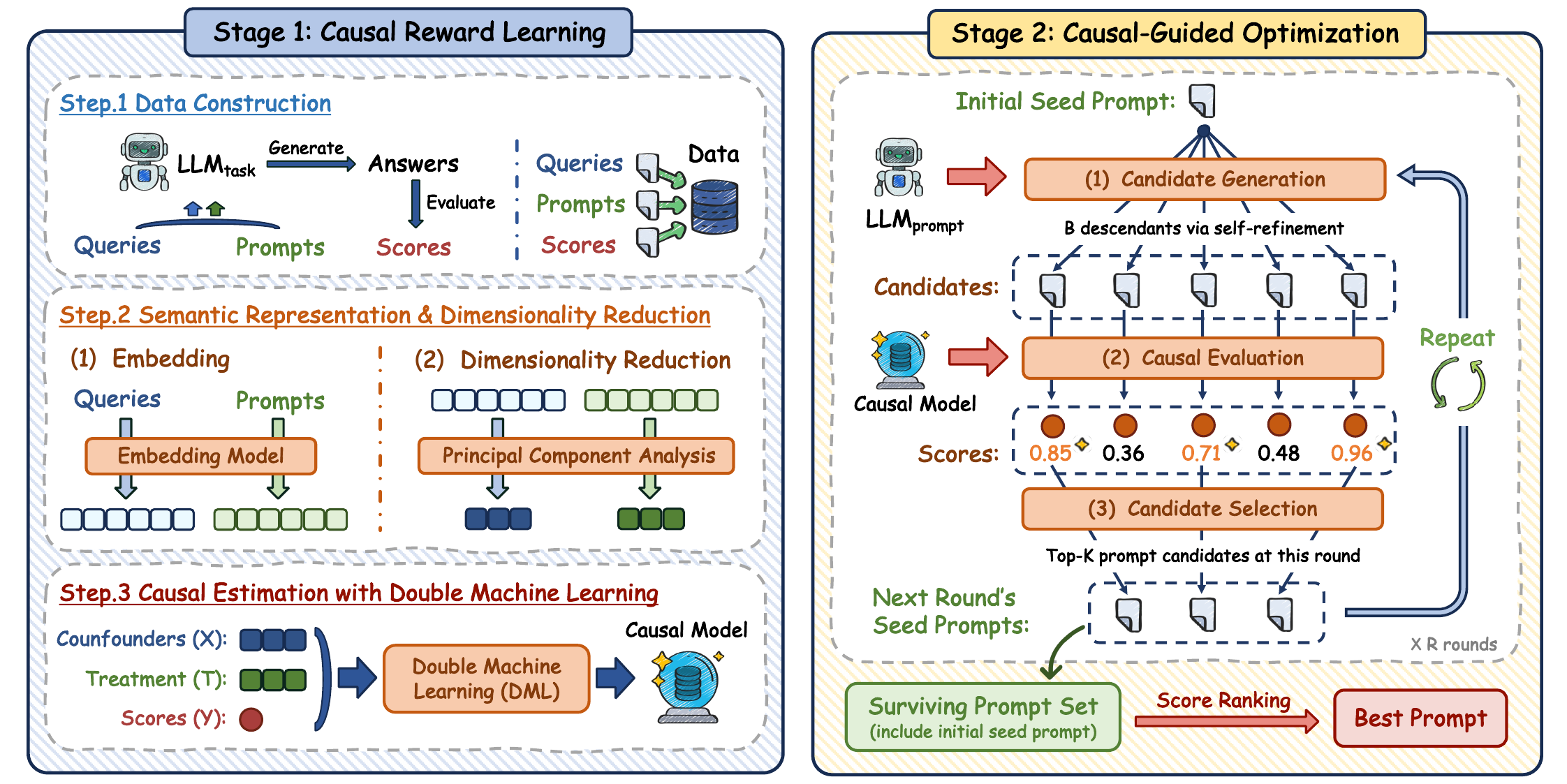}
    \end{center}
    \caption{Overview of the Proposed Causal Prompt Optimization (CPO) Framework}
    \label{fig:framework}
\end{figure*}

Figure~\ref{fig:framework} illustrates the overall architecture of the Causal Prompt Optimization (CPO) framework, which proceeds in two stages: (1) Causal Reward Learning and (2) Causal-Guided Optimization.

\subsection{Stage 1: Causal Reward Learning}
\label{subsec:stage1}
\subsubsection{Step 1: Data Construction}
\label{subsubsec:data-construction}
To estimate $\tau(x,t)$, we construct a dataset consisting of triplets $\{(x_i, t_i, y_i)\}_{i=1}^N$. For a benchmark dataset $\mathcal{D}_{\text{task}}$, we systematically vary prompt templates while holding the query $x_i$ constant. These variations differ along key dimensions, such as constraint inclusion, instructional style, or few-shot examples, creating a rich treatment space. To anchor causal estimation, we also define a \emph{control prompt} $t_0$ for each dataset, which is typically a simple instruction without few-shot examples. This balanced design ensures that for any given query $x$, we observe outcomes under multiple prompt treatments, providing the necessary variation to disentangle prompt effects from effects of query characteristics (see Section \ref{subsec:offline-data-construction} for experimental implementation details).

\subsubsection{Step 2: Semantic Representation and ``Latent Treatments''}
\label{subsubsec:semantic-representation}
A central challenge in causal modeling over text is that both prompts and queries are inherently discrete, making it difficult to define meaningful treatment and control variables. We address this by defining a continuous ``latent treatment'' space through a two-stage transformation.

We first map each query $x$ and prompt $t$ into dense embeddings using a sentence-transformer encoder \citep{nussbaum2025nomic}. These embeddings capture latent linguistic information (e.g., tone, structure) but remain too high-dimensional for efficient causal inference. We then apply PCA to obtain compact semantic representations. PCA orthogonalizes correlated linguistic features and yields a low-dimensional semantic space suitable for causal estimation:
\begin{equation}
\textbf{$\mathbf{x} = \psi_X(x) \in \mathbb{R}^{d_x}$}, \qquad
\textbf{$\mathbf{z} = \psi_T(t) \in \mathbb{R}^{d_t}$}.
\end{equation}
Here, $\mathbf{x}$ and $\mathbf{z}$ are vector representations in the learned semantic space: $\mathbf{x}$ summarizes query characteristics, and $\mathbf{z}$ provides a continuous treatment representation of the prompt. With a slight abuse of notation, we write $\hat{\tau}(x, t)$ to denote $\hat{\tau}(\mathbf{x}, \mathbf{z})$, emphasizing that all causal evaluation is conducted in the learned embedding space.

Given these representations, our object of interest is how prompt semantics $\mathbf{z}$ causally affect task performance $Y$, potentially in a query-adaptive manner characterized by $\mathbf{x}$. A natural starting point is the conditional expectation $\mathbb{E}[Y \mid \mathbf{x}, \mathbf{z}]$. However, because both $\mathbf{x}$
and $\mathbf{z}$ are high-dimensional vectors, this relationship is unlikely to admit a simple parametric form. In particular, performance varies substantially across queries even in the absence of prompt variation, implying that baseline query difficulty must be accounted for before isolating the causal effect of prompts. These considerations motivate a causal estimation strategy that (i) flexibly models outcome heterogeneity induced by $\mathbf{x}$, while (ii) isolating the causal contribution of $\mathbf{z}$ without imposing restrictive functional form assumptions.

\subsubsection{Step 3: Causal Estimation with Double Machine Learning}
Our goal in this step is to estimate the causal effect of prompt semantics $\mathbf{z}$ on task performance $Y$, allowing this effect to vary across queries characterized by $\mathbf{x}$. Although prompt assignment is experimentally controlled, estimation remains challenging because both 
$\mathbf{x}$ and $\mathbf{z}$ reside in high-dimensional semantic spaces. In such settings, standard parametric estimators impose restrictive functional form assumptions that are unlikely to hold, while classical semiparametric methods become infeasible due to the curse of dimensionality. To address these challenges, we adopt the DML framework \citep{chernozhukov2018double}, which combines flexible machine learning for nuisance estimation with orthogonalized estimating equations to deliver valid causal inference in high-dimensional environments \citep{shi2025what}. 

\begin{enumerate}
    \item[(1)] \textbf{Modeling Query Characteristics and Treatment Assignment} 
    We begin by modeling systematic variation in both outcomes and treatments as a function of query semantics. Even in the absence of prompt variation, queries differ substantially, implying that observed performance reflects both query characteristics and prompt effects. Failing to account for this heterogeneity can distort causal estimates, particularly when the mapping from semantic embeddings to outcomes is nonlinear. 
    
    To operationalize this objective within the DML framework, we impose a partially linear structure on the conditional mean, assuming that prompt effects enter linearly through their embedding representation while allowing unrestricted heterogeneity with respect to query semantics \citep{chernozhukov2018double}. This assumption preserves flexibility in how query characteristics affect performance, while enabling orthogonalized estimation of prompt effects in high-dimensional treatment spaces.
    
    Under this structure, DML introduces nuisance functions that summarize how outcomes and treatments vary with $\mathbf{x}$:
    \begin{equation}
        m(\mathbf{x}) = \mathbb{E}[Y \mid \mathbf{x}], \qquad e(\mathbf{x}) = \mathbb{E}[\mathbf{z} \mid \mathbf{x}].
    \end{equation}
    Here, $m(\mathbf{x})$ represents the baseline performance level implied by a query’s characteristics, while $e(\mathbf{x})$ captures systematic dependence between query characteristics and the continuous treatment representation. Because both relationships may be highly nonlinear in high-dimensional embedding spaces, we estimate $m(\cdot)$ and $e(\cdot)$ using flexible machine learning methods (Gradient Boosting), rather than imposing parametric restrictions.

    To prevent overfitting in this first stage from biasing causal estimates, we implement cross-fitting as prescribed by \cite{chernozhukov2018double}. Specifically, we partition the data into $K$ folds. For each fold $k$, we estimate the nuisance functions using the other $k-1$ folds and generate out-of-sample predictions for fold $k$, ensuring that subsequent residuals are constructed using cross-fitted predictions.

    \item[(2)] \textbf{Residualization via Orthogonalization} While flexible machine learning improves approximation accuracy, it introduces regularization bias that can contaminate causal estimates if left unaddressed. A central insight of DML is that this bias can be neutralized by constructing an estimating equation that is Neyman orthogonal to the nuisance functions. Orthogonality ensures that small estimation errors in $m(\cdot)$ and $e(\cdot)$ affect the treatment effect estimator only at second order.

    Operationally, orthogonality is achieved by residualizing both the outcome and the treatment using cross-fitted nuisance estimates:
    \begin{equation}
        \tilde{Y} = Y - m(\mathbf{x}), \qquad \tilde{\mathbf{z}} = \mathbf{z} - e(\mathbf{x}).
    \end{equation}
    This step orthogonalizes the prompt signal against the query signal. For instance, if a query is inherently difficult, $m(\mathbf{x})$ will predict a low score; the residual $\tilde{Y}$ thus isolates the \emph{excess performance} attributable specifically to the prompt variation, independent of the query's baseline difficulty.

    \item[(3)] \textbf{CATE Estimation} After orthogonalization, we estimate heterogeneous treatment effects by exploiting the partially linear structure implied by DML. Specifically, the orthogonalized outcome satisfies
    \begin{equation}
        \tilde{Y} = \theta(\mathbf{x})^\intercal \tilde{\mathbf{z}} + \varepsilon,
        \label{eq:residual_dml}
    \end{equation}
    where $\theta(\mathbf{x})$ denotes the conditional average treatment effect as a function of query characteristics. This specification preserves linearity in the treatment while allowing treatment effects to vary flexibly across the semantic query space.

    We estimate $\theta(\mathbf{x})$ using a Generalized Random Forest \citep{athey2019generalized}, which is well suited for learning heterogeneous coefficients in high-dimensional settings. Because the estimating equation is Neyman orthogonal to the nuisance functions, the resulting estimator remains locally insensitive to regularization bias in the first-stage nuisance estimates.

\end{enumerate}

The estimated conditional average treatment effect is therefore given by
\begin{equation}
    \hat{\tau}(x, t) = \hat{\theta}(\mathbf{x})^\intercal(\mathbf{z}-\mathbf{z_0}),
    \label{eq:tau_def}
\end{equation}
which yields a robust, de-biased reward signal that guides the optimization in Stage 2.

\subsection{Stage 2: Causal-Guided Optimization}
\label{subsec:stage2}

In the second stage, we leverage the estimated causal effect $\hat{\tau}(x,t)$ as an offline reward function for prompt optimization. This design offloads the heavy prompt-evaluation workload, which otherwise requiring repeated calls to the task-execution LLM (i.e., $LLM_{\text{task}}$), to an offline causal reward model as a one-time fixed investment, leaving only lightweight marginal costs during prompt search. As a result, we directly address the primary economic bottleneck of standard APO methods, which scale linearly with the number of prompt evaluations.

\begin{figure*}[!ht]
    \begin{center}
        \includegraphics[width=0.7\textwidth]{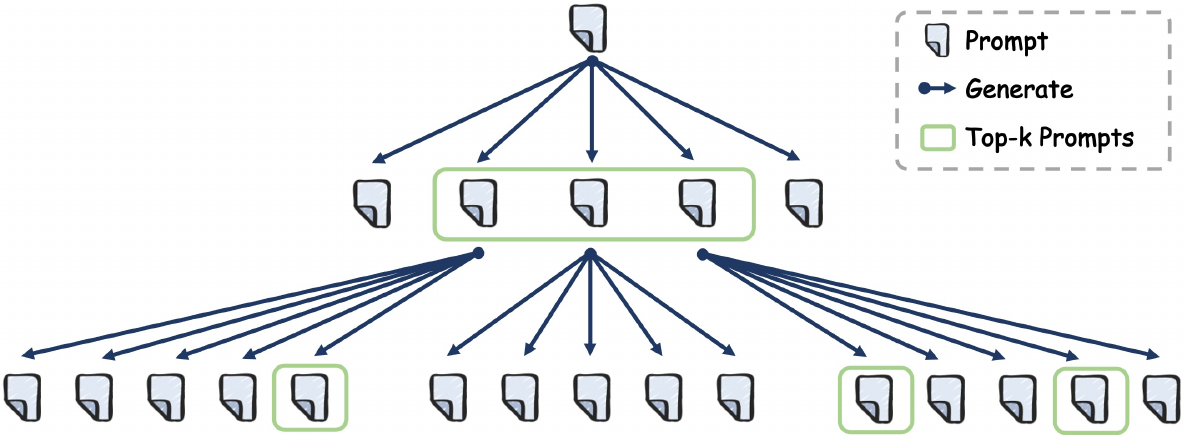}
    \end{center}
    \caption{An Illustration of the Tree Search Process in Prompt Optimization ($B{=}5$, $K{=}3$, $R{=}2$)}
    \label{fig:tree-search}
\end{figure*}

Using this offline reward, we conduct prompt search via an iterative tree-search procedure (Figure~\ref{fig:tree-search}):
\begin{enumerate}
    \item[(1)] \textbf{Candidate Generation:} Starting from a set of seed prompts (baseline prompt $t_0$ in the first round), a generator model $LLM_{\text{prompt}}$, which is typically more lightweight than the task-execution model, produces $B$ candidate variants, where $B$ denotes the number of generated variants per prompt. To promote meaningful exploration, we embed an explicit \emph{self-refinement instruction} (e.g., ``develop superior prompts based on the seed prompt'') within the generation prompt. This encourages the model to propose semantically improved prompts rather than superficial paraphrases, enabling exploration of higher-quality regions of the prompt space. Figure \ref{fig:prompt-math} illustrates the template used in the mathematical-reasoning task (see Section~\ref{subsec:benchmark-datasets} for further details of this task), which consists of (1) a system prompt defining the generation goal and style where a self-refinement instruction is included, (2) instruction and example blocks that guide reasoning, and (3) a user prompt specifying the number and format of new candidate prompts.
    \item[(2)] \textbf{Causal Evaluation:} Instead of querying the expensive $LLM_{\text{task}}$ for every candidate, we project the new prompts into the PCA space and evaluate them using the learned causal model to obtain $\hat{\tau}(x, t)$. This step provides a fast estimate of each prompt’s counterfactual performance gain relative to the baseline, without incurring online inference costs.
    \item[(3)] \textbf{Selection and Global Ranking:} We retain the top-$K$ candidates with the highest predicted causal gain to seed the next search iteration. This iterative selection-and-expansion process is repeated for $R$ rounds, after which we aggregate all surviving prompts into a global set and re-rank them according to $\hat{\tau}(x,t)$, selecting the final optimal prompt $t^*$.
\end{enumerate}

This approach effectively decouples exploration from execution. By evaluating thousands of candidates via the lightweight causal reward model, CPO affords deep exploration of the semantic space at a fraction of the cost of online methods (e.g., LLM-as-a-Judge). Moreover, because the reward is explicitly causal rather than correlational, the resulting prompt selection is more robust to query heterogeneity and distribution shifts, yielding prompts that generalize beyond the specific evaluation sample.

\begin{tcolorbox}[breakable, enhanced, base, title={Prompt for MATH to Generate New Prompt Candidates}]
    \begin{lstlisting}
SYSTEM PROMPT
As a creative prompt engineer, your mission is to craft diverse and innovative prompts for solving math problems, inspired by the provided seed prompt. These prompts should include clear `instructions` and `examples`. The `instructions` should provide general guidance for the language model to solve a wide range of math problems, avoiding specific directions for any particular type of math problem. The `examples` are question-solution demonstrations. Your objective is to develop superior prompts based on the seed prompt, aiming to boost the language model's performance.

Note: 
1. For `instructions`, you should provide general instructions for the language model to solve math problems, but DO NOT write something like `Instruction:` or `Instructions:` before the instruction, use the instruction itself as the beginning.
2. For `examples`, you can design examples that are not limited to the specific type of math problem. Each example should follow the format of `Question` and `Solution`, and use \\boxed{} to wrap the final answer. For instance, this is an example of `examples`:
```
Examples:
Example 1:
Question: Find the positive base $b$ in which the equation $13\\cdot15=243$ is valid.
Solution: When we rewrite the above equation with the base numbers as sums of digit bundles we arrive at the following to work with: \\begin{align*}\n13_b\\cdot15_b&=243_b\\quad\\Rightarrow\\\\\n(b+3)(b+5)&=2b^2+4b+3\\quad\\Rightarrow\\\\\nb^2+8b+15&=2b^2+4b+3\\quad\\Rightarrow\\\\\n0&=b^2-4b-12\\quad\\Rightarrow\\\\\n0&=(b-6)(b+2).\n\\end{align*} Since $b$ must be positive, the necessary base is base $\\boxed{6}$.
                         
Example 2:
Question: Find $x$ such that $\\lceil x \\rceil + x = \\dfrac{23}{7}$. Express $x$ as a common fraction.
Solution: First, we note that $x$ must be positive, since otherwise $\\lceil x \\rceil + x$ is nonpositive. Next, we know that the decimal part of $x$ must be $\\dfrac{2}{7}$. We write $x$ as $n+\\dfrac{2}{7}$, where $n$ is the greatest integer less than $x.$ Then, $\\lceil x \\rceil = n + 1.$ Therefore, we can write $\\lceil x \\rceil + x$ as $n+1+n+\\dfrac{2}{7}=\\dfrac{23}{7}$. Solving, we get $n=1$. Therefore, the only value $x$ that satisfies the equation is $1+\\dfrac{2}{7}=\\boxed{\\dfrac{9}{7}}$.
```

USER PROMPT
Here is a seed prompt for solving math problems:

{seed prompt}

The seed prompt may only include `instructions`, i.e., no `examples`, but you are required to generate prompts including `instructions` and several `examples` (the number of examples is from 0 to 7). Now, please generate {self.candidates_per_prompt} new, diverse prompts that could be used for solving math problems. 

Generate {self.candidates_per_prompt} prompts in the following format:

### Prompt 1:
<prompt>
[Your generated prompt text here]
</prompt>

### Prompt 2:
<prompt>
[Your generated prompt text here]
</prompt>

And so on...
\end{lstlisting}
\end{tcolorbox}
\vspace{-1em}
\captionof{figure}{Prompt Template for Generating New Candidate Prompts in the MATH Task}
\label{fig:prompt-math}
\setlength{\parindent}{1em}

\section{Experiments}
\subsection{Data Preparation}
\label{subsec:data-preparation}

\subsubsection{Benchmark Datasets}
\label{subsec:benchmark-datasets}

To comprehensively evaluate the proposed CPO framework, we conduct experiments on three diverse tasks, including mathematical reasoning, natural language to visualization generation, and data analytics. For each task, we select a corresponding typical benchmark dataset: MATH~\citep{hendrycks2021measuring} for mathematical reasoning, VisEval~\citep{chen2024viseval} for natural language to visualization generation, and DABench~\citep{hu2024infiagent} for data analytics. See the detailed descriptions and the evaluation metrics (e.g., accuracy and pass rate) of the benchmarks in the Appendix~\ref{sec:data_description}.

\subsubsection{Offline Data Construction}
\label{subsec:offline-data-construction}
As outlined conceptually in Section \ref{subsubsec:data-construction}, we construct offline datasets of $\langle$query, prompt, score$\rangle$ triplets to enable causal estimation in Stage 1 of CPO. The data-collection pipeline follows a unified procedure across all three benchmarks while allowing task-specific adaptations. Table \ref{tab:offline-stats} summarizes the dataset statistics.

\begin{table*}[!ht]
    \centering
    \caption{Statistics of the Constructed Offline Data for the Three Benchmark Datasets}
    \tabcolsep=0.029\linewidth
    \small
    \renewcommand{\arraystretch}{1.15}
    \label{tab:offline-stats}
    \begin{threeparttable}
    \begin{tabular}{l c c c c c}
        \toprule
        Task & \#Queries & \#Prompts & \#Instructions & \#Examples & \#Triplets \\
        \midrule
        MATH    & 185 & 201 & 116 & 29 & 37,185 \\
        VisEval & 102  & 201  & 30 & 64 & 20,502 \\
        DABench & 164 & 160 & 10 & 3 & 26,240    \\
        \bottomrule
    \end{tabular}
    \end{threeparttable}
\end{table*}

For the MATH and VisEval datasets, each prompt consists of two elements: an \emph{instruction} and a set of \emph{few-shot examples}. We begin with the baseline prompt $t_0$ and expand the prompt pool in a binary-tree manner. Specifically, we use $LLM_{\text{prompt}}$ to generate two new variants from the initial prompt ($t_0$); then, for each newly generated prompt, we again generate two variants, and repeat this process recursively. During this process, the model is required to preserve the original intent while varying linguistic style, structural framing, and explanatory tone, thereby producing diverse yet semantically aligned instructions. Few-shot examples are query–answer pairs randomly sampled from the benchmark, combined to form multiple demonstration sets. Pairing distinct instructions with these examples yields a large collection of complete prompts. 
We also designed a baseline prompt $t_0$ for each task to anchor the CATE estimation, which is a simple instruction that does not include any few-shot examples. For example, for the MATH dataset, the baseline prompt is ``Solve the following math problem." The details of the baseline prompts are shown in Appendix~\ref{sec:baseline-prompts}.
For each $\langle$query, prompt$\rangle$ pair, we use $LLM_{\text{task}}$ (i.e., Qwen2.5-14B) to generate an answer, which is subsequently evaluated to obtain the performance metric. This process yields a comprehensive offline dataset of $\langle$query, prompt, score$\rangle$ triplets for each task.

For the DABench dataset, we follow a similar procedure but introduce two additional types of prompt variation to accommodate its data-analysis nature.
First, each query in DABench includes explicit \emph{constraints} that influence the solution approach or output format. We therefore treat the inclusion of these constraints as one dimension of prompt variation.
Second, because tasks require referencing external data tables, we create a standardized JSON description for each dataset containing metadata such as column names, data types, extrema, and representative values. The inclusion or exclusion of this JSON context constitutes another design dimension.
In addition, we design three few-shot problem-solving demonstrations, each with an associated dataset, correct code implementation, and expected output. Prompts are generated under two configurations: (1) a \emph{one-shot} setting, which randomly incorporates a single demonstration, and (2) a \emph{three-shot} setting, which embeds all three demonstrations. This procedure ensures diversity in prompt complexity and informativeness. A baseline prompt is similarly designed for this task, as reported in Appendix~\ref{sec:baseline-prompts}.

\subsection{Validation Strategy}
\label{sec:validation}

To establish the reliability and practical value of the proposed CPO framework, we develop a validation strategy that evaluates both stages of CPO: (1) the causal reward model’s ability to estimate unbiased and generalizable treatment effects, and (2) the effectiveness of using these causal rewards to guide prompt optimization in practice.

\subsubsection{Stage 1: Validating Causal Reward Learning}
\label{subsubsec:stage1-validate}

The first stage evaluates whether the causal estimator $\hat{\tau}(x,t)$ can recover true treatment effects more accurately than non-causal predictors. It answers our first research question: \emph{If our model predicts that prompt A will outperform prompt B, does this prediction actually hold when tested on unseen queries and prompts?} Specifically, we focus on evaluating whether the estimated causal effects maintain generalizability across previously unseen queries and prompts. To this end, each offline dataset, drawn from the three aforementioned benchmarks, is partitioned into distinct training and validation sets. Both the causal and non-causal reward models are trained using the training split, and their performance is subsequently evaluated on the validation set. This evaluation design is essential because, in our causal formulation, prompts correspond to the \textit{treatments} and questions serve as potential \textit{confounders}. Robust performance on unseen prompts and queries demonstrates that the causal model successfully isolates the true causal contribution of prompts to model performance, ensuring reliable generalization beyond the observed data. A significant improvement over the non-causal ML baseline confirms that causal reward learning provides more reliable and generalizable effect estimates.

To provide a fair comparison, we introduce a non-causal machine learning (ML) baseline that employs the same underlying machine learning architecture as our DML estimator but is trained without causal adjustment. The baseline directly predicts task outcomes from concatenated prompt and query embeddings, focusing purely on predictive accuracy. In contrast, the DML-based causal estimator explicitly separates treatments from confounders and estimates their conditional effects, yielding unbiased measures of prompt influence. Therefore, comparing these two models isolates the unique contribution of causal reasoning to the reliability and generalizability of the learned reward signal.

We evaluate the quality of causal effect estimation using Kendall’s tau-b, a rank-based statistic that measures the ordinal consistency between the estimated causal effects and the ground-truth causal outcomes in held-out data~\citep{agresti2010analysis}. This metric is particularly appropriate because subsequent prompt optimization depends on ranking candidate prompts by their estimated effects, making rank consistency more relevant than absolute accuracy. We adopt Kendall’s tau-b rather than the traditional Kendall’s tau because it properly accounts for tied rankings~\citep{agresti2010analysis}---a frequent occurrence in our setting, where the true effects of multiple prompt–query pairs may be identical (e.g., performance scores of 1 or 0). By correcting for such ties, Kendall’s tau-b provides a robust and unbiased measure of ranking quality, ensuring fair comparison even when many samples share the same ground-truth outcomes. See the calculation details of Kendall’s tau-b in the Appendix \ref{sec:calc-kendall}.

\subsubsection{Stage 2: Validating Causal-Guided Optimization}

The second stage examines whether using $\hat{\tau}(x,t)$ as a reward function leads to practical performance gains in prompt optimization. Specifically, it answers our second research question: \emph{Do the prompts identified by causal-guided optimization actually yield superior task performance compared to existing methods?} Positive results would demonstrate that the causal reward learned in Stage~1 provides actionable guidance that consistently translates into real-world gains. We evaluate CPO against both human-designed and automated baselines on held-out test sets from the same three benchmarks, ensuring no overlap with the offline data used for causal training. This separation guarantees that evaluation reflects genuine generalization rather than memorization.

We compare CPO against three categories of baseline methods. The first category, ordinary human prompts (baseline/control prompt in our setting), represents standard manually crafted instructions that serve as a baseline for typical prompt engineering practice. The second category, conventional prompting approaches, includes Chain-of-Thought (CoT)~\citep{wei2022chain} and Reflexion~\citep{shinn2023reflexion}, both of which enhance reasoning by introducing explicit intermediate steps or iterative self-reflection mechanisms. The third category comprises automated prompt optimization methods, such as APE~\citep{zhou2022large}, OPRO~\citep{yang2023large}, PromptAgent~\citep{wangpromptagent}, PromptBreeder~\citep{fernando2024promptbreeder}, TextGrad~\citep{yuksekgonul2025optimizing}, and DSPy~\citep{khattab2023dspy}. These methods seek to identify a single optimized prompt that generalizes across an entire dataset. Compared with these automated prompt optimization techniques, CPO conditions on both query and prompt representations to generate query-specific optimized prompts. This design enables fine-grained adaptation to query heterogeneity and supports superior generalization across diverse input distributions. The implementation details of these baselines, including optimization hyperparameters and final optimized prompts, are provided in Appendix~\ref{sec:baseline_configuration}.

\subsection{Experimental Implementation}

\subsubsection{LLM Model Configuration}
This part specifies the model configuration and experimental environment used across all benchmarks.
We employ Qwen2.5-14B as both the task-execution model ($LLM_{\text{task}}$) and the prompt-generation model ($LLM_{\text{prompt}}$). In all experiments, the model temperature is fixed at 0.8, unless otherwise specified in baseline configurations.
For the embedding model, we employ the \texttt{nomic-embed-text-v1.5} model~\citep{nussbaum2025nomic}, a high-performance open embedding model known for its strong semantic alignment across diverse text domains. It produces 768-dimensional embeddings that encode the semantic and structural properties of natural language queries and prompts into continuous vector representations. To enhance efficiency and interpretability in causal estimation, we apply Principal Component Analysis (PCA) to the embeddings, thereby reducing dimensionality and extracting the most salient prompt patterns. Because the semantic complexity of queries and prompts varies across tasks, we set task-specific PCA dimensions. For dimensionality reduction, we retain 40 and 15 PCA components for queries and prompts in MATH, 20 and 10 in VisEval, and 40 and 30 in DABench, respectively.

\subsubsection{Task-Specific Execution Environment}
Because the three benchmark tasks involve different types of model interaction, ranging from direct text generation to code execution and data visualization, we adopt task-specific frameworks for integrating $LLM_{\text{task}}$ into the evaluation pipeline.

(1) For the MATH dataset, the task involves direct query answering and reasoning. We therefore use $LLM_{\text{task}}$ through the standard OpenAI-compatible API provided in the \texttt{openai} Python library, which returns plain-text responses for each query–prompt pair.

(2) For the VisEval dataset, which requires natural-language-to-visualization code generation, we follow the original benchmark setup and adopt the CoML framework~\citep{zhang2024mlcopilot}. $LLM_{\text{task}}$ serves as the LLM within the CoML agent, which structures prompts using the ``Variables–Executed Code–Request'' format: variables describe the dataset, executed code specifies preprocessing steps, and the request expresses the user’s visualization query.

(3) For the DABench dataset, which centers on data-analysis tasks requiring code execution and structured output, we employ the \texttt{smolagents} framework~\citep{smolagents} developed by HuggingFace. $LLM_{\text{task}}$ acts as the agent's reasoning module, capable of accessing tabular data in CSV format to autonomously perform code generation, file reading, execution, result extraction, and answer evaluation, enabling reproducible and end-to-end testing across tasks.

\subsubsection{Configuration of DML Model in Stage 1}
In Stage~1, we implement causal reward learning using the \textit{CausalForestDML} estimator, which decomposes estimation into two nuisance models: the outcome model $m(\mathbf{x})$ and the treatment model $e(\mathbf{x})$.
The outcome model $m(\mathbf{x})$ is a \textit{GradientBoostingClassifier} configured with 100 estimators, a maximum tree depth of 3, and a minimum leaf size of 20.\footnote{For the DABench dataset, the number of estimators is reduced to 50 to control memory usage and prevent kernel crashes.}
The treatment model $e(\mathbf{x})$ is implemented as a \textit{MultiOutputRegressor} wrapping a \textit{GradientBoostingRegressor} with identical hyperparameters.
Unless otherwise specified, we use a 0.9/0.1 split for training and held-out evaluation when fitting the DML estimator.

\subsubsection{Configuration of Prompt Optimization in Stage 2}

In Stage~2, prompt optimization proceeds through the iterative causal-guided search described in Section~\ref{subsec:stage2}. 
Using the same notation as before ($R$ for optimization rounds, $K$ for the number of top-performing prompts retained per round, and $B$ for the number of new candidates generated from each prompt), we set $R = 3$, $K = 3$, and $B = 5$, resulting in $35$ newly produced prompt variations overall. 
These hyperparameters strike a balance between exploration and refinement, enabling efficient traversal of the discrete prompt space without excessive computational cost.

\subsection{Performance of Causal Reward Learning}

Following the validation design described in Section~\ref{subsubsec:stage1-validate}, we assess model performance on unseen queries or prompts. 
For both the causal estimator in CPO and the non-causal ML baseline introduced in Section~\ref{subsubsec:stage1-validate}, we report Kendall’s tau-b scores for all three tasks, as shown in Table~\ref{tab:ktaub-results}.

Across all three benchmarks, the causal reward model in CPO consistently achieves higher Kendall’s tau-b scores than the non-causal ML baseline, indicating that causal reward learning provides a more reliable and generalizable estimation of prompt effects. 
Specifically, the causal estimator of CPO improves Kendall’s tau-b from 0.0441 to 0.0608 (+38\%) on MATH, from 0.0980 to 0.1283 (+31\%) on VisEval, and from 0.1347 to 0.1509 (+12\%) on DABench. 
These results provide quantitative evidence that the proposed causal reward model captures the true effects of prompt variation more reliably than a purely predictive approach, reinforcing the value of causal inference as the foundation for prompt optimization.

\begin{table}[!ht]
    \centering
    \caption{Performance Comparison of Causal and Non-Causal Estimators on Unseen Query/Prompt}
    \label{tab:ktaub-results}
    \tabcolsep=0.029\linewidth
    \small
    \renewcommand{\arraystretch}{1.15}
    \begin{threeparttable}
    \begin{tabular}{
            >{\centering\arraybackslash}m{1.6cm}
            >{\centering\arraybackslash}m{3.6cm}
            >{\centering\arraybackslash}m{2.5cm}
        }
        \toprule
        Task & Method & Kendall's Tau-b \\
        \midrule
        \multirow{2}{*}{MATH} 
            & Non-causal ML baseline & 0.0441 \\
            & Causal estimator of CPO & \textbf{0.0608} \\ 
        \hline
        \multirow{2}{*}{VisEval} 
            & Non-causal ML baseline & 0.0980 \\
            & Causal estimator of CPO & \textbf{0.1283} \\ 
        \hline
        \multirow{2}{*}{DABench} 
            & Non-causal ML baseline & 0.1347 \\
            & Causal estimator of CPO & \textbf{0.1509} \\
        \bottomrule
    \end{tabular}
    \end{threeparttable}
\end{table}

\subsection{Performance of Causal-Guided Optimization}
In Stage~2, we design dedicated test datasets for each task to rigorously evaluate the effectiveness of CPO in guiding prompt optimization on previously unseen queries. All test queries are sampled to ensure no overlap with the offline dataset used for causal reward learning, providing a robust basis for assessing generalization. Because each benchmark categorizes queries by difficulty, we construct stratified test sets by sampling across multiple difficulty levels, with an emphasis on higher-difficulty categories to better reflect challenging real-world scenarios. The detailed composition of each test set is summarized in Table~\ref{tab:testset-construction}.

\begin{table}[!ht]
    \centering
    \caption{Construction Details of Test Sets for Each Task in Stage 2}
    \label{tab:testset-construction}
    \small
    \renewcommand{\arraystretch}{1.15}
    \begin{threeparttable}
    \begin{tabular}{l l p{0.5\linewidth} c}
        \toprule
        Task & Difficulty Levels & Selection Criteria & Size \\
       \midrule
        MATH & Levels 3, Level 4, Level 5 & Randomly sample 100 queries from each level & 300 \\
        VisEval & Easy, Medium, Hard, Extra Hard & Sample 100 \(\langle\)NL, VIS\(\rangle\) pairs per level with each VIS unique & 400 \\
        DABench & Easy, Medium, Hard & Randomly sample 30 queries from each level & 90 \\
 \bottomrule 
    \end{tabular}
    \end{threeparttable}
\end{table}

Tables~\ref{tab:math-prompt-optimization}-\ref{tab:dabench-prompt-optimization} report the prompt optimization performance on the MATH, VisEval and DABench test sets respectively, detailing accuracy across each difficulty level as well as the overall performance. 
As shown in Table~\ref{tab:math-prompt-optimization}, CPO exhibits two notable advantages on the MATH benchmark. First, it maintains strong and consistent performance across all three difficulty levels, achieving the highest overall accuracy of 90.00\%. In contrast, while several automated methods (e.g., APE and PromptBreeder) perform well on individual levels, their results fluctuate across tasks, reflecting less stable generalization. This consistency highlights CPO’s robustness and adaptability in mathematical reasoning tasks. Second, CPO attains the top accuracy (82\%) on Level~5, the most challenging subset, outperforming all baselines except APE, with which it is tied. This result underscores CPO’s effectiveness in handling complex reasoning problems that demand more advanced prompt engineering.

\begin{table*}[!ht]
    \centering
    \caption{Prompt Optimization Results on MATH}
    \label{tab:math-prompt-optimization}
    \tabcolsep=0.029\linewidth
    \small
    \renewcommand{\arraystretch}{1.15}
    \begin{threeparttable}
    \begin{tabular}{l l c c c c}
        \toprule
        \multirow{2}{*}{Type} & \multirow{2}{*}{Method} & \multicolumn{3}{c }{Accuracy (\%) by Difficulty Level} & \multirow{2}{*}{Overall} \\  
        \cline{3-5}
        & & Level 3 & Level 4 & Level 5 & \\
        \midrule
        Human Prompts & Human & \underline{95} & 91 & 79 & 88.33 \\
        \hline
        \multirow{2}{*}{Conventional Prompting} 
            & CoT (1-shot) & 93 & 92 & 74 & 86.33 \\
            & Reflexion & 90 & 92 & 77 & 86.33 \\
        \hline
        \multirow{6}{*}{Automated Prompting} 
            & APE & 94 & 92 & \textbf{82} & \underline{89.33} \\
            & OPRO & 94 & 92 & 79 & 88.33 \\
            & PromptAgent & \underline{95} & \underline{93} & 77 & 88.33 \\
            & PromptBreeder & 92 & \textbf{94} & \underline{80} & 88.67 \\
            & TextGrad & 92 & 87 & 79 & 86.00 \\
            & DSPy & 78 & 75 & 62 & 71.67 \\
        \rowcolor{gray!30} Ours & \textbf{CPO} & \textbf{96} & 92 & \textbf{82} & \textbf{90.00} \\
        \bottomrule
    \end{tabular}
    \tabnote{\textit{Note:} For each column, the best performance is highlighted in bold and the second best one is highlighted in underline.}
    \end{threeparttable}
\end{table*}

In Table~\ref{tab:viseval-prompt-optimization}, we can see that CPO achieves the highest overall accuracy of 54.75\% on the VisEval test set, outperforming all other baselines with a clear margin. This result underscores the robustness of our framework, as CPO consistently delivers top-tier performance across all difficulty levels, whereas other methods exhibit substantial variation. For instance, PromptAgent attains the best accuracy (77\%) on the \textit{Easy} subset but performs poorly on the remaining three levels. Similarly, while PromptBreeder ranks second on the \textit{Extra Hard} subset, its performance on the \textit{Medium} and \textit{Hard} subsets lags considerably. In contrast, CPO maintains balanced and stable accuracy across all difficulty levels, a property critical for real-world deployment where query complexity can vary widely.

\begin{table*}[!ht]
    \centering
    \caption{Prompt Optimization Results on VisEval}
    \label{tab:viseval-prompt-optimization}
    \tabcolsep=0.02\linewidth
    \small
    \renewcommand{\arraystretch}{1.15} 
    \begin{threeparttable}
    \begin{tabular}{l l c c c c c}
        \toprule
        \multirow{2}{*}{Type} & \multirow{2}{*}{Method} & \multicolumn{4}{c}{Accuracy (\%) by Difficulty Level} & \multirow{2}{*}{Overall} \\  
        \cline{3-6}
        & & Easy & Medium & Hard & Extra Hard & \\
        \midrule
        Human Prompts & Human & 74 & 59 & 42 & 26 & 50.25 \\
        \hline
        \multirow{2}{*}{Conventional Prompting} 
            & CoT (1-shot) & 73 & 58 & 44 & 27 & 50.50 \\
            & Reflexion & 73 & 54 & 43 & 29 & 49.75 \\
        \hline
        \multirow{6}{*}{Automated Prompting} 
            & APE & 75 & \underline{59} & 43 & \textbf{36} & \underline{53.25} \\
            & OPRO & 74 & 57 & \underline{48} & 34 & \underline{53.25} \\
            & PromptAgent & \textbf{77} & 50 & 39 & 26 & 48.00 \\
            & PromptBreeder & \underline{76} & 51 & 42 & \underline{35} & 51.00 \\
            & TextGrad & 74 & 59 & 42 & 26 & 51.25 \\
            & DSPy & 58 & 43 & 39 & 13 & 38.25 \\
        \rowcolor{gray!30} Ours & \textbf{CPO (ours)} & \underline{76} & \textbf{60} & \textbf{49} & 34 & \textbf{54.75} \\
        \bottomrule
    \end{tabular}
    \tabnote{\textit{Note:} For each column, the best performance is highlighted in bold and the second best one is highlighted in underline.}
    \end{threeparttable}
    
\end{table*}

On DABench, as shown in Table~\ref{tab:dabench-prompt-optimization}, CPO achieves the highest overall accuracy of 65.33\%, outperforming all baseline methods, including the strongest automated competitor, PromptBreeder (62.33\%). Beyond overall performance, CPO demonstrates remarkable consistency across difficulty levels, achieving 77\% on the \textit{Easy} subset and 69\% on the \textit{Medium} subset. This stability contrasts sharply with methods such as APE and PromptBreeder, whose performance fluctuates more substantially across tasks. Moreover, CPO achieves a clear advantage on the \textit{Hard} subset, attaining 50\% accuracy, substantially higher than any other method. This result highlights CPO’s effectiveness in handling complex reasoning tasks that challenge both conventional and automated prompting approaches. Collectively, these findings underscore CPO’s robustness, stability, and strong generalization in automated prompt optimization for data analysis tasks.

\begin{table*}[!ht]
    \centering
    \caption{Prompt Optimization Results on DABench}
    \label{tab:dabench-prompt-optimization}
    \tabcolsep=0.029\linewidth
    \small
    \renewcommand{\arraystretch}{1.14} 
    \begin{threeparttable}
    \begin{tabular}{l l c c c c}
        \toprule
        \multirow{2}{*}{Type} & \multirow{2}{*}{Method} & \multicolumn{3}{c }{Accuracy (\%) by Difficulty Level} & \multirow{2}{*}{Overall} \\  
        \cline{3-5}
        & & Easy & Medium & Hard & \\
        \midrule
        Human Prompts & Human & \underline{77} & 57 & 28 & 54.00 \\
        \hline
        \multirow{2}{*}{Conventional Prompting} 
        & CoT (1-shot) & 57 & 56 & 30 & 47.67 \\
            & Reflexion & 73 & 50 & 28 & 50.33 \\
        \hline
        \multirow{6}{*}{Automated Prompting} 
            & APE & \underline{77} & 65 & 39  & 60.33 \\
            & OPRO & 58 & 56 & 37 & 50.33 \\
            & PromptAgent & \underline{77} & 65 & 42 & 61.33 \\
            & PromptBreeder & \textbf{80}  & \textbf{72} & 35 &  \underline{62.33}\\
            & TextGrad & 73 & 60 & 25 & 52.67 \\
            & DSPy & 70 & 59 & 37 & 55.33 \\
                \rowcolor{gray!30} Ours & \textbf{CPO} & \underline{77} &  \underline{69} & \textbf{50} & \textbf{65.33} \\
        \bottomrule
    \end{tabular}
    \tabnote{\textit{Note:} For each column, the best performance is highlighted in bold and the second best one is highlighted in underline.}
    \end{threeparttable}
\end{table*}

\section{Why and Where Does CPO Work?}
\label{sec:analysis}

\subsection{The Impact of Causal Modeling}
\begin{figure}[h!]
    \centering
    \begin{subfigure}[b]{0.32\linewidth}
        \includegraphics[width=\linewidth]{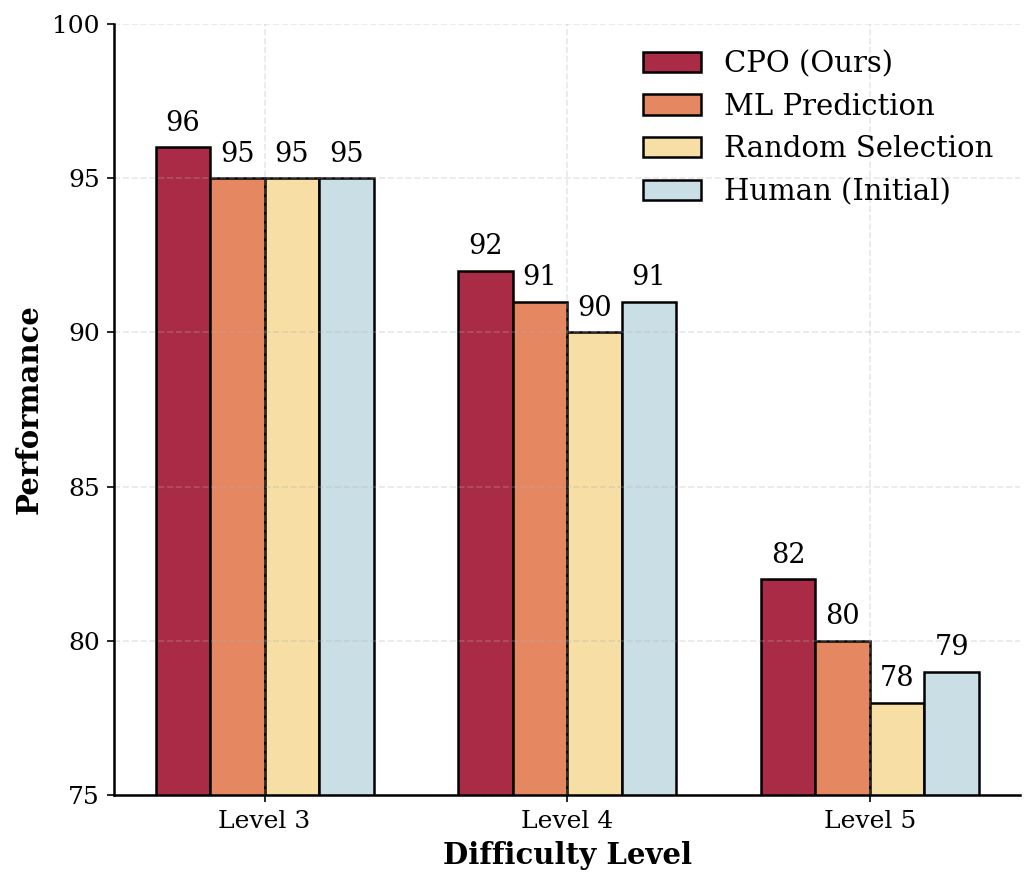}
        \caption{MATH}
        \label{fig:prompt-selection-a}
    \end{subfigure}
    \begin{subfigure}[b]{0.31\linewidth}
        \includegraphics[width=\linewidth]{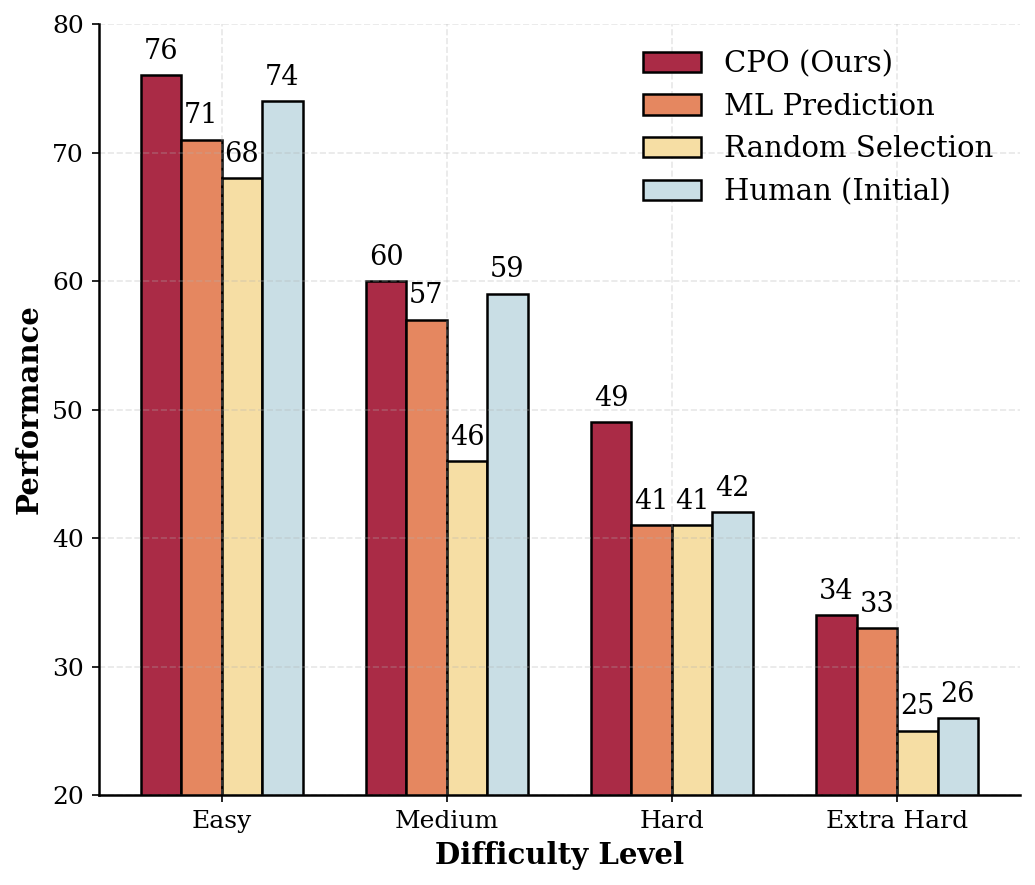}
        \caption{VisEval}
        \label{fig:prompt-selection-b}
    \end{subfigure}
    \begin{subfigure}[b]{0.31\linewidth}
        \includegraphics[width=\linewidth]{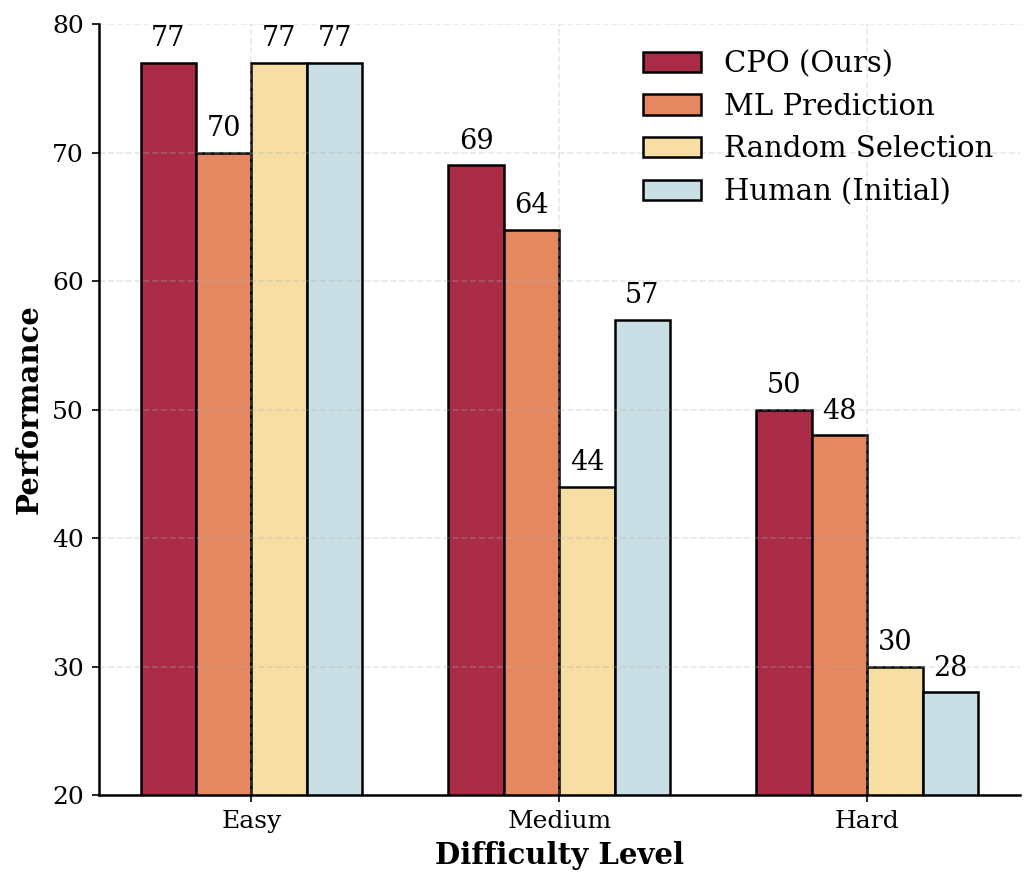}
        \caption{DABench}
        \label{fig:prompt-selection-c}
    \end{subfigure}
    \caption{Ablation Study for the Causal Reward Model in CPO.}
    \label{fig:ablation}
\end{figure}

To rigorously examine whether the performance gains of CPO stem from genuine causal identification, we conduct a component-level ablation study comparing our full CPO framework against two variants and the human baseline across benchmarks. Figure~\ref{fig:ablation} illustrates the performance comparison of four settings: (1) CPO, our framework employing DML-based causal reward learning to estimate the $\hat{\tau}(x,t)$ (the CATE) of prompt candidates; (2) ML prediction variant, which replaces the causal reward model with machine learning prediction model while keeping all other components identical; (3) Random selection variant, which further removes the reward-based prompt selection mechanism and randomly selects prompts from the pool of LLM-generated candidates. Notably, since we let the LLM refine the previous prompts during the prompt generation process, this variant can represent the performance based purely on the ``self-refinement" capability of $LLM_{\text{prompt}}$; (4) Human (initial), which is the unoptimized control prompt.

The ablation results reveal several important insights about the contribution of causal modeling. \textit{First}, CPO demonstrates superior performance compared to the other variants, as well as the human baseline across all difficulty levels. The comparison between CPO and ML prediction is the most theoretically significant. ML prediction estimates the expected model performance for each query-prompt combination, which may be confounded by the query-specific characteristics, such as varying levels of difficulty. In contrast, CPO aims to isolate the causal effect attributable solely to prompt modifications from confounding factors. Since both methods share the same architecture and search strategy, the performance gap of the two approaches (e.g., CPO achieving 82\% vs. ML's 80\% on MATH Level 5) is attributable strictly to the prompt selection preference guided by the reward model. This suggests that the causal framework's ability to isolate the true effect of prompt variations from confounding factors is crucial for effective prompt selection. 
\textit{Second}, CPO exhibits more stable performance across queries from different difficulty levels compared to other methods. Such stability further confirms the superiority of causal reward learning for handling unseen queries and prompts at test time. 
\textit{Third}, the results show that the Random selection baseline not only fails to consistently improve upon the initial human prompt, but in many cases actually underperforms it (for example, achieving only 44\% versus 57\% on DABench Hard). This phenomenon underscores an important limitation of relying solely on the self-refinement capabilities of LLMs: new prompts generated in this way are not guaranteed to be better, and may even introduce significant variance and instability into performance. These findings highlight the necessity of employing a robust causal filter to effectively identify genuinely beneficial prompt modifications from the vast and noisy pool of LLM-generated prompt candidates.

\subsection{Scalability and Robustness to Training Data Size}

The benefits of causal modeling become more pronounced as the size of the offline dataset increases. As data accumulates, a causal learner can increasingly disentangle true prompt effects from confounding factors, whereas a correlational learner may instead overfit to spurious associations.

\begin{figure*}[!ht]
    \centering
    \begin{subfigure}[b]{0.48\textwidth}
        \centering
        \includegraphics[width=\linewidth]{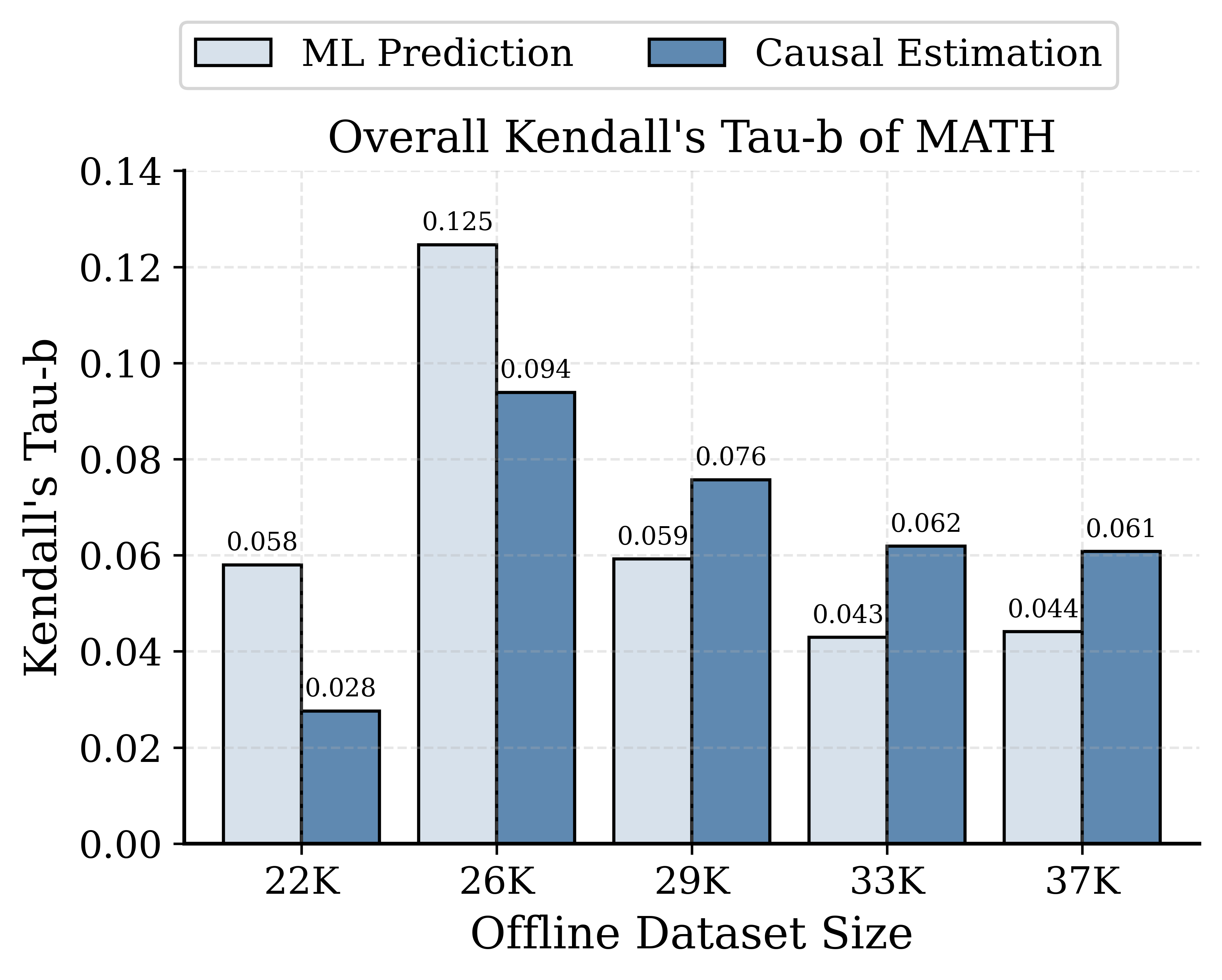}
        \caption{Stage 1: Causal Reward Learning}
        \label{fig:ktaub_by_question_ratio_math}
    \end{subfigure}
    \quad
    \begin{subfigure}[b]{0.48\textwidth}
        \centering
        \includegraphics[width=\linewidth]{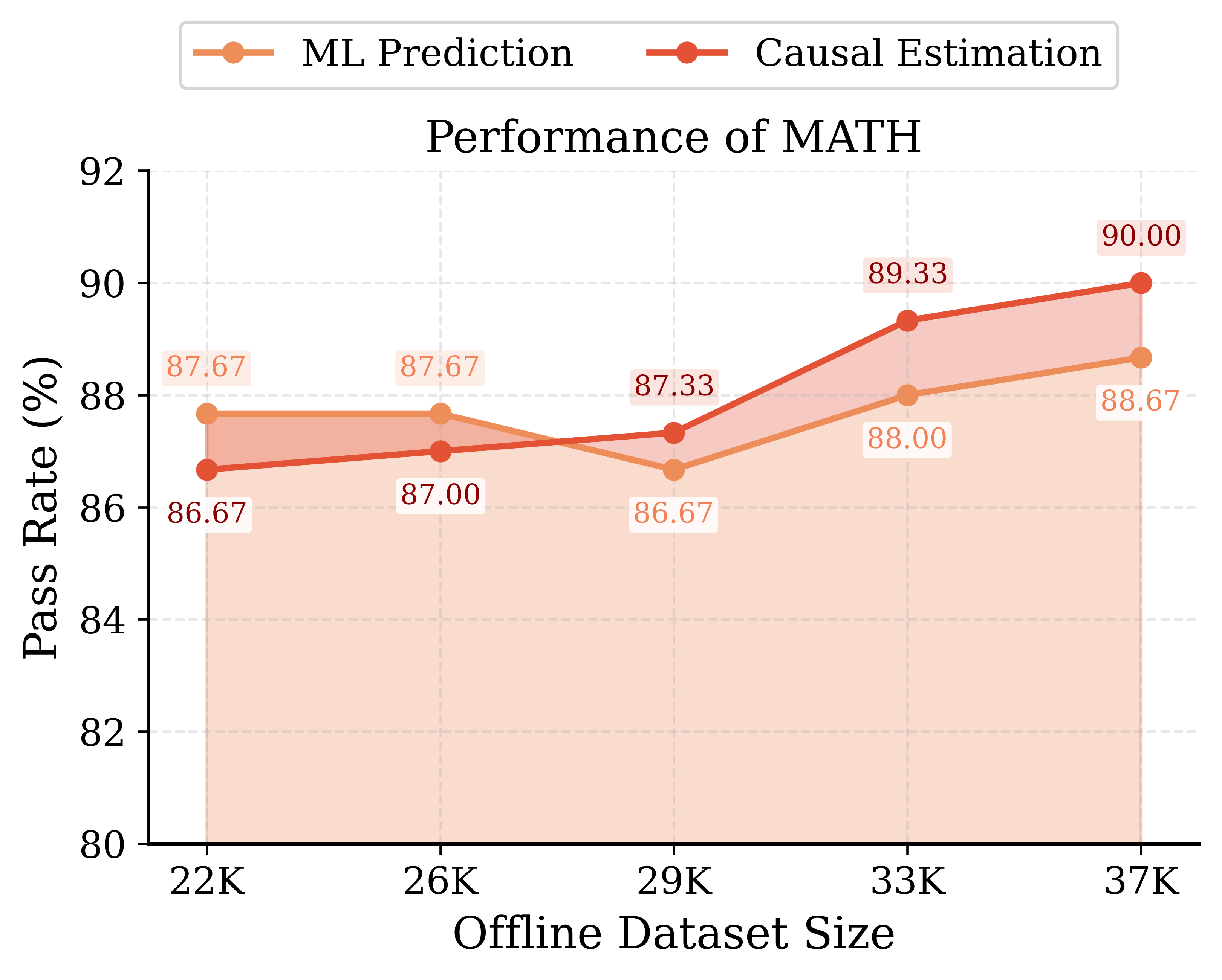}
        \caption{Stage 2: Causal-Guided Optimization}
        \label{fig:performance_by_question_ratio_math}
    \end{subfigure}
    \caption{Performance of CPO as Data Accumulates (Dataset: MATH)}
    \label{fig:ktaub_and_performance_by_question_ratio_math}
\end{figure*}

\begin{figure*}[!ht]
    \centering
    \begin{subfigure}[b]{0.48\textwidth}
        \centering
        \includegraphics[width=\linewidth]{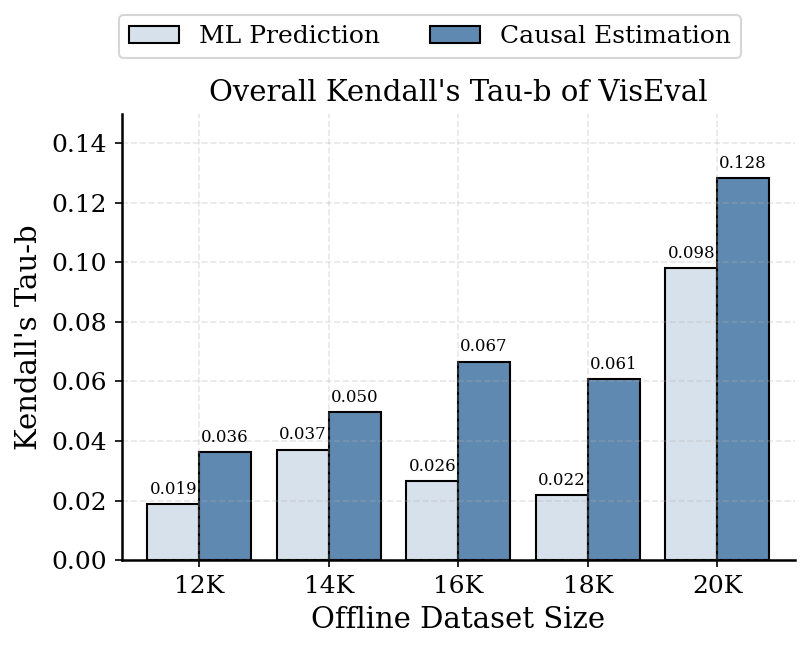}
        \caption{Stage 1: Causal Reward Learning}
        \label{fig:ktaub_by_question_ratio_viseval}
    \end{subfigure}
    \quad
    \begin{subfigure}[b]{0.48\textwidth}
        \centering
        \includegraphics[width=\linewidth]{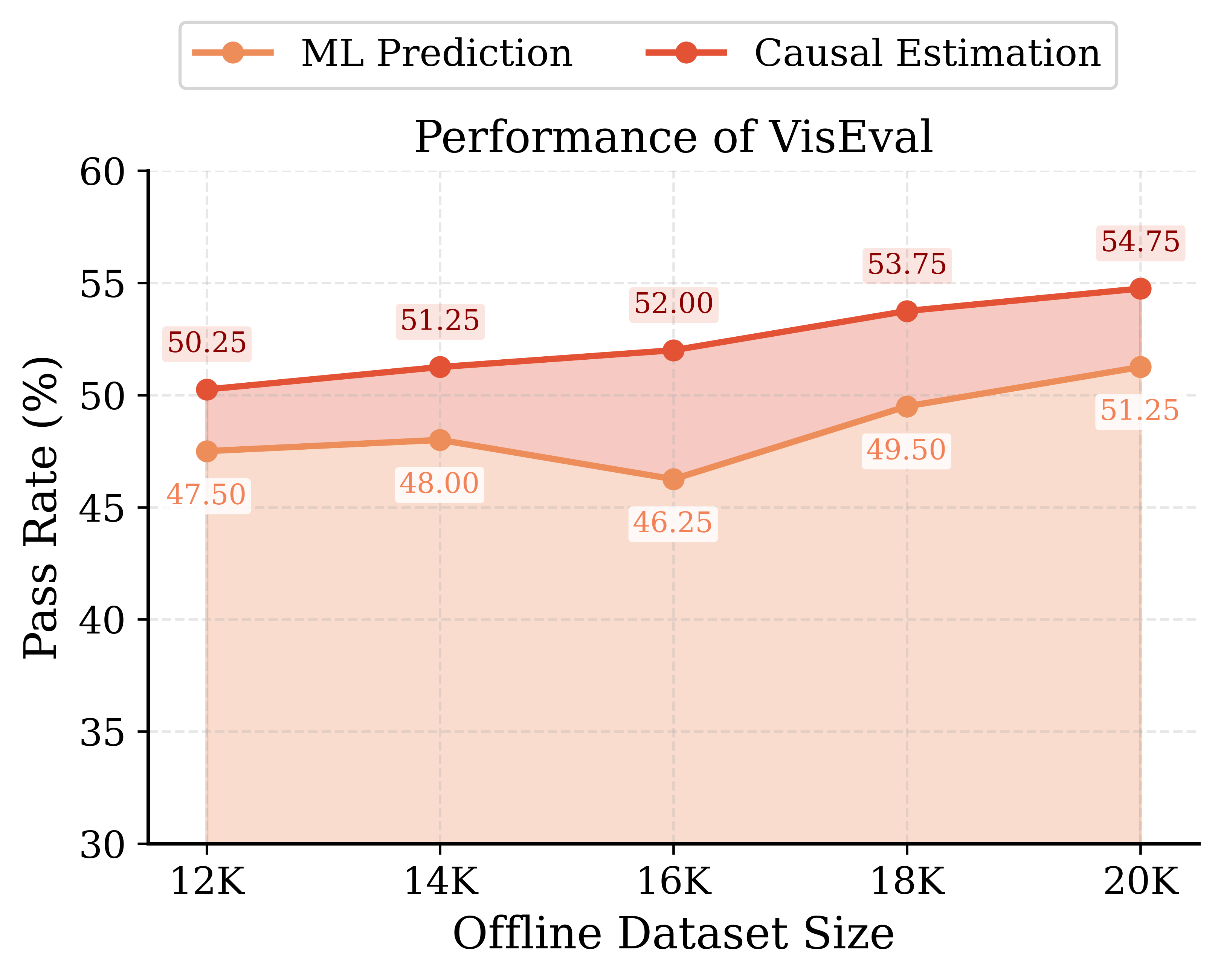}
        \caption{Stage 2: Causal-Guided Optimization}
        \label{fig:performance_by_question_ratio_viseval}
    \end{subfigure}
    \caption{Performance of CPO as Data Accumulates (Dataset: VisEval)}
    \label{fig:ktaub_and_performance_by_question_ratio_viseval}
\end{figure*}

\begin{figure*}[!ht]
    \centering
    \begin{subfigure}[b]{0.48\textwidth}
        \centering
        \includegraphics[width=\linewidth]{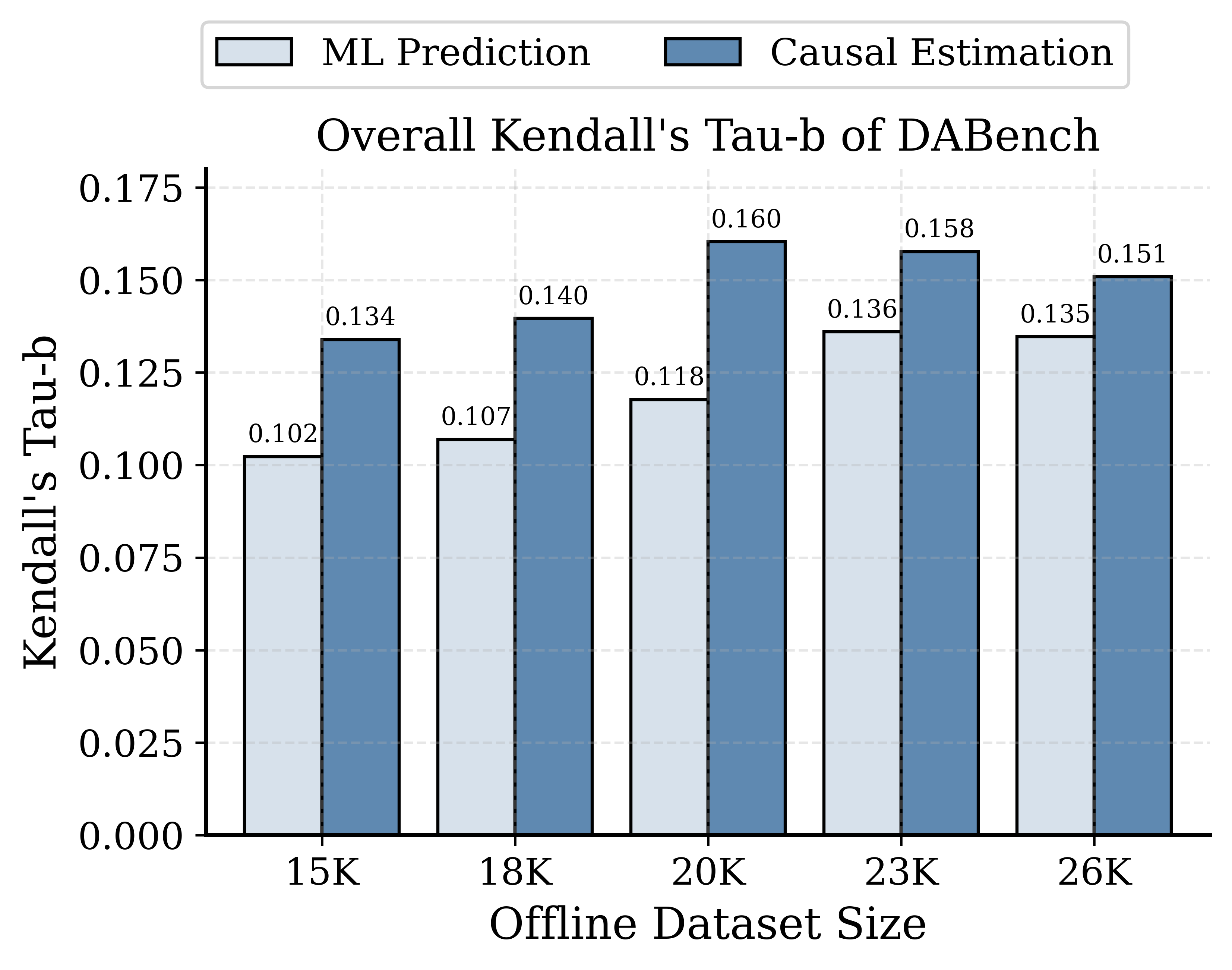}
        \caption{Stage 1: Causal Reward Learning}
        \label{fig:ktaub_by_question_ratio_dabench}
    \end{subfigure}
    \quad
    \begin{subfigure}[b]{0.48\textwidth}
        \centering
        \includegraphics[width=\linewidth]{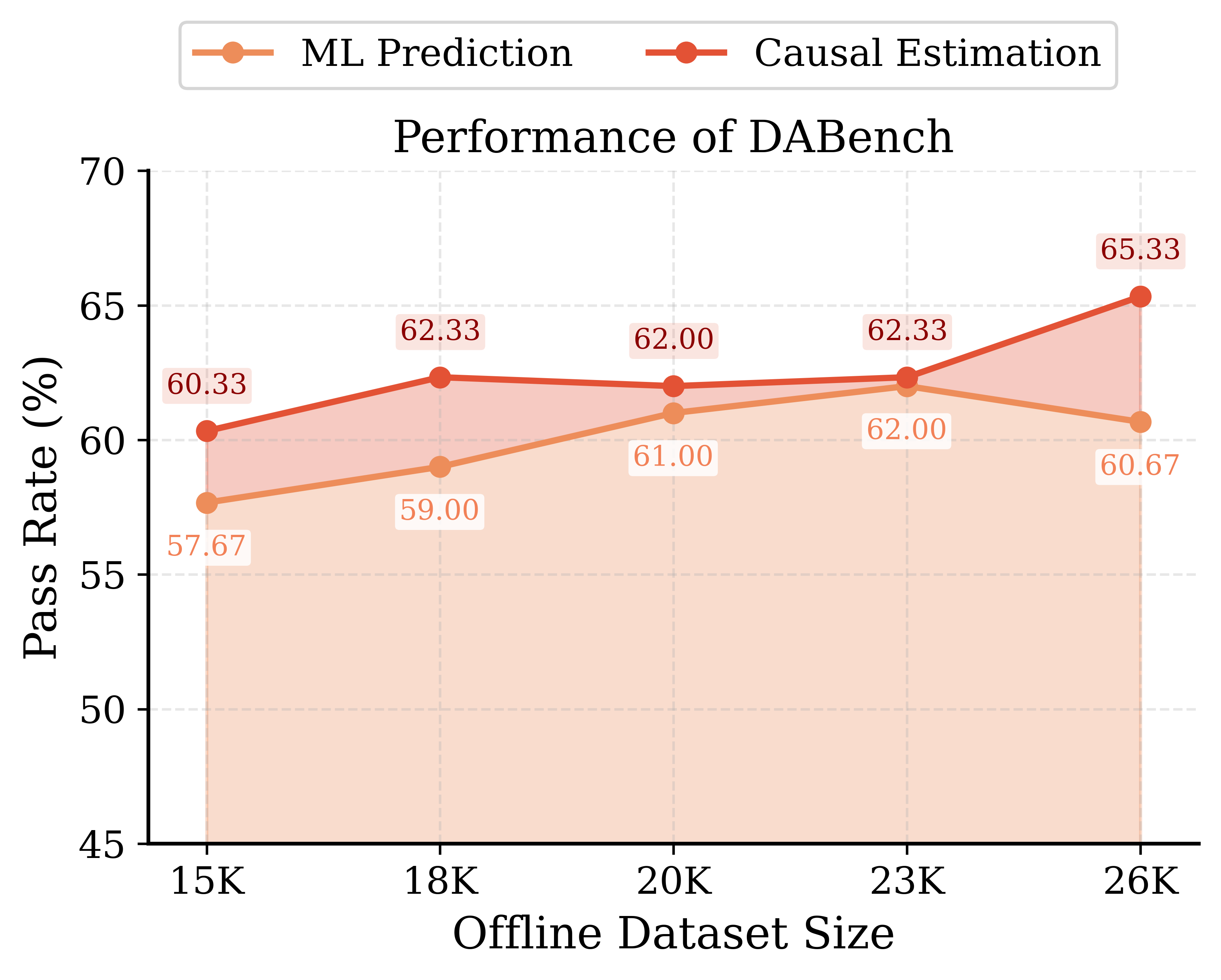}
        \caption{Stage 2: Causal-Guided Optimization}
        \label{fig:performance_by_question_ratio_dabench}
    \end{subfigure}
    \caption{Performance of CPO as Data Accumulates (Dataset: DABench)}
    \label{fig:ktaub_and_performance_by_question_ratio_dabench}
\end{figure*}

Figures~\ref{fig:ktaub_and_performance_by_question_ratio_math} to~\ref{fig:ktaub_and_performance_by_question_ratio_dabench} confirm this hypothesis across all three benchmarks. 
In Stage 1 (Causal Reward Learning), as the offline dataset size increases, the causal reward model of CPO demonstrates a more favorable \emph{data scaling behavior} compared to the non-causal ML model, with performance advantage becoming more pronounced as more data accumulates.
This pattern is most evident on MATH, where CPO initially underperforms the ML baseline at smaller dataset sizes (22K and 26K), but overtakes and stabilizes ahead of the baseline once sufficient data is available (from 29K onward, CPO consistently achieves Kendall's $\tau_b$ around 0.061--0.076, while the baseline drops to 0.043--0.059).
This crossover pattern reveals a critical insight: causal estimators require a certain volume of observational data to effectively disentangle confounding effects, but once this threshold is reached, they deliver more reliable and stable effect estimates than purely predictive models.

Notably, in Stage 2 (Causal-Guided Optimization), a clear contrast emerges across all three benchmarks: CPO performance shows a consistent upward tendency with increasing data volume, while the ML prediction baseline either fluctuates irregularly or undergoes a notable decline. Moreover, the improvements in task performance achieved by CPO directly reflect the gains in causal effect estimation quality observed in Stage 1, whereas for the ML baseline, the trends of Stage 1 and Stage 2 sometimes contradict each other. This further highlights the reliability of CPO when generalizing to new queries and prompts, benefiting from the ability of the causal reward model to disentangle true prompt effects from confounding factors.

Taken together, these findings suggest that for enterprise applications where interaction logs accumulate continuously, CPO offers a superior ``return on data", that is, CPO extracts greater value from each additional observation, converting accumulated historical logs into progressively more accurate causal estimates and, ultimately, into more effective optimization policies.

\subsection{Validating the ``Latent Treatment" Assumption}
\label{sec:interpretability-pca}

Our causal framework relies on the assumption that the continuous PCA components of prompt embeddings ($\mathbf{z} \subset \mathbb{R}^{d_t}$) act as valid ``latent treatments." One potential concern is whether mathematical operations in this reduced space correspond to meaningful semantic interventions. To validate this, we analyze the alignment between the extracted PCA components and human-interpretable prompt design patterns.

We defined 7 distinct prompt patterns (e.g., \emph{Constraint Strictness}, \emph{Chain-of-Thought Guidance}, \emph{Tone}) based on recent taxonomies~\citep{ white2023prompt, geroimenko2025key} (see Table~\ref{sec:prompt-patterns} in Appendix). We then scored the VisEval offline prompts as an example against these patterns using GPT-4o and computed the Pearson correlation between the pattern scores and the PCA coordinates.

\begin{figure*}[!ht]
    \begin{center}
        \includegraphics[width=0.9\textwidth]{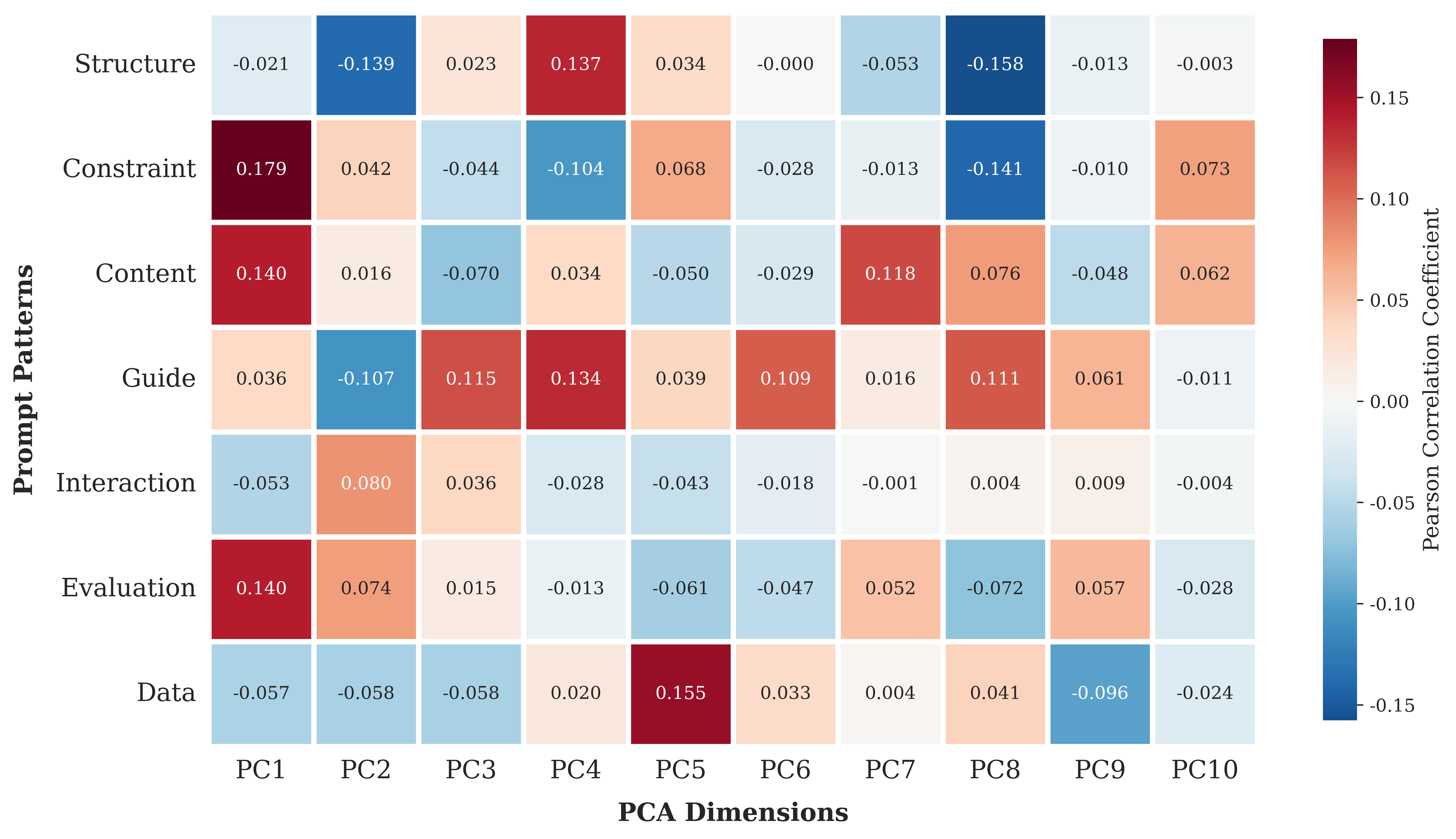}
    \end{center}
    \caption{Correlation Coefficients between PCA Components and Prompt Patterns (Dataset: VisEval)}
    \label{fig:pca-visualization}
\end{figure*}

Figure~\ref{fig:pca-visualization} demonstrates that the leading PCA components serve as \emph{composite latent treatments}. For instance, PC1 correlates strongly with \emph{Constraint} ($\rho=0.179$) and \emph{Content} ($\rho=0.140$), while PC4 captures \emph{Structure} ($\rho=0.137$). While no single dimension maps one-to-one to a linguistic feature, the matrix confirms that variance in the PCA space corresponds to variance in design choices. This validates our methodological choice: by intervening on these orthogonal components, CPO systematically explores the semantic design space rather than navigating random noise.

\subsection{Sensitivity to the Number of PCA Components}
To examine how the performance of the proposed CPO framework is affected by the choice of PCA dimensionality, we conduct a sensitivity analysis reported in Table~\ref{tab:sensitivity-to-dimension}. We vary the number of retained PCA dimensions for query and prompt embeddings, assessing the robustness of CPO to different levels of representation compression.

\begin{table*}[!ht]
    \centering
    \caption{Sensitivity Analysis to the Number of PCA Components (Dataset: VisEval)}
    \label{tab:sensitivity-to-dimension}
    \tabcolsep=0.035\linewidth
    \small
     \renewcommand{\arraystretch}{1.15} 
    \begin{threeparttable}
    \begin{tabular}{c c | c c c c | c}
        \toprule
        \multicolumn{2}{c|}{PCA Dimensions} & \multicolumn{4}{c|}{Accuracy (\%) by Difficulty Level} & \multirow{2}{*}{Overall} \\
        \multicolumn{1}{c}{Query} & \multicolumn{1}{c|}{Prompt} & Easy & Medium & Hard & Extra Hard &  \\
        \midrule
        10 & 5    & 76 & 51 & 37 & 23 & 46.75 \\
        15 & 10   & 74 & 60 & 42 & 33 & 52.25 \\
        20 & 10   & 76 & 60 & 49 & 34 & \textbf{54.75} \\
        25 & 15   & 75 & 58 & 47 & 33 & \underline{53.50} \\
        40 & 20   & 76 & 59 & 46 & 32 & 53.25 \\
        80 & 40   & 72 & 49 & 41 & 26 & 47.25 \\
        \bottomrule
    \end{tabular}
    \tabnote{\textit{Note:} For each column, the best performance is highlighted in bold and the second best one is highlighted in underline.}
    \end{threeparttable}
\end{table*}

The results reveal an inverted U-shaped relationship. Performance degrades at high dimensions (e.g., $d_x=80$), likely due to the \emph{positivity violation} in causal inference: as dimensionality grows, the probability of observing treated and control units in the same region of the covariate space diminishes, making propensity estimation unstable~\citep{chernozhukov2018double}. This confirms that PCA is not merely a computational convenience but a necessary regularization step to ensure valid causal identification in the semantic space.

\subsection{Economic Analysis: Breaking the Static-Dynamic Trade-off}
\label{sec:cost_analysis}

An operational comparison must distinguish between two distinct deployment strategies: static approaches (one prompt for all queries) and dynamic approaches (unique prompts for each query). Existing baselines (e.g., APE, OPRO) typically employ a \emph{static approach}, which usually obtain an optimized prompt on the training data and then freeze it for incoming queries. While this approach avoids high marginal costs, it incurs a significant performance opportunity cost: the static prompt fails to adapt to the ``long tail" of heterogeneous queries that deviate from the average training distribution. CPO belongs to the dynamic approach category, which adapts the prompt to every specific user query in real-time. Ideally, an enterprise would prefer a \emph{dynamic approach} ($t^*(x)$) that adapts the prompt to every specific user query in real-time. However, static approaches are fundamentally incapable of achieving query adaptation due to their inherent optimization mechanism.

CPO makes the dynamic approach both operationally feasible and highly cost-efficient. Table~\ref{tab:cost_analysis} shows the cost structure of CPO. By offloading the heavy evaluation burden to the offline causal model (Fixed Cost), CPO incurs only a lightweight marginal cost for generating prompt candidates during real-time adaptation to each query. For example, under the default parameters in this study, starting from a single initial seed prompt, only 7 calls to the $LLM_{\text{prompt}}$ can generate 35 prompt candidates. This structural shift not only makes dynamic adaptation feasible in practice, but also ensures that per-query customization is achieved at a manageable and economic cost, enabling organizations to fully capitalize on the benefits of real-time personalized prompt optimization without incurring prohibitive overhead.

\begin{table}[ht]
    \centering
    \caption{CPO Cost Structure: Fixed Investment vs. Marginal Execution}
    \label{tab:cost_analysis}
    \small
    \tabcolsep=0.035\linewidth
    \renewcommand{\arraystretch}{1.15} 
    \begin{tabular}{l c c c}
        \toprule
        \textbf{Dataset} & \multicolumn{2}{c}{\textbf{Fixed Investment (Stage 1)}} & \textbf{Marginal Cost (Stage 2)} \\
        & \multicolumn{2}{c}{($LLM_{\text{task}}$ calls - \textit{One-time Sunk Cost})} & ($LLM_{\text{prompt}}$ calls - \textit{Per Query}) \\
        \midrule
        MATH     & \multicolumn{2}{c}{37,185} & 7 (by default)\\
        VisEval  & \multicolumn{2}{c}{20,502} & 7 (by default)\\
        DABench  & \multicolumn{2}{c}{26,240} & 7 (by default)\\
        \bottomrule
    \end{tabular}
    \begin{tablenotes}
        \footnotesize
        \item \textit{Note:} The table summarizes the CPO framework's two-stage cost structure. Fixed investment reflects the one-time, offline evaluation cost using the $LLM_{\text{task}}$, usually amortized as a sunk cost in organizations with historical data. The marginal cost is the per-query online cost for $LLM_{\text{prompt}}$, which is lightweight. This separation enables practical, economic per-query dynamic prompt optimization at scale.
    \end{tablenotes}
\end{table}

By comparing the baseline methods using static approaches versus CPO from a long-term perspective in real-world enterprise applications, CPO demonstrates significant advantages in both cost efficiency and the stability of returns.
Taking the DABench dataset as an example, the baseline methods using static approaches require approximately 5,000 to 10,000 LLM calls for a single optimization, which stems from the need to evaluate the performance of prompt candidates on the training dataset. For only a single optimization, static approaches appear to be less expensive than the fixed investment cost for DABench shown in Table~\ref{tab:cost_analysis}. However, in practical applications, the enterprises will continuously accumulate data through ongoing operations and periodically perform prompt optimization on the ever-growing dataset in pursuit of greater performance and returns. In this situation, the cost of static approaches on each optimization escalates significantly as data volume grows, because the prompt candidates need to be evaluated on this ever-larger dataset. In contrast, CPO offers a substantial advantage in such continuous, long-term optimization: when retraining the causal model on larger datasets, it relies solely on DML algorithm computations and incurs no additional LLM invocation costs, ensuring consistently low operational expenses beyond the initial sunk cost. Therefore, from a long-term perspective, the total cost of static approaches will far exceed that of CPO. Moreover, CPO is better positioned to reap greater benefits from accumulating data: as the dataset grows, the causal model becomes more accurate, which in turn enhances the performance of subsequent query-adaptive prompt optimization and enables steady improvement in business returns. In contrast, static-policy methods, due to their trial-and-error nature, still cannot guarantee that prompts optimized on historical data will generalize well to new queries. This combination of escalating costs and unstable returns makes static-policy methods undesirable for enterprises seeking reliable and scalable optimization.

Furthermore, CPO also demonstrates superior cost efficiency compared to existing dynamic-policy approaches. It is worth noting that recent dynamic-policy approaches achieve query adaptation by fine-tuning a pretrained large language model into a ``prompt model'' that maps inputs directly to prompts. While effective in query-level prompt adaptation, these approaches are highly demanding in technical and resource requirements (including technical staff, large-scale training data, GPU clusters, and so on), constraining their practicality for many organizations, particularly small-scale and non-technical enterprises. By contrast, CPO achieves dynamic adaptation with a markedly lower adoption barrier and superior cost efficiency: it leverages off-the-shelf LLMs together with an offline causal model, avoids model fine-tuning, and confines online expenditure to lightweight prompt generation.

\section{Concluding Remarks}
\label{sec:conclusion}
Prompt design has emerged as the primary interface through which organizations control LLM behavior, yet it remains an ad-hoc practice: labor-intensive, brittle, and prohibitively costly to scale. Existing automated approaches offer limited relief: static methods optimize for average performance while sacrificing adaptation to query heterogeneity, and dynamic methods rely on correlational reward signals that confound prompt effectiveness with query characteristics. This paper introduces Causal Prompt Optimization, a framework that reconceptualizes prompt design as a problem of causal inference, enabling systematic identification of prompts that generalize reliably across heterogeneous queries.

Our research advances the literature on AI system design in three ways. First, we expose a foundational limitation in correlation-based optimization: standard reward models systematically misattribute performance gains to prompts when outcomes actually reflect intrinsic query difficulty. By formalizing this confounding bias, we explain why correlation-driven optimization deteriorates under distribution shift. Second, we operationalize causal inference over semantic spaces through a novel integration of principal component analysis and Double Machine Learning. This extends causal machine learning, traditionally applied to structured managerial interventions, to the domain of natural language processing, demonstrating that continuous semantic representations can serve as valid treatments for causal estimation of prompt effects. Third, we resolve a key economic constraint that has limited practical prompt optimization. By converting evaluation from expensive online LLM calls to an offline causal model, CPO enables query-adaptive prompt selection at marginal costs that make dynamic optimization operationally feasible at scale.

Empirically, CPO achieves top overall accuracy across all three benchmarks, mathematical reasoning, visualization, and data analytics, while exhibiting two operational advantages critical for enterprise deployments. First, it maintains substantially higher stability on difficult queries where correlational methods deteriorate: on MATH Level 5, CPO achieves 82\% accuracy compared to 77–80\% for the strongest baselines; on DABench Hard, CPO attains 50\% versus 25–42\% for competitors. This robustness directly addresses the long-tail challenge in enterprise settings, , where rare or complex queries often carry disproportionate economic value. Second, CPO demonstrates favorable data-scaling properties: as offline datasets grow, the causal reward model exhibits consistent improvements in ranking accuracy and optimization performance, while predictive baselines fluctuate or decline, confirming that causal estimation extracts enduring value from accumulated interaction logs.

These results carry direct managerial implications. Existing APO baselines typically adopt a static deployment policy, optimizing a single prompt on training data and  freezing it for all future queries. Under this design, query adaptation is not merely costly but structurally infeasible. CPO breaks this constraint through a two-stage architecture that treats optimization as a capital investment rather than a recurring operational expense. Stage 1 leverages accumulated interaction logs, data organizations already possess, to train an offline causal reward model, eliminating the need for continuous online evaluation. Stage 2 enables query-adaptive prompting at marginal costs comparable to static approaches, requiring only lightweight prompt-generation calls rather than thousands of task-execution queries. This design makes per-query customization economically viable, unlocking value from heterogeneous query distributions that static policies systematically overlook. Moreover, as enterprises accumulate data, the causal model improves without incurring additional LLM costs, converting historical logs into progressively stronger optimized prompts.

Several limitations suggest promising directions for future research. First, while CPO is designed for heterogeneous queries, operational environments can be nonstationary; extending causal reward learning to settings with continuous distribution drift and periodic retraining policies is an important next step. Second, our semantic treatment representation relies on embedding and dimensionality reduction choices; future work can explore alternative representations and causal estimators that improve robustness under weaker overlap or more complex prompt spaces. Third, many enterprise applications are increasingly multimodal (e.g., vision-language agents) and interactive (tool-using agents); adapting causal prompt optimization to these settings would extend the framework to a broader class of AI-enabled workflows and decision processes.

\makeatletter
\@ifundefined{appendix}{\def\appendix{\par\setcounter{section}{0}\setcounter{subsection}{0}\setcounter{subsubsection}{0}%
\renewcommand\thesection{\Alph{section}}\renewcommand\thesubsection{\Alph{section}.\arabic{subsection}}%
\renewcommand\thesubsubsection{\Alph{section}.\arabic{subsection}.\arabic{subsubsection}}}}{}
\makeatother
\appendix
\ECSwitch
\ECHead{Appendix -- ``Optimizing Prompts for Large Language Models: A Causal Approach''}
\input{appendix_body}

\bibliographystyle{informs2014}
\bibliography{references-short,references}

\end{document}

%% file: appendix_body.tex
\label{sec:notation}
Table~\ref{tab:notation} summarizes the major notation used in the paper.

\begin{table}[!htbp]
	\centering
	\small
	\caption{Summary of Notation}
	\renewcommand{\arraystretch}{1.3}
	\begin{tabular}{p{0.18\textwidth} p{0.82\textwidth}}
		\hline
		\textbf{Notation} & \textbf{Description} \\
		\hline
		\multicolumn{2}{l}{\textbf{Setting}} \\

        $\mathcal{X},\mathcal{T}$ & Space of query texts and candidate prompt texts \\
		$LLM_{\text{task}}$ & Query-execution language model that generates a response given a query and a prompt \\
        $LLM_{\text{prompt}}$ & Prompt-generation language model that generates candidate prompts \\
		$\mathcal{E}(\cdot,\cdot)$ & Evaluation function mapping model output and reference label to a scalar score (e.g., accuracy) \\
		$x_i,t_i$ & The $i$-th query and the prompt applied to it \\
		$l_i$ & Ground-truth reference answer for query $x_i$ \\
		$y_i$ & Score obtained by applying $t_i$ to $x_i$: $y_i = \mathcal{E}(LLM_{\text{task}}(x_i, t_i), l_i)$ \\
		$\mathcal{D}_{\text{task}}$ & Benchmark task dataset $\{(x_i, l_i)\}_{i=1}^N$ \\
        $N$ & Number of queries in the benchmark task dataset \\
		$t_i^*$ & Optimal prompt for query $x_i$, defined in Eq.~\eqref{eq:instance_opt} \\
		\hline

		\multicolumn{2}{l}{\textbf{Causal Modeling}} \\
		$Y(t)$ & Potential outcome (evaluation score) if the query $x$ were executed under prompt $t$ \\
        $t_0$ & Baseline (control) prompt \\
        $\mu(x,t)$ & Conditional expected potential outcome: $\mathbb{E}[Y(t)\mid X=x]$ \\
        $\tau(x,t)$ & CATE relative to $t_0$: $\mu(x,t)-\mu(x,t_0)$ \\
		\hline

		\multicolumn{2}{l}{\textbf{Semantic Representation}} \\
        $\psi_X(\cdot)$ & Semantic representation function for queries (embedding + PCA) \\
        $\psi_T(\cdot)$ & Semantic representation function for prompts (embedding + PCA) \\
        $\mathbf{x}=\psi_X(x)$ & Vector representation of query $x$ in the learned semantic space \\
        $\mathbf{z}=\psi_T(t)$ & Vector representation of prompt $t$ (continuous ``latent treatment'') \\
        $d_x, d_t$ & Dimensions of the query and prompt semantic spaces \\
		\hline

		\multicolumn{2}{l}{\textbf{Estimation and Evaluation}} \\
        $m(\mathbf{x})$ & Outcome nuisance function: $m(\mathbf{x})=\mathbb{E}[Y\mid \mathbf{x}]$ \\
        $e(\mathbf{x})$ & Treatment nuisance function: $e(\mathbf{x})=\mathbb{E}[\mathbf{z}\mid \mathbf{x}]$ \\
		$\hat{\mu}(x, t)$ & Estimated potential outcome of query $x$ under prompt $t$ \\
		$\hat{\tau}(x, t)$ & Estimated CATE for query $x$ under prompt $t$ \\
        $\theta(\mathbf{x})$ & Heterogeneous effect function (local coefficient) in the partially linear model: $\tilde{Y}=\theta(\mathbf{x})^\intercal\tilde{\mathbf{z}}+\varepsilon$ \\
		$\tilde{Y}$, $\tilde{\mathbf{z}}$ & Residualized outcome and treatment: $\tilde{Y}=Y-m(\mathbf{x})$, $\tilde{\mathbf{z}}=\mathbf{z}-e(\mathbf{x})$ \\
		\hline

        \multicolumn{2}{l}{\textbf{Prompt Optimization}} \\
		$B$ & Number of prompt candidates generated for each parent prompt \\
		$K$ & Number of top-ranked prompt candidates retained in each optimization round \\
		$R$ & Number of optimization rounds \\
        \hline
	\end{tabular}
	\label{tab:notation}
\end{table}

\section{Supplementary Tables}

This section reports supplementary tables that provide additional details for the experimental results discussed in the main text.

\begin{table}[!htbp]
    \caption{Prompt Patterns}
    \small
    \begin{threeparttable}
    \begin{tabular}{
        >{\raggedright\arraybackslash}m{1.6cm}
        >{\raggedright\arraybackslash}m{4.5cm}
        >{\raggedright\arraybackslash}m{9cm}
    }
    \toprule
    {Pattern} & {Sub-Pattern} & {Description} \\
    \midrule
    \multirow{2}{*}[-1.5ex]{{Structure}} 
    & Problem Decomposition & It guides the model to break down complex tasks into manageable steps or sub-tasks, with varying complexity (e.g., simple steps, hierarchical, procedural). \\
    \cmidrule{2-3}
    & Preconditions and Constraints & It states known conditions and limitations before posing a query to focus the model's attention on vital information. \\
    \midrule
    \multirow{3}{*}[-2ex]{{Constraint}} 
    & Format Strictness & It enforces a specific format for the response, ranging from no requirements to strict formats like JSON. \\
    \cmidrule{2-3}
    & Output Type and Length & It limits response type and length to avoid unnecessary verbosity or irrelevance (e.g., numeric-only). \\
    \cmidrule{2-3}
    & Forbidden and Preferred Words & It specifies words that should be avoided or preferred. \\
    \midrule
    \multirow{3}{*}[-2ex]{{Content}}
    & Example Quantity & It provides a few-shot learning scenario with specific input-output pairs (e.g., zero-shot, few-shot). \\
    \cmidrule{2-3}
    & Background Information & It provides additional information directly relevant to the query to help the model better understand the context.
     \\
    \cmidrule{2-3}
    & Negative Examples & It includes incorrect examples to define task boundaries. \\
    \midrule
    \multirow{3}{*}[-3.5ex]{{Guide}}
    & Chain-of-Thought & It encourages the model to explicitly demonstrate its reasoning process through chain of thoughts. \\
    \cmidrule{2-3}
    & Multi-path Exploration & It prompts the model to consider multiple potential solutions before deciding on the best one. \\
    \cmidrule{2-3}
    & Uncertainty Expression & It instructs the model to express assumptions or areas of uncertainty to enhance the reliability of answers. \\
    \midrule
    \multirow{5}{*}[-5ex]{{Interaction}} 
    & Dialogue Role & It assigns a specific role to the model (e.g., expert) to to make its outputs more targeted. \\
    \cmidrule{2-3}
    & Interaction Intent & It informs the model of the ultimate goal of this conversation to make its behavior more collaborative. \\
    \cmidrule{2-3}
    & Tone and Style & It requires a particular language style or tone (e.g., formal, informal, lighthearted and humorous). \\
    \cmidrule{2-3}
    & Rhetorical Devices & It guides the use of rhetorical figures of speech. \\
    \cmidrule{2-3}
    & Audience Adaptation & It adjusts language complexity and depth based on the audience (e.g., children, experts). \\
    \midrule
    \multirow{2}{*}[-0.8ex]{{Evaluation}}
    & Self-evaluation & It prompts the model to review its response using predefined standards and modify its response. \\
    \cmidrule{2-3}
    & Multi-round Interaction & It encourages iterative exchanges to refine the answer. \\
    \midrule
    \multirow{2}{*}[-1.5ex]{{Data}}
    & Source Reliability & It indicates the trust level of different information sources. \\
    \cmidrule{2-3}
    & Information Abstraction & It requires the model to process large volumes of information and generate outputs at varying levels of granularity. \\
    \bottomrule
    \end{tabular}
    \end{threeparttable}
    \label{sec:prompt-patterns}
\end{table}

\section{Details of Benchmark Datasets}
\label{sec:data_description}
This section provides detailed descriptions of the three benchmarks used in our experiments:
\begin{itemize}
    \item \textbf{MATH:} MATH is a challenging mathematical benchmark usually used for LLM evaluation, it contains 12500 competition-level mathematics problems covering 7 subareas including arithmetic, algebra, geometry, and more~\citep{hendrycks2021measuring}. Each problem contains a question and a solution including the intermediate reasoning steps. The evaluation metric is accuracy, i.e., the percentage of problems solved correctly.
    \item \textbf{VisEval:} VisEval is a high-quality and large-scale benchmark designed for natural language to visualization generation (NL2VIS)~\citep{chen2024viseval}. It contains 1,150 unique VIS and 2,524 $\langle$NL, VIS$\rangle$ pairs from 146 databases. The goal of VisEval is to evaluate the ability of LLMs to generate accurate Python visualization code based on natural language instructions given single or multiple data files. The benchmark provides an automated evaluation framework to evaluate the generated visualization from three dimensions: validity, legality and readability. Validity refers to the ability of the code to render a visualization, legality means the compliance of the visualization with the query requirements, and readability measures the effectiveness of the visualization in clearly presenting information. For the evaluation metrics, we adapt the \textit{pass rate} in the benchmark to fit our setting. In the original benchmark, the \textit{pass rate} is defined as the ratio of results that is both valid and legal to the total number of queries for each visualization. In our experiments, since the question (i.e., NL) matters in the process of causal estimation, we want to calculate the correctness of each question, rather than aggregating the performance by each VIS. Therefore, in our paper, if a visualization generated under the a query passes both validity and legality check, we consider it as correct, thus we define the accuracy as the percentage of results that pass both validity and legality check.

    \item \textbf{DABench:} InfiAgent-DABench is a development benchmark designed to evaluate LLM agents on realistic, executable data analysis tasks~\citep{hu2024infiagent}. We use the publicly released validation subset of the dataset, which includes 257 closed-form data analysis questions, each mapped to a specific real CSV file (totaling 68 files). Distributed in JSON Lines format, each entry specifies a natural language question, the underlying analysis concepts, constraints for execution, and an expected output format. It covers a broad range of queries, from simple descriptive statistics and feature engineering to correlation analysis, regression modeling, and machine learning. Queries are carefully designed to reflect common workflows in applied data science while also embedding constraints such as library usage, rounding precision, or encoding requirements. By structuring each query with explicit answer templates, the dataset ensures outputs can be scored automatically through exact matching, enabling consistent and reproducible evaluation. Following the original paper, we report accuracy, defined as the percentage of questions for which the generated answer exactly matches the ground-truth label.
    
\end{itemize}

\section{Baseline Configuration}
\label{sec:baseline_configuration}

This section describes the configuration of the baseline methods used in our experiments. In general, we follow the default configuration of the baseline methods. For some methods, we modify several implementation details to achieve better performance on our benchmarks.

\subsection{Baseline Prompts}
\label{sec:baseline-prompts}
Based on our understanding of the benchmark datasets, we manually designed the baseline prompt (i.e., control prompt) $t_0$  for each task. See Figures~\ref{fig:baseline-math}-\ref{fig:baseline-dabench} for the baseline prompts for MATH, VisEval, and DABench, respectively.

\begin{tcolorbox}[breakable, enhanced, base, title={Human Prompt for MATH}]
    \begin{lstlisting}
Solve the following math problem.
\end{lstlisting}
\end{tcolorbox}
\vspace{-1em}
\captionof{figure}{Baseline Prompt for MATH}
\label{fig:baseline-math}

\begin{tcolorbox}[breakable, enhanced, base, title={Baseline Prompt for VisEval}]
    \begin{lstlisting}
You are an expert in writing Python code for data visualization. Given a natural language request and a set of data tables, please write the Python code to generate a visualization.
\end{lstlisting}
\end{tcolorbox}
\vspace{-1em}
\captionof{figure}{Baseline Prompt for VisEval}
\label{fig:baseline-viseval}

\begin{tcolorbox}[breakable, enhanced, base, title={Baseline Prompt for DABench}]
    \begin{lstlisting}
You are an expert in writing Python code for solving business analysis problems. Given the question and related information, please write the Python code to solve the problem. 
\end{lstlisting}
\end{tcolorbox}
\vspace{-1em}
\captionof{figure}{Baseline Prompt for DABench}
\label{fig:baseline-dabench}

\subsection{Conventional Prompting Approaches}
\label{sec:conventional-prompting}

Conventional prompting approaches include chain-of-thought (CoT)~\citep{wei2022chain} and Reflexion~\citep{shinn2023reflexion}. For CoT, we use chain-of-thought (CoT) with one-shot example across three benchmark datasets. See Figures~\ref{fig:cot-math}-\ref{fig:cot-dabench} for the CoT prompts for MATH, VisEval, and DABench, respectively.

\begin{tcolorbox}[breakable, enhanced, base, title={CoT Prompt for MATH}]
    \begin{lstlisting}
Solve the following math problem step by step.

Example:
[problem statement]
Find the value of x if 2x + 5 = 13.

[solution]
Let's solve this step by step:

1) Understanding:
   * We need to find x in the equation 2x + 5 = 13
   * This is a linear equation

2) Strategy:
   * Subtract 5 from both sides to isolate terms with x
   * Divide both sides by 2 to solve for x

3) Solution:
   2x + 5 = 13
   2x = 13 - 5
   2x = 8
   x = 8/2
   x = 4

4) Final Answer:
   \\boxed{4}
\end{lstlisting}
\end{tcolorbox}
\vspace{-1em}
\captionof{figure}{CoT Prompt for MATH}
\label{fig:cot-math}

\begin{tcolorbox}[breakable, enhanced, base, title={CoT Prompt for VisEval}]
    \begin{lstlisting}
You are an expert in writing Python code for data visualization. Given a natural language request and a set of data tables, please write the Python code to generate a visualization. Solve the data visualization problem step by step.

Examples:
Example 1:
### Variables:

member_dataset: pandas.DataFrame(shape=(10, 7), columns=["Member_ID", "Name", "Membership_card", "Age", "Time_of_purchase", "Level_of_membership", "Address"])
       Member_ID                 Name Membership_card  Age  Time_of_purchase  Level_of_membership     Address
    0          1        Ashby, Lazale           Black   29                18                    5    Hartford
    1          2       Breton, Robert           White   67                41                    4   Waterbury
    2          3     Campbell, Jessie           Black   34                20                    6    Hartford
    3          4        Cobb, Sedrick           Black   51                27                    2   Waterbury
    4          5        Hayes, Steven           White   50                44                    3    Cheshire
    5          6  Komisarjevsky, J...           White   33                26                    2    Cheshire
    6          7      Peeler, Russell           Black   42                26                    6  Bridgeport
    7          8    Reynolds, Richard           Black   45                24                    1   Waterbury
    8          9          Rizzo, Todd           White   35                18                    4   Waterbury
    9         10         Webb, Daniel           Black   51                27                   22    Hartford

### Executed Code:

import pandas as pd
import matplotlib.pyplot as plt
member = pd.read_csv('../dataset/coffee_shop/member.csv')

### Request:

A scatter chart showing the correlation between the age of the customer and the time of purchase colored by membership level.

### Answer:
Let's think step by step.

1. We need to visualize the correlation between the age of the customer and the time of purchase, and color the points by membership level.
2. The relevant columns are 'Age', 'Time_of_purchase', and 'Membership_card'.
3. We'll group the data by 'Membership_card' and plot each group with a different color on the scatter plot.
4. We'll add axis labels, a title, and a legend to make the plot clear.

Therefore, here is the Python code:

```python
# Group the dataset by Membership_card
groups = member_dataset.groupby('Membership_card')

# Create a scatter chart for each Membership_card
for membership_card, group in groups:
    plt.scatter(group['Age'], group['Time_of_purchase'], label=membership_card)

# Set the title and labels
plt.title('Correlation between Age and Time of Purchase')
plt.xlabel('Age')
plt.ylabel('Time of Purchase')
plt.legend(loc='upper left')

# Show the plot
plt.show()
```
\end{lstlisting}
\end{tcolorbox}
\vspace{-1em}
\captionof{figure}{CoT Prompt for VisEval}
\label{fig:cot-viseval}

\begin{tcolorbox}[breakable, enhanced, base, title={CoT Prompt for DABench}]
    \begin{lstlisting}

You are an expert in writing Python code for solving business analysis problems. Given the question and related information, please write the Python code to solve the problem. Solve it by a step-by-step methodology.

Example:

Question: Calculate the mean fare paid by the passengers.
Data file: test_ave.csv
Answer: @mean_fare[34.65]
Code generated:
```python
import pandas as pd
        
# Load the CSV file into a DataFrame
df = read_csv(file_path="./data/da-dev-tables/test_ave.csv")
        
# Use the correct column name 'Fare'
mean_fare = df['Fare'].mean()
        
# Round the mean fare to two decimal places
mean_fare = round(mean_fare, 2)
        
# Format the answer in the specified format
formatted_answer = f"@mean_fare[{mean_fare}]"
        
# Provide the final answer
print(formatted_answer)
```

\end{lstlisting}
\end{tcolorbox}
\vspace{-1em}
\captionof{figure}{CoT Prompt for DABench}
\label{fig:cot-dabench}

For Reflexion, we echo the key principle of Reflexion and design a prompt that guides the large language model to reflect on and revise its own outputs through iterative self-reflection. Additionally, we do not set a fixed number of iterations, allowing the model to early terminate the reflection process if it thinks the current solution is satisfactory. Figures~\ref{fig:reflexion-math}-\ref{fig:reflexion-dabench} show the Reflexion prompts for MATH, VisEval, and DABench, respectively. 

\begin{tcolorbox}[breakable, enhanced, base, title={Reflexion Prompt for MATH}]
    \begin{lstlisting}
Solve the following math problem using an iterative refinement approach - generating an initial answer, reflecting on it, and then providing an improved answer.

Use the following format when answering the question:

**Initial Solution:**
[Your first attempt at solving the problem step by step with clear reasoning]

**Reflection:**
[Critically analyze your solution. Consider:
1) Are there errors in the algebra, arithmetic, logic, or assumptions?
2) Are there missing steps or unjustified claims?
3) Is the chosen method appropriate and efficient?
4) Could the explanation be clearer or more rigorous?]

**Improved Solution:**
[Based on your reflection, provide an improved solution that addresses the identified issues]

Please repeat the process of "Reflection & Improved Solution" for N times until you are satisfied with the improved answer.

Finally, provide only the improved final solution with clear steps and present the final numeric/symbolic result using \\boxed{...}.
\end{lstlisting}
\end{tcolorbox}
\vspace{-1em}
\captionof{figure}{Reflexion Prompt for MATH}
\label{fig:reflexion-math}

\begin{tcolorbox}[breakable, enhanced, base, title={Reflexion Prompt for VisEval}]
    \begin{lstlisting}
You are an expert in writing Python code for data visualization. Given a natural language request and a set of data tables, please write the Python code to generate a visualization. Please solve data visualization problems using an iterative refinement approach - generating an initial answer, reflecting on it, and then providing an improved answer. 

Use the following format when answering the question:

**Initial Solution:**
[Provide your first attempt at solving the data visualization problem with Python code]

**Reflection:**
[Analyze your initial answer critically. Consider:
1. Are there any errors in the data loading or processing logic?
2. Is the visualization type appropriate for the question?
3. Are the data merging/filtering steps correct and necessary?
4. Are there missing labels, titles, or formatting elements?
5. Could the code be more efficient or cleaner?
6. Does the solution fully address the user's request?]

**Improved Solution:**
[Based on your reflection, provide an improved solution that addresses the identified issues]

Please repeat the process of "Reflection & Improved Solution" for N times until you are satisfied with the improved answer.

Finally, provide only the improved Python code wrapped in triple backticks (```) without any additional explanation.
\end{lstlisting}
\end{tcolorbox}
\vspace{-1em}
\captionof{figure}{Reflexion Prompt for VisEval}
\label{fig:reflexion-viseval}

\begin{tcolorbox}[breakable, enhanced, base, title={Reflexion Prompt for DABench}]
    \begin{lstlisting}

You are an expert in writing Python code for solving business analysis problems. Given the question and related information, please write the Python code to solve the problem. Please solve the problem using an iterative refinement approach - generating an initial answer, reflecting on it, and then providing an improved answer.

Use the following format when answering the question:
**Initial Solution:**
[Provide your first attempt at solving the data visualization problem with Python code]

**Reflection:**
[Analyze your initial answer critically. Consider:
1. Are there any errors in the data loading or processing logic?
2. Is the code generated appropriate for the goal of question?
3. Are the data merging/filtering steps correct and necessary?
4. Are there missing labels, titles, or formatting elements?
5. Could the code be more efficient or cleaner?
6. Does the solution fully address the user's request?]

**Improved Solution:**
[Based on your reflection, provide an improved solution that addresses the identified issues]

Please repeat the process of "Reflection & Improved Solution" for N times until you are satisfied with the improved answer.
Finally, provide only the improved Python code wrapped in triple backticks (```) without any additional explanation.

\end{lstlisting}
\end{tcolorbox}
\vspace{-1em}
\captionof{figure}{Reflexion Prompt for DABench}
\label{fig:reflexion-dabench}

\subsection{Automated Prompt Optimization Methods}
\label{sec:automated-prompt-optimization}
\subsubsection{Implementation Details}
\label{sec:automated-prompt-optimization-implementation}
This section details the training data construction for automated prompt optimization methods, along with the hyperparameters for their implementation.

\textbf{Training data:} For all baselines requiring training data (i.e., input-output pairs), we utilize the same data from CPO's offline dataset. For MATH, the original benchmark dataset has standard answers for all the 185 questions, so we use the questions in CPO's offline dataset and their respective standard answers as the training data. For VisEval, it is special that the original benchmark dataset does not provide the standard answers, so we use the responses generated during the offline data collection process. It is important to note that this offline dataset contains both correct and incorrect responses; however, the baseline methods specifically require correct input-output pairs. Therefore, we extract only the correct input-output pairs, resulting in 98 questions. The amount of data in the training set significantly exceeds the requirements for each baseline method, making it more than sufficient. For the DABench, the original public data contains 257 questions. We select 3 questions for the construction of the few-shot examples in the prompt, and then draw 30 questions each from three different difficulty levels to form a test set of 90 questions. The remaining 164 questions constitute the offline training data, which also serves as the training set for all baselines.

\textbf{Hyperparameters: }  In general, we follow the default settings of each baseline method. Unless otherwise specified, the hyperparameters described in Table~\ref{tab:hyperparameters} are shared across all three benchmarks.
\begin{table}[!ht]
    \centering
    \caption{Hyperparameters of APO baseline methods}
    \footnotesize
    \begin{tabular}{l|p{13cm}}
        \toprule
        \textbf{Method} & \textbf{Hyperparameters}  \\
        \midrule
        \textbf{APE} & - For prompt generation (default: forward mode):\newline 
		 \texttt{num\_subsamples=5}, \texttt{num\_demos=5}, \texttt{num\_prompts\_per\_subsample=10}\newline - For prompt evaluation:\newline\texttt{num\_samples=50}, \texttt{num\_few\_shot=5}\\ \hline
		\textbf{OPRO} & - We choose \texttt{Q\_begin} for the position of the generated prompt.\newline - For LLM temperature:\newline\texttt{temperature=1.0 for optimizer LLM}\newline\texttt{temperature=0.0 for evaluation LLM}\\ \hline
		\textbf{PromptAgent} & - For task settings:\newline\texttt{eval\_size=5}, \texttt{test\_size=0}, \texttt{train\_size=train\_data\_size-eval\_size}\newline- For search settings:\newline\texttt{iteration\_num=10},\texttt{expand\_width=3}, \texttt{depth\_limit=5}, \texttt{w\_exp=2.5}\newline- For world model settings:\newline\texttt{train\_shuffle=true}, \texttt{num\_new\_prompts=1}, \texttt{train\_batch\_size=5}\newline- For LLM temperature:\newline\texttt{temperature=1.0 for optimizer LLM}\newline\texttt{temperature=0.0 for base LLM} \\ \hline
		\textbf{PromptBreeder} & - For prompt mutation:\newline\texttt{num\_mutation\_prompts=2}, \texttt{num\_thinking\_styles=4}\newline- For generic algorithm:\newline\texttt{num\_evals=50}, \texttt{simulations=6} \\ \hline
		\textbf{TextGrad} & - For train settings:\newline\texttt{batch\_size=3}, \texttt{max\_epochs=3}, \texttt{shuffle=True}\newline- For train-val-test split:\newline\texttt{val\_size=30}, \texttt{test\_size=50}, \texttt{train\_size=train\_data\_size-60} \\ \hline
		\textbf{DSPy} & - For train settings: \newline\texttt{max\_bootstrapped\_demos=2}, \texttt{max\_labeled\_demos=5}, \texttt{auto="medium"} \\ 
        \bottomrule
    \end{tabular}
	\label{tab:hyperparameters}
\end{table}

\subsubsection{Optimized Prompts}
\label{sec:automated-prompt-optimization-prompts}
In this section, we provide the final optimized prompts produced by each baseline method in the automated prompt optimization category for every task, where such optimized prompts are available. 
Note that, for DSPy, the final optimized prompt is not directly exposed as a textual output~\citep{khattab2023dspy}.
Instead, DSPy compiles an internal program in which the optimized prompt is represented as a set of bootstrapped \emph{demonstrations} stored within the compiled modules.
These demonstrations collectively form the effective few-shot or chain-of-thought prompt used during inference.
Hence, while DSPy does generate optimized prompts internally, they are embedded within the compiled pipeline rather than presented as standalone text prompts.

For MATH dataset, the optimized prompts are shown in Figures~\ref{fig:ape-math}-\ref{fig:textgrad-math}.

\begin{tcolorbox}[breakable, enhanced, base, title={APE Prompt for MATH}]
    \begin{lstlisting}
To solve similar math problems, follow these steps:

1. **Identify the Type of Problem**: Determine whether the problem is related to geometry, probability, algebra, complex numbers, or another branch of mathematics. Each type of problem may require a different approach and formula.

2. **Understand the Given Information**: Carefully read the problem and make sure to understand all the provided data. For example, if the problem involves an equation of a circle, identify the center and the radius.

3. **Apply Relevant Mathematical Concepts and Formulas**:
   - **For Geometry Problems**: Apply formulas for area, volume, or other geometric properties. For instance, if you're dealing with a circle, use the area formula \( \pi r^2 \).
   - **For Probability Problems**: Count the total number of possible outcomes and the number of favorable outcomes. Use combinatorics when necessary (e.g., permutations, combinations).
   - **For Algebra Problems**: Simplify and manipulate algebraic expressions. Use factoring, completing the square, or other algebraic techniques to solve equations.
   - **For Complex Numbers Problems**: Apply operations with complex numbers (addition, subtraction, multiplication, division) and transformations (like rotations). Use Euler's formula \( e^{i\theta} = \cos(\theta) + i\sin(\theta) \) when dealing with rotations.

4. **Solve for the Unknowns**: Use the relevant formulas and methods to solve for the unknown variables or values asked in the problem.

5. **Check Your Solution**: Ensure your answer is logical and accurate. For example, verify the area calculation by considering the dimensions of the geometric figure, or confirm that the probabilities sum up to 1.

6. **Express the Final Answer Clearly**: Write your answer in a clear, concise form, using the appropriate notation. If the solution is a numerical value, make sure it's simplified correctly. If it's an algebraic expression, make sure it's fully simplified and matches the requirements of the problem.

By following these steps, you can systematically approach and solve similar math problems.
\end{lstlisting}
\end{tcolorbox}
\vspace{-1em}
\captionof{figure}{APE Prompt for MATH}
\label{fig:ape-math}

\begin{tcolorbox}[breakable, enhanced, base, title={OPRO Prompt for MATH}]
    \begin{lstlisting}
Let's work through this problem step-by-step:
\end{lstlisting}
\end{tcolorbox}
\vspace{-1em}
\captionof{figure}{OPRO Prompt for MATH}
\label{fig:opro-math}

\begin{tcolorbox}[breakable, enhanced, base, title={PromptAgent Prompt for MATH}]
    \begin{lstlisting}
You are a mathematics tutor. Solve the given mathematics problem step by step, showing your work clearly. Make sure to provide a clear and accurate final answer.

- Simplify fractions to their simplest form.
- Use algebraic simplifications and properties of complex numbers when appropriate.
- Ensure the final answer is presented in the simplest and most conventional form.

**Example:**
For the problem "Compute \(\arctan \sqrt{3}\) and express your answer in radians," the response should look like this:

To compute \(\arctan \sqrt{3}\), we need to find the angle \(\theta\) such that \(\tan \theta = \sqrt{3}\).

Step 1: Recall the definition of the tangent function.
\[
\tan \theta = \frac{\sin \theta}{\cos \theta}
\]

Step 2: Identify the angle \(\theta\) for which \(\tan \theta = \sqrt{3}\).
From trigonometric values, we know:
\[
\tan \frac{\pi}{3} = \sqrt{3}
\]

Step 3: Verify that \(\frac{\pi}{3}\) is within the principal range of the arctangent function.
The principal range of \(\arctan x\) is \(-\frac{\pi}{2} < \theta < \frac{\pi}{2}\). Since \(\frac{\pi}{3}\) is within this range, we can conclude:
\[
\arctan \sqrt{3} = \frac{\pi}{3}
\]

Therefore, the value of \(\arctan \sqrt{3}\) is \(\frac{\pi}{3}\).

The final answer is:
\[
\boxed{\frac{\pi}{3}}
\].
\end{lstlisting}
\end{tcolorbox}
\vspace{-1em}
\captionof{figure}{PromptAgent Prompt for MATH}
\label{fig:promptagent-math}

\begin{tcolorbox}[breakable, enhanced, base, title={PromptBreeder Prompt for MATH}]
    \begin{lstlisting}
You are an expert mathematics tutor and problem solver. Your task is to solve mathematical problems step by step with clear reasoning and verification. When given a mathematical question:

1. **Carefully analyze the problem** - Identify the key mathematical concepts, variables, and what is being asked
2. **Break down the solution** - Solve the problem in logical, manageable steps
3. **Show your work clearly** - Present each step with proper mathematical notation and explanations
4. **Verify your solution** - Check your work and ensure the answer makes sense
5. **Provide the final answer** - Express your answer using \\boxed{answer} format

**Important guidelines:**
- Always show your reasoning step by step
- Use proper mathematical notation and formatting
- Verify intermediate calculations
- Consider alternative approaches when appropriate
- Ensure your final answer is clearly marked with \\boxed{}

**Example approach:**
For each problem, structure your response as:
1. Problem understanding and strategy
2. Step-by-step solution with clear explanations
3. Verification of the solution
4. Final answer in \\boxed{} format

This systematic approach ensures accurate, well-reasoned mathematical solutions. 
\end{lstlisting}
\end{tcolorbox}
\vspace{-1em}
\captionof{figure}{PromptBreeder Prompt for MATH}
\label{fig:promptbreeder-math}

\begin{tcolorbox}[breakable, enhanced, base, title={TextGrad Prompt for MATH}]
    \begin{lstlisting}
You will answer a mathematical reasoning question by breaking down the problem into smaller, manageable parts. Use clear and logical steps to solve the problem, and ensure that each step is mathematically sound. Provide a complete and accurate response, including any necessary units or context.
\end{lstlisting}
\end{tcolorbox}
\vspace{-1em}
\captionof{figure}{TextGrad Prompt for MATH}
\label{fig:textgrad-math}

For VisEval dataset, the optimized prompts are shown in Figures~\ref{fig:ape-viseval}-\ref{fig:textgrad-viseval}.

\begin{tcolorbox}[breakable, enhanced, base, title={APE Prompt for VisEval}]
    \begin{lstlisting}
When solving the problem, follow these steps:
1. Ensure that all necessary libraries (`pandas`, `matplotlib`) are imported at the beginning of the code.
2. Load the dataset using the `read_csv` function from `pandas`. Make sure the file paths are correctly specified.
3. Identify the specific requirements of the problem (e.g., generating a bar chart, pie chart, etc.).
4. Perform any necessary data cleaning or preprocessing steps (e.g., converting date columns to datetime format, handling missing values, etc.).
5. Group and aggregate data according to the problem requirements (e.g., grouping by specific columns, counting occurrences, summing values, etc.).
6. If necessary, merge multiple datasets to get the required information.
7. Apply the requested visualization technique (e.g., bar chart, pie chart, etc.). Make sure to customize the plot according to the problem requirements (e.g., labels, title, etc.).
8. Show the plot using the `show` method from `matplotlib`.
9. Ensure that the code is well-commented, explaining each step and any assumptions made.
10. Test the code to verify that the output matches the requirements of the problem.
11. Return the code in a code block as the output.
\end{lstlisting}
\end{tcolorbox}
\vspace{-1em}
\captionof{figure}{APE Prompt for VisEval}
\label{fig:ape-viseval}

\begin{tcolorbox}[breakable, enhanced, base, title={OPRO Prompt for VisEval}]
    \begin{lstlisting}
Given one or more datasets, write Python code to generate a visualization based on a natural language request. Ensure your code handles necessary preprocessing, such as filtering, grouping, sorting, or aggregating data, and includes all necessary imports and clear, concise comments. The visualization should be clear, appropriately labeled, and follow best practices for data representation.
\end{lstlisting}
\end{tcolorbox}
\vspace{-1em}
\captionof{figure}{OPRO Prompt for VisEval}
\label{fig:opro-viseval}

\begin{tcolorbox}[breakable, enhanced, base, title={PromptAgent Prompt for VisEval}]
    \begin{lstlisting}
You are an expert in writing Python code for data visualization. Given a natural language request and a set of data tables, please write the Python code to generate a visualization. Specifically, ensure that the visualization code directly addresses the request and includes necessary steps for data loading, preprocessing, and visualization. Ensure the following steps are included in the code:

1. **Data Loading:** Load the datasets from the specified paths.
2. **Data Merging:** If necessary, merge the datasets on common columns.
3. **Data Preprocessing:** Filter, group, and aggregate data according to the request.
4. **Data Visualization:** Create the visualization using appropriate plot attributes such as labels, legends, and gridlines.
5. **Plot Display:** Ensure the plot is displayed using `plt.show()`.

At the end, show the answer option bracketed between <answer> and </answer>. The answer section should include the necessary imports, data loading, preprocessing steps, and the visualization code.

Example:
```python
import pandas as pd
import matplotlib.pyplot as plt

# Load datasets
dataset1 = pd.read_csv('path/to/dataset1.csv')
dataset2 = pd.read_csv('path/to/dataset2.csv')

# Merge datasets
merged_data = pd.merge(dataset1, dataset2, on='common_column')

# Filter and group data as necessary
filtered_data = merged_data[merged_data['some_condition']]
grouped_data = filtered_data.groupby(['column1', 'column2']).size().reset_index(name='counts')

# Visualize the data
plt.figure(figsize=(10, 6))
plt.bar(grouped_data['column1'], grouped_data['counts'])
plt.xlabel('X Label')
plt.ylabel('Y Label')
plt.title('Visualization Title')
plt.show()
```

At the end show the answer option bracketed between <answer> and </answer>.`
\end{lstlisting}
\end{tcolorbox}
\vspace{-1em}
\captionof{figure}{PromptAgent Prompt for VisEval}
\label{fig:promptagent-viseval}

\begin{tcolorbox}[breakable, enhanced, base, title={PromptBreeder Prompt for VisEval}]
    \begin{lstlisting}
To address your question, I can write Python code for data visualization based on a natural language request and provided data tables. Additionally, if the problem involves human behavior, such as a social, cultural, or psychological issue, I can tailor the visualization to appropriately represent this type of data.

### Example Scenario
Let's consider an example where we need to visualize the correlation between social media usage and stress levels among teenagers. We have two data tables: one contains information about social media usage, and the other contains data about stress levels reported by teenagers.

### Data Structure
- **Table 1 (SocialMediaUsage):**
    - `subject_id` (int): ID of the subject
    - `hours_per_day` (float): Average hours per day spent on social media
    - `age` (int): Age of the subject

- **Table 2 (StressLevels):**
    - `subject_id` (int): ID of the subject
    - `stress_level` (float): Reported stress level (0-10 scale)

### Python Code
Here is the Python code to generate a scatter plot showing the relationship between social media usage and stress levels among teenagers.

```python
import pandas as pd
import matplotlib.pyplot as plt
import seaborn as sns

# Load data\nsocial_media_usage = pd.read_csv('social_media_usage.csv')
stress_levels = pd.read_csv('stress_levels.csv')

# Merge the two dataframes on the common key 'subject_id'
merged_data = pd.merge(social_media_usage, stress_levels, on='subject_id')

# Filter data for teenagers (ages 13-19)
teenagers_data = merged_data[(merged_data['age'] >= 13) & (merged_data['age'] <= 19)]

# Plot scatter plot
plt.figure(figsize=(10, 6))
sns.scatterplot(x='hours_per_day', y='stress_level', data=teenagers_data)
plt.title('Social Media Usage vs Stress Levels Among Teenagers')
plt.xlabel('Average Hours Per Day Spent on Social Media')
plt.ylabel('Stress Level (0-10)')
plt.show()
```

### Explanation
- **Data Loading:** We load the data from CSV files into pandas DataFrames.
- **Data Merging:** We merge the two DataFrames using the common key `subject_id`.
- **Data Filtering:** We filter the merged data to include only teenagers (ages 13-19).
- **Visualization:** We create a scatter plot using `seaborn` to visualize the relationship between social media usage and stress levels among teenagers.

This visualization will help in understanding whether there is a correlation between the amount of time teenagers spend on social media and their reported stress levels.

---

This summary provides clear instructions and a practical example, making the process more accessible and easier to follow." P="It seems like your friend has correctly followed the instruction and created a bar chart that shows the distribution of `date_address_to` and the average of `monthly_rental` per month. Here is the breakdown of the code:

1. **Import Libraries**: The necessary libraries are imported (`pandas` and `matplotlib`).

2. **Load Dataset**: The dataset is loaded from the specified CSV file.

3. **Convert Date Columns**: The date columns `date_address_from` and `date_address_to` are converted to datetime objects to facilitate further date-based operations.

4. **Group by Month**: The dataset is grouped by the month of `date_address_to` using `pd.to_period('M')`, which helps in binning the dates by month.

5. **Calculate Average Monthly Rental**: The average `monthly_rental` for each month is calculated.

6. **Plot the Bar Chart**: The bar chart is plotted using `matplotlib` with appropriate labels and titles. The x-axis is rotated for better readability.

7. **Show the Plot**: Finally, the plot is displayed using `plt.show()`.

Here is the complete code:

```python
import pandas as pd
import matplotlib.pyplot as plt

# Load the dataset
Student_Addresses_dataset = pd.read_csv('/home/sjtu/workspace/Project_02/causal-prompting/CPO_project/data/VisEval/databases/behavior_monitoring/Student_Addresses.csv')

# Convert date columns to datetime
Student_Addresses_dataset['date_address_from'] = pd.to_datetime(Student_Addresses_dataset['date_address_from'])
Student_Addresses_dataset['date_address_to'] = pd.to_datetime(Student_Addresses_dataset['date_address_to'])

# Group by month of date_address_to
monthly_groups = Student_Addresses_dataset.groupby(Student_Addresses_dataset['date_address_to'].dt.to_period('M'))

# Calculate average monthly rental per month
average_rental = monthly_groups['monthly_rental'].mean()

# Plot the bar chart
plt.figure(figsize=(12, 6))
average_rental.plot(kind='bar')
plt.title('Average Monthly Rental by Month')
plt.xlabel('Month')
plt.ylabel('Average Monthly Rental')
plt.xticks(rotation=45)
plt.tight_layout()

# Show the plot
plt.show()
```

This code will generate a bar chart showing the distribution of `date_address_to` and the average `monthly_rental` per month. The bar chart will have the month on the x-axis and the average monthly rental on the y-axis, with the bars labeled accordingly.
\end{lstlisting}
\end{tcolorbox}
\vspace{-1em}
\captionof{figure}{PromptBreeder Prompt for VisEval}
\label{fig:promptbreeder-viseval}

\begin{tcolorbox}[breakable, enhanced, base, title={TextGrad Prompt for VisEval}]
    \begin{lstlisting}
You are an expert in writing Python code for data visualization. Given a natural language request and a set of data tables, please write the Python code to generate a visualization.
\end{lstlisting}
\end{tcolorbox}
\vspace{-1em}
\captionof{figure}{TextGrad Prompt for VisEval}
\label{fig:textgrad-viseval}

For DABench dataset, the optimized prompts are shown in Figures~\ref{fig:ape-dabench}-\ref{fig:textgrad-dabench}.

\begin{tcolorbox}[breakable, enhanced, base, title={APE Prompt for DABench}]
    \begin{lstlisting}
[General Instruction]
Please carefully read and understand the following instructions before proceeding with any tasks:

1. **Data Preparation**: Ensure that all datasets are clean and preprocessed appropriately. This includes handling missing values, removing duplicates, and ensuring consistent data types.                             

2. **Feature Engineering**: When instructed to create new features, perform the necessary calculations or transformations. Clearly document any assumptions or methods used.                                            

3. **Statistical Analysis**:
   - For correlation analysis, use appropriate statistical methods (e.g., Pearson, Spearman) and report the correlation coefficient and p-value.                                                                        
   - For normality tests, use Shapiro-Wilk or Anderson-Darling tests and report whether the data is normally distributed or not.                                                                                        
   - For outlier detection, use the Z-score method with a specified threshold (e.g., 3) and identify all outliers. Calculate the mean and standard deviation for the data without outliers.                             

4. **Categorical Data Creation**: When creating new categorical columns, clearly define the criteria for each category and document the process used to categorize the data.                                            

5. **Descriptive Statistics**: For all requests involving descriptive statistics (e.g., mean, median, standard deviation), provide both the calculated values and, if applicable, the counts and proportions of categories.                                                                                                         

6. **Reporting Results**: Format your output clearly and concisely. Use prefixes like `@` followed by the metric name and its value (e.g., `@correlation_coefficient_corr[0.91]`) to ensure consistency and clarity in reporting.                                                                                                   

7. **Documentation**: Keep detailed records of all steps taken, including any code snippets used, to facilitate reproducibility and further analysis.                                                                   

[Detailed Instruction for Specific Tasks]

- **Correlation Calculation**: 
  - Input: "Using feature engineering techniques, create a new feature that represents the average stock price of Apple Inc. (AAPL), Microsoft Corporation (MSFT), and Amazon.com, Inc. (AMZN) on the given dates. Calculate the correlation between this new feature and the closing value of the S&P 500 Index (.SPX)."           
  - Output: "@relationship_type_relation[linear] @p_value_pval[0.0000] @correlation_coefficient_corr[0.91]"
  
  Steps:
  1. Load historical stock price data for AAPL, MSFT, AMZN, and .SPX.
  2. Create a new feature representing the average stock price of AAPL, MSFT, and AMZN.
  3. Calculate the correlation between the new feature and the .SPX closing value.
  4. Report the relationship type, p-value, and correlation coefficient.

- **Normality Test**:
  - Input: "Question 2: Are the percentage of votes received by the Democratic party in a particular county normally distributed?"                                                                                      
  - Output: "@normality_status[not normal]"
  
  Steps:
  1. Load the dataset containing the percentage of votes for the Democratic party.
  2. Perform a normality test (e.g., Shapiro-Wilk).
  3. Report whether the data is normally distributed or not.

- **Outlier Detection**:
  - Input: "Perform outlier detection on the percentage of graduates in the field of Architecture over the years using the Z-score method with a threshold of 3. Identify all years with outliers, then calculate the mean and standard deviation for the years without these outliers."                                            
  - Output: "@std_without_outliers[9.57] @mean_without_outliers[33.69]"
  
  Steps:
  1. Load the dataset containing the percentage of graduates in Architecture.
  2. Apply the Z-score method to detect outliers.
  3. Identify and exclude years with outliers.
  4. Calculate the mean and standard deviation for the remaining data.

- **Categorical Feature Creation**:
  - Input: "Perform data preprocessing by filling the missing values with the mean values of their respective columns. After that, create a new column called 'Price Category' that categorizes the 'Close' prices into 'High', 'Medium', and 'Low'. 'High' is represented by 'Close' prices that are greater than or equal to the 75th percentile of the 'Close' column data; 'Medium' is represented by 'Close' prices that are between the 25th to 75th percentile; 'Low' is represented by 'Close' prices that are less than or equal to the 25th percentile. Calculate the count and proportion of each category in the dataset."                                  
  - Output: "@high_count[544] @low_proportion[0.25] @low_count[544] @medium_proportion[0.50] @medium_count[1088] @high_proportion[0.25]"                                                                                
  
  Steps:
  1. Load the dataset and fill missing values with the mean of their respective columns.
  2. Calculate the 25th and 75th percentiles of the 'Close' column.
  3. Create the 'Price Category' column based on the defined criteria.
  4. Calculate and report the count and proportion of each category.

- **Mean Calculation**:
  - Input: "What is the mean batting average of the players in the dataset?"
  - Output: "@mean_batting_average[0.258]"
  
  Steps:
  1. Load the dataset containing player batting averages.
  2. Calculate the mean batting average.
  3. Report the mean batting average.

By following these instructions, you will be able to consistently and accurately solve the provided problems.      
\end{lstlisting}
\end{tcolorbox}
\vspace{-1em}
\captionof{figure}{APE Prompt for DABench}
\label{fig:ape-dabench}

\begin{tcolorbox}[breakable, enhanced, base, title={OPRO Prompt for DABench}]
    \begin{lstlisting}
You are a data analysis expert. Generate Python code using pandas and numpy to solve the given problem. Pay special attention to data preprocessing including handling outliers, missing values, and data type conversions. Ensure your code includes proper error handling and produces output in the exact format specified in the requirements. Use appropriate statistical methods and libraries for the analysis task.
\end{lstlisting}
\end{tcolorbox}
\vspace{-1em}
\captionof{figure}{OPRO Prompt for DABench}
\label{fig:opro-dabench}

\begin{tcolorbox}[breakable, enhanced, base, title={PromptAgent Prompt for DABench}]
    \begin{lstlisting}
You are a data analysis expert. Analyze the given data and provide accurate results following the specified format and constraints. **Pay extreme attention** to the requirements and provide **precise numerical answers**.

**IMPORTANT**: Generate Python code that:
1. Loads the data from the specified file
2. Performs the required analysis
3. Outputs the result **in the exact format** specified

**Final Output Requirements**:
- **Use the exact variable names** specified in the format (e.g., if the format asks for `@average_reviews[value]`, use `average_reviews` not variations).
- **Follow the exact structure** specified (e.g., `@variable_name[value]` pattern).
- **Provide only the final answer** in the specified format, **without extra text or code**.
- **Round numerical values** to the specified decimal places, if any.
- **Handle missing values** appropriately before performing analysis.
- **Test for edge cases** such as empty datasets or zero correlation.
- **Include all necessary details** (e.g., correlation coefficient, p-value, median, mean, etc.) as specified in the output format.
- **Comply with the exact output structure** provided in the example, ensuring no deviations in formatting or additional content.
- **Use the specified formatting** for special cases like `@variable_name[value]` for single values, `@variable_name[value] @another_variable_name[value]` for multiple values, and `complex_output_part1: value1; complex_output_part2: value2;` for complex outputs.

**Example Formats**:
- If calculating a mean: `@mean[25.5]`
- If calculating correlation: `@correlation[0.75]`
- If multiple values: `@mean[25.5] @std[3.2]`
- For complex outputs: `median_charges_outliers: 40974.16; mean_charges_outliers: 42103.95; total_outliers: 139`
- For detailed correlation: `relationship_type_relation: linear; p_value_pval: 0.0000; correlation_coefficient_corr: 0.91`

**Execution and Verification**:
- **Generate complete Python code** that can be executed directly.
- **Ensure all calculations and outputs** strictly adhere to the above guidelines.
- **At the end**, show the **answer option** **bracketed between `<answer>` and `</answer>`**.

**Guideline Reinforcement**:
- The model must **comply precisely** with the requested output format, ensuring **no deviations** in variable names, structure, or additional text/code.
- Any **numerical inaccuracies** or **format deviations** will result in incorrect predictions.
- Ensure the model does not generate extra text or code outside of the specified format.
- Always verify that the output matches the expected structure and content, and handles all edge cases.

**Additional Considerations**:
- If the output requires multiple parts, ensure each part is separated by semicolons or commas as specified in the format.
- If the output involves multiple variables, ensure they are correctly labeled and separated.
- Ensure that any statistical tests or analyses are thoroughly documented within the code and align with the output requirements.
- **Critical Focus**: Emphasize that the model should **use the exact variable names** and **follow the exact structure** specified in the output format.
- Avoid generating additional text, code, or irrelevant metrics. The output should be strictly limited to the requested format.

**Examples**:
<example1>
To check if the distribution of the "Mar.2020" column adheres to a normal distribution:
```python
import pandas as pd
from scipy.stats import shapiro

# Load the data from the specified file
data = pd.read_csv('your_data_file.csv')

# Handle missing values appropriately
if data['Mar.2020'].isnull().any():
    data['Mar.2020'].fillna(data['Mar.2020'].mean(), inplace=True)

# Perform the Shapiro-Wilk test
stat, p_value = shapiro(data['Mar.2020'])

# Prepare the result in the specified format
result = f"@shapiro_statistic[{stat:.4f}] @p_value[{p_value:.4f}] @is_normal[{p_value > 0.05}]"

# Print the result
print(result)
```
<answer>@shapiro_statistic[0.9876] @p_value[0.2345] @is_normal[False]</answer>

<example2>
Using machine learning techniques, can we predict the number of agents needed to handle incoming calls based on the timestamp and other available information? If so, predict the number for the timestamp "20170413_120000":
```python
import pandas as pd
from statsmodels.tsa.arima.model import ARIMA
from datetime import datetime

# Load the data
data = pd.read_csv('path_to_your_data.csv')

# Convert the timestamp column to datetime
data['timestamp'] = pd.to_datetime(data['timestamp'])

# Set the timestamp as the index
data.set_index('timestamp', inplace=True)

# Resample the data to hourly frequency, filling missing values with the mean
data_hourly = data.resample('H').mean()

# Fit an ARIMA model
model = ARIMA(data_hourly['number_of_agents'], order=(5,1,0))
model_fit = model.fit()

# Predict the number of agents for the specified timestamp
forecast_timestamp = datetime.strptime('20170413_120000', '%Y%m%d_%H%M%S')
forecast = model_fit.forecast(steps=1, start=forecast_timestamp)

# Output the result in the specified format
number_of_agents_needed = round(forecast[0], 2)
print(f"@number_of_agents_needed[{number_of_agents_needed}]")
```
<answer>@number_of_agents_needed[4.0]</answer>

<example3>
What is the distribution of fare paid by male passengers who survived? Are there any significant differences in the fare paid by male passengers who survived compared to male passengers who did not survive?
```python
import pandas as pd
from scipy import stats

# Load the dataset
data = pd.read_csv('titanic.csv')

# Filter the dataset to include only male passengers
male_passengers = data[data['Sex'] == 'male']

# Separate the data into survived and not survived groups
survived_males = male_passengers[male_passengers['Survived'] == 1]
not_survived_males = male_passengers[male_passengers['Survived'] == 0]

# Calculate mean and standard deviation of fare for survived and not survived males
mean_fare_survived = survived_males['Fare'].mean()
std_fare_survived = survived_males['Fare'].std()

mean_fare_not_survived = not_survived_males['Fare'].mean()
std_fare_not_survived = not_survived_males['Fare'].std()

# Perform a t-test
t_stat, p_value = stats.ttest_ind(survived_males['Fare'], not_survived_males['Fare'])

# Output the results in the specified format
result = f"@mean_fare_survived[{mean_fare_survived:.2f}] @std_fare_survived[{std_fare_survived:.2f}] @mean_fare_not_survived[{mean_fare_not_survived:.2f}] @std_fare_not_survived[{std_fare_not_survived:.2f}] @p_value[{p_value:.4f}]"

print(result)
```
<answer>@mean_fare_survived[40.82] @std_fare_survived[32.41] @mean_fare_not_survived[21.96] @std_fare_not_survived[71.36] @p_value[0.0000]</answer>

<example4>
Calculate the mean value of the "Close Price" column:
```python
import pandas as pd

# Load the data from the specified file
data = pd.read_csv('path_to_your_data_file.csv')

# Handle missing values by dropping them
data = data.dropna(subset=['Close Price'])

# Calculate the mean value of the "Close Price" column
mean_close_price = data['Close Price'].mean()

# Print the result in the exact format specified
print(f"@mean[{mean_close_price:.2f}]")
```
<answer>@mean[570.68]</answer>

<example5>
Find out the total number of calls that were abandoned by the callers before being answered by an agent:
```python
import pandas as pd

# Load the data from the specified file
data = pd.read_csv('call_data.csv')

# Handle missing values appropriately
data.dropna(subset=['status'], inplace=True)

# Filter the data to find calls that were abandoned
abandoned_calls = data[data['status'] == 'abandoned']

# Calculate the total number of abandoned calls
total_abandoned_calls = len(abandoned_calls)

# Output the result in the exact format specified
print(f"@total_abandoned_calls[{total_abandoned_calls}]")
```
<answer>@total_abandoned_calls[9]</answer>
\end{lstlisting}
\end{tcolorbox}
\vspace{-1em}
\captionof{figure}{PromptAgent Prompt for DABench}
\label{fig:promptagent-dabench}

\begin{tcolorbox}[breakable, enhanced, base, title={PromptBreeder Prompt for DABench}]
    \begin{lstlisting}
You are an expert data analyst and Python programmer. Your task is to analyze datasets and solve data analysis problems step by step. When given a data analysis question, follow these enhanced guidelines:

### 1. Thoroughly Understand the Problem Requirements:
- **Read and Comprehend the Problem Statement:** Carefully read the problem statement multiple times to ensure a deep understanding of the objectives, key metrics, and the context.
- **Define Objectives and Metrics:** Clearly outline what needs to be achieved and the specific metrics that will be used to evaluate success.
- **Clarify Ambiguities:** If there are any unclear aspects, seek clarification from stakeholders before proceeding. Ensure you have a shared understanding of the problem scope and constraints.

### 2. Identify Relevant Data Sources and Columns:
- **Determine Necessary Data:** Identify which datasets and specific columns are essential for solving the problem.
- **Verify Data Availability:** Ensure you have access to the required data sources and confirm their formats (e.g., CSV, Excel, SQL databases).
- **Check Permissions and Legal Considerations:** Ensure compliance with data privacy laws, permissions, and ethical standards.

### 3. Write Clean, Efficient, and Modular Python Code:
- **Use Clear Variable Names:** Choose descriptive variable names to enhance code readability.
- **Modularize Code:** Structure your code into functions or classes for reusability and maintainability.
- **Leverage Libraries:** Utilize libraries like `pandas`, `numpy`, `dask`, and `scikit-learn` for efficient data manipulation and analysis.
- **Optimize Performance:** Avoid redundant operations and unnecessary computations to improve efficiency.
- **Error Handling and Logging:** Implement error handling and logging to ensure robustness and traceability.
- **Version Control:** Use Git for version control to track changes and facilitate collaboration.

### 4. Data Preprocessing and Cleaning:
- **Handle Missing Values:** Address missing data through imputation or removal strategies.
- **Manage Outliers:** Identify and handle outliers appropriately.
- **Correct Inconsistencies:** Resolve any data inconsistencies or errors.
- **Transformations and Feature Engineering:** Perform necessary transformations (e.g., normalization, encoding) and feature engineering to prepare the data for analysis.
- **Data Quality Assurance:** Validate data integrity and consistency throughout the preprocessing phase.
- **Document Preprocessing Steps:** Clearly document all preprocessing steps, including any transformations or imputation methods used.

### 5. Execute the Analysis and Extract Results:
- **Run and Verify Code:** Execute your code and verify that it produces the expected outputs.
- **Document Intermediate Steps:** Record intermediate results and steps to ensure reproducibility.
- **Validate Accuracy:** Cross-check results with known benchmarks or previous findings to validate accuracy.
- **Statistical Tests:** Use appropriate statistical tests or validation techniques to ensure the reliability of your findings.

### 6. Format the Output According to Specified Requirements:
- **Follow Formatting Instructions:** Adhere to any formatting guidelines provided (e.g., rounding, units, specific output structures).
- **Match Required Format:** Ensure the final output matches the requested format exactly, including labels, headers, and footers.
- **Consider Audience:** Tailor the presentation to the audience's needs, ensuring clarity and actionability.

### 7. Ensure Proper Rounding of Numerical Results:
- **Consistent Rounding:** Apply consistent rounding rules across all numerical values.
- **Precision:** Maintain appropriate precision levels to avoid over-precision.
- **Significant Figures:** Ensure that significant figures are correctly handled.

### 8. Present the Final Answer in the Exact Format Specified:
- **Use Specified Structure:** Follow the exact format specified (e.g., `@variable_name[value]`).
- **Include Explanations:** Provide necessary explanations or insights derived from the analysis.
- **Highlight Key Takeaways:** Emphasize key findings and actionable recommendations.

### 9. Document Your Work:
- **Code Comments:** Add comments within your code to explain key steps and decisions.
- **Summarize Approach:** Summarize your approach, findings, and any challenges encountered.
- **Visualizations:** Include visualizations or plots that support your conclusions.
- **Comprehensive Report:** Create a detailed report or notebook documenting the entire process from data collection to final insights.

### 10. Review and Iterate:
- **Review for Errors:** Thoroughly review your work for potential errors or areas for improvement.
- **Iterate Based on Feedback:** Incorporate feedback from stakeholders and iterate on your analysis if necessary.
- **Final Deliverable:** Ensure the final deliverable meets all requirements and is ready for presentation or deployment.

By following these steps, you will produce high-quality, efficient, and reproducible solutions to data analysis problems.

## Few-Shot Example

**Question:** Perform a distribution analysis on the 'Fare' column for each passenger class ('Pclass') separately. Calculate the mean, median, and standard deviation of the fare for each class.

**Answer:** @mean_fare_class1[87.96] @median_fare_class1[69.30] @std_dev_fare_class1[80.86] @mean_fare_class2[21.47] @median_fare_class2[15.05] @std_dev_fare_class2[13.19] @mean_fare_class3[13.23] @median_fare_class3[13.23] @std_dev_fare_class3[10.04]

**Data File:** test_ave.csv

**Difficulty:** medium

\end{lstlisting}
\end{tcolorbox}
\vspace{-1em}
\captionof{figure}{PromptBreeder Prompt for DABench}
\label{fig:promptbreeder-dabench}

\begin{tcolorbox}[breakable, enhanced, base, title={TextGrad Prompt for DABench}]
    \begin{lstlisting}
You are a seasoned data analyst with extensive experience in transforming complex datasets into actionable insights. You possess a deep understanding of statistical analysis, data visualization, and machine learning techniques. You are dedicated to delivering high-quality reports and are committed to continuous learning and improvement.
\end{lstlisting}
\end{tcolorbox}
\vspace{-1em}
\captionof{figure}{TextGrad Prompt for DABench}
\label{fig:textgrad-dabench}

\section{Additional Details of Calculation of Kendall's tau-b}
\label{sec:calc-kendall}

Formally, let $\boldsymbol{\hat{\tau}} = [\hat{\tau}_1, \hat{\tau}_2, \ldots, \hat{\tau}_n]$ denote the estimated causal effects for $n$ samples, and $\boldsymbol{\tau} = [\tau_1, \tau_2, \ldots, \tau_n]$ the corresponding ground-truth effects. To compute Kendall’s tau-b, we consider all possible unordered pairs of samples, indexed by $(i, j)$ with $i < j$. For each pair, if the ordering of the model estimates matches the ordering of the ground truth, that is, $(\hat{\tau}_i - \hat{\tau}_j)(\tau_i - \tau_j) > 0$, the pair is considered \emph{concordant}. If the orderings disagree, meaning $(\hat{\tau}_i - \hat{\tau}_j)(\tau_i - \tau_j) < 0$, the pair is classified as \emph{discordant}. If either $\hat{\tau}_i = \hat{\tau}_j$ or $\tau_i = \tau_j$, the pair is considered as \emph{tied}. Let $N_c$ be the number of concordant pairs, $N_d$ the number of discordant pairs, $T_{\hat{\tau}}$ the number of pairs tied only on the predictions (i.e., $\hat{\tau}_i = \hat{\tau}_j$, but $\tau_i \neq \tau_j$), and $T_{\tau}$ the number of pairs tied only on the ground truth (i.e., $\tau_i = \tau_j$, but $\hat{\tau}_i \neq \hat{\tau}_j$). Kendall’s tau-b is then defined as:
\begin{equation}
\tau_b = \frac{N_c - N_d}{\sqrt{(N_c + N_d + T_{\hat{\tau}})\,(N_c + N_d + T_{\tau})}}
\end{equation}
where the denominator corrects for the presence of tied pairs to ensure a fair ranking comparison even when ties are frequent. The value of $\tau_b$ ranges from $-1$ (perfectly discordant) to $+1$ (perfectly concordant), with higher values indicating stronger agreement between the estimated and true causal rankings.

In implementation, for each query in the validation set, we compute Kendall’s tau-b between the sequence of predicted prompt effects $\boldsymbol{\hat{\tau}}$ and the corresponding true performance outcomes $\boldsymbol{\tau}$ across all prompts. The final score is obtained by averaging the tau-b values across all queries, providing a stable and interpretable aggregate measure of ranking quality.

\section{Additional Details of Causal-Guided Optimization on DABench}
\label{sec:example}

This section outlines additional experimental details of \textit{Stage 2: Causal-Guided Optimization} in our CPO optimization framework on DABench.

\subsection{Initial Prompt Selection}
\label{bps}
In the first step, we constructed the initial set of prompts used for optimization, which were directly drawn from the previously curated prompt pool and served as the starting point for optimization. An initial prompt may contain the following components (not all were required):
\begin{itemize}
    \item \emph{Instruction}: the role and general procedure the model should follow;
    \item \emph{Question}: core query, corresponds to $x$;
    \item \emph{File path}: pointers to data files referenced by the query;
    \item \emph{Format requirements}: output structure or markup the model must follow;
    \item \emph{Treatment information}: optional ``dimensions'' such as constraints, dataset descriptions, or illustrative few-shot examples.
\end{itemize}

In our causal view, the optional components (constraints, dataset descriptions, few-shot examples) 
constituted distinct treatment dimensions embedded in prompt $t$. Thus, not every prompt contains all treatment dimensions, and the absence or presence of such components constitutes meaningful variation in $t$.


\subsection{Optimization Process}
\label{op}

The optimization stage was designed to iteratively refine the prompts obtained in Section~\ref{bps}. This stage consisted of several key phases: (i) placeholder substitution, (ii) variant prompts generation, 
(iii) reconstruction of complete prompts, (iv) embedding transformation, (v) causal effects estimation and (vi) ranking for selection. 

\textbf{Placeholder substitution}. To isolate the role of instructions in shaping model behavior, each prompt was converted into a normalized template in which all query-dependent elements—including the question text, file path, output-format specification, constraints, and JSON metadata—were replaced with symbolic placeholders. Only the general instructional component remained in natural language. This step ensured that the optimization process focused exclusively on refining the prompt $t$ while keeping the query-specific information intact.

\textbf{Variant prompts generation}. The placeholder-included prompts were then given as examples to the $LLM_{prompt}$ to produce new prompt variants for the same query $x$, where the instructions differed but the placeholders for the question, file path and format requirements remained unchanged. Auxiliary components such as constraints or dataset descriptions could be optionally retained or omitted, allowing controlled variation across treatment dimensions.

It is important to note that when an initial prompt contained few-shot examples, these examples were not replaced with placeholders. Therefore, the model occasionally rewrote or reformatted such components during variant generation, introducing additional heterogeneity in the resulting prompt pool. This reflects realistic variability in the prompt $t$, which is desirable for causal effect estimation.

\textbf{Reconstruction of complete prompts}. Once the new instruction variants were produced, the placeholders were substituted back with the actual content corresponding to each specific query. This reconstruction step yielded complete candidate prompts that combine the original problem-specific information with alternative instruction formulations. 

\textbf{Embedding transformation}. Once the reconstructed prompts were obtained, each candidate prompt $t$ was mapped into an embedding space using the sentence encoder and then projected by query-specific and prompt-specific PCA operators $\psi_X$ and $\psi_T$.

\textbf{Causal effects estimation}. To define the control group, we adopted the same control condition as in the offline training phase without any detailed instructions.

\textbf{Ranking for selection}. For each query, the new candidate prompts were available at this stage, and they will be ranked by descending order by estimated treatment effects. The top three prompts for each query were selected as the seed set for the next round.


\subsection{Evaluation Stage}
\label{es}

To rigorously assess the effectiveness of the optimized prompts obtained from Section~\ref{op}, we constructed a fully automated evaluation pipeline. The pipeline was implemented within the \texttt{smolagents} framework ~\citep{smolagents}, which integrated the programming-oriented $LLM_{task}$ with automated code execution and verification capabilities. 

The evaluation workflow proceeds as follows. Upon receiving a fully reconstructed prompt $t$, the model (i) generated executable code according to the instructions contained in $t$, (ii) executed that code to load and process the relevant dataset, (iii) produced an analysis output and (iv) formatted the output according to the requirements specified in the prompt. Given the formatted output, the agent would compare the generated output against the ground-truth label $l$. The scoring function $\mathcal{E}\left(LLM_{\text{task}}(x, t), l\right)$ then determines correctness, producing the observed outcome $y$ for each prompt. By combining automated execution with standardized scoring, this evaluation protocol provided consistent and reproducible measures of prompt performance.


%% file: references-short.bib
@STRING{PNAS        = "Proc. Natl. Acad. Sci."}


%% file: references.bib
@article{leng2024calibration,
  title={Calibration of heterogeneous treatment effects in randomized experiments},
  author={Leng, Yan and Dimmery, Drew},
  journal={Information Systems Research},
  volume={35},
  number={4},
  pages={1721--1742},
  year={2024},
  publisher={INFORMS}
}

@article{zhou2024generative,
  title={Generative artificial intelligence, human creativity, and art},
  author={Zhou, Eric and Lee, Dokyun},
  journal={PNAS nexus},
  volume={3},
  number={3},
  pages={pgae052},
  year={2024},
  publisher={Oxford University Press US}
}

@article{chen2024large,
  title={Large Language Model in Creative Work: The Role of Collaboration Modality and User Expertise},
  author={Chen, Zenan and Chan, Jason},
  journal={Management Science},
  volume={70},
  number={12},
  year={2024},
  publisher={INFORMS}
}

@article{rai2024pathways,
  title={Pathways for Design Research on Artificial Intelligence},
  author={Rai, Arun},
  journal={Information Systems Research},
  volume={35},
  number={2},
  year={2024},
  publisher={INFORMS}
}

@article{susarla2025inventing,
  title={Inventing with Machines: Generative {AI} and the Evolving Landscape of IS Research},
  author={Susarla, Anjana and others},
  journal={Information Systems Research},
  volume={36},
  number={4},
  year={2025},
  publisher={INFORMS}
}

@article{yoo2024next,
  title={The Next Frontiers of Digital Innovation Research},
  author={Yoo, Youngjin and others},
  journal={Information Systems Research},
  volume={35},
  number={4},
  year={2024},
  publisher={INFORMS}
}

@article{eloundou2024gpts,
  title={{GPTs} are {GPTs}: Labor market impact potential of {LLMs}},
  author={Eloundou, Tyna and Manning, Sam and Mishkin, Pamela and Rock, Daniel},
  journal={Science},
  volume={384},
  number={6702},
  pages={1306--1308},
  year={2024},
  publisher={American Association for the Advancement of Science}
}

@article{brynjolfsson2025generative,
  title={Generative {AI} at work},
  author={Brynjolfsson, Erik and Li, Danielle and Raymond, Lindsey},
  journal={The Quarterly Journal of Economics},
  volume={140},
  number={2},
  pages={889--942},
  year={2025},
  publisher={Oxford University Press}
}

@article{wager2018estimation,
  title={Estimation and inference of heterogeneous treatment effects using random forests},
  author={Wager, Stefan and Athey, Susan},
  journal={Journal of the American Statistical Association},
  volume={113},
  number={523},
  pages={1228--1242},
  year={2018},
  publisher={Taylor \& Francis}
}

@article{shi2025what,
  title={What, Why, and How: An Empiricist's Guide to Double/Debiased Machine Learning},
  author={Shi, Bowen and Mao, Xiaojie and Yang, Mochen and Li, Bo},
  journal={Information Systems Research},
  year={2025}
}

@article{athey2019generalized,
  title={Generalized random forests},
  author={Athey, Susan and Tibshirani, Julie and Wager, Stefan},
  journal={The Annals of Statistics},
  volume={47},
  number={2},
  pages={1148--1178},
  year={2019},
  publisher={JSTOR}
}

@article{brown2020language,
  title={Language models are few-shot learners},
  author={Brown, Tom and Mann, Benjamin and Ryder, Nick and Subbiah, Melanie and Kaplan, Jared D and Dhariwal, Prafulla and Neelakantan, Arvind and Shyam, Pranav and Sastry, Girish and Askell, Amanda and others},
  journal={Advances in Neural Information Processing Systems},
  volume={33},
  pages={1877--1901},
  year={2020}
}

@article{wei2022chain,
  title={Chain-of-thought prompting elicits reasoning in large language models},
  author={Wei, Jason and Wang, Xuezhi and Schuurmans, Dale and Bosma, Maarten and Xia, Fei and Chi, Ed and Le, Quoc V and Zhou, Denny and others},
  journal={Advances in Neural Information Processing Systems},
  volume={35},
  pages={24824--24837},
  year={2022}
}

@inproceedings{kong2024better,
  title={Better Zero-Shot Reasoning with Role-Play Prompting},
  author={Kong, Aobo and Zhao, Shiwan and Chen, Hao and Li, Qicheng and Qin, Yong and Sun, Ruiqi and Zhou, Xin and Wang, Enzhi and Dong, Xiaohang},
  booktitle={Proceedings of the 2024 Conference of the North American Chapter of the Association for Computational Linguistics: Human Language Technologies (Volume 1: Long Papers)},
  pages={4099--4113},
  year={2024}
}

@inproceedings{zhou2022large,
  title={Large language models are human-level prompt engineers},
  author={Zhou, Yongchao and Muresanu, Andrei Ioan and Han, Ziwen and Paster, Keiran and Pitis, Silviu and Chan, Harris and Ba, Jimmy},
  booktitle={The Eleventh International Conference on Learning Representations},
  year={2022}
}

@inproceedings{reynolds2021prompt,
  title={Prompt programming for large language models: Beyond the few-shot paradigm},
  author={Reynolds, Laria and McDonell, Kyle},
  booktitle={Extended Abstracts of the 2021 CHI Conference on Human Factors in Computing Systems},
  pages={1--7},
  year={2021}
}

@article{sahoo2024systematic,
  title={A systematic survey of prompt engineering in large language models: Techniques and applications},
  author={Sahoo, Pranab and Singh, Ayush Kumar and Saha, Sriparna and Jain, Vinija and Mondal, Samrat and Chadha, Aman},
  journal={arXiv preprint arXiv:2402.07927},
  year={2024}
}

@inproceedings{deng2022rlprompt,
  title={{RLPrompt}: Optimizing Discrete Text Prompts with Reinforcement Learning},
  author={Deng, Mingkai and Wang, Jianyu and Hsieh, Cheng-Ping and Wang, Yihan and Guo, Han and Shu, Tianmin and Song, Meng and Xing, Eric and Hu, Zhiting},
  booktitle={Proceedings of the 2022 Conference on Empirical Methods in Natural Language Processing},
  pages={3369--3391},
  year={2022}
}

@inproceedings{wangpromptagent,
  title={{PromptAgent}: Strategic Planning with Language Models Enables Expert-level Prompt Optimization},
  author={Wang, Xinyuan and Li, Chenxi and Wang, Zhen and Bai, Fan and Luo, Haotian and Zhang, Jiayou and Jojic, Nebojsa and Xing, Eric and Hu, Zhiting},
  booktitle={The Twelfth International Conference on Learning Representations},
  year={2024}
}

@inproceedings{cheng2024black,
  title={Black-Box Prompt Optimization: Aligning Large Language Models without Model Training},
  author={Cheng, Jiale and Liu, Xiao and Zheng, Kehan and Ke, Pei and Wang, Hongning and Dong, Yuxiao and Tang, Jie and Huang, Minlie},
  booktitle={Proceedings of the 62nd Annual Meeting of the Association for Computational Linguistics (Volume 1: Long Papers)},
  pages={3201--3219},
  year={2024}
}

@inproceedings{pryzant2023automatic,
  title={Automatic Prompt Optimization with “Gradient Descent” and Beam Search},
  author={Pryzant, Reid and Iter, Dan and Li, Jerry and Lee, Yin and Zhu, Chenguang and Zeng, Michael},
  booktitle={Proceedings of the 2023 Conference on Empirical Methods in Natural Language Processing},
  pages={7957--7968},
  year={2023}
}

@article{khattab2023dspy,
  title={{DSPy}: Compiling declarative language model calls into self-improving pipelines},
  author={Khattab, Omar and Singhvi, Arnav and Maheshwari, Paridhi and Zhang, Zhiyuan and Santhanam, Keshav and Vardhamanan, Sri and Haq, Saiful and Sharma, Ashutosh and Joshi, Thomas T and Moazam, Hanna and others},
  journal={arXiv preprint arXiv:2310.03714},
  year={2023}
}

@article{radford2019language,
  title={Language models are unsupervised multitask learners},
  author={Radford, Alec and Wu, Jeffrey and Child, Rewon and Luan, David and Amodei, Dario and Sutskever, Ilya and others},
  journal={OpenAI blog},
  volume={1},
  number={8},
  pages={9},
  year={2019}
}

@article{zhao2024revolutionizing,
  title={Revolutionizing Finance with {LLMs}: An Overview of Applications and Insights},
  author={Zhao, Huaqin and Liu, Zhengliang and Wu, Zihao and Li, Yiwei and Yang, Tianze and Shu, Peng and Xu, Shaochen and Dai, Haixing and Zhao, Lin and Mai, Gengchen and others},
  journal={CoRR},
  year={2024}
}

@article{liu2023pre,
  title={Pre-train, prompt, and predict: A systematic survey of prompting methods in natural language processing},
  author={Liu, Pengfei and Yuan, Weizhe and Fu, Jinlan and Jiang, Zhengbao and Hayashi, Hiroaki and Neubig, Graham},
  journal={ACM computing surveys},
  volume={55},
  number={9},
  pages={1--35},
  year={2023},
  publisher={ACM New York, NY}
}

@inproceedings{li2021prefix,
  title={Prefix-Tuning: Optimizing Continuous Prompts for Generation},
  author={Li, Xiang Lisa and Liang, Percy},
  booktitle={Proceedings of the 59th Annual Meeting of the Association for Computational Linguistics and the 11th International Joint Conference on Natural Language Processing (Volume 1: Long Papers)},
  pages={4582--4597},
  year={2021}
}

@inproceedings{yang2023large,
  title={Large language models as optimizers},
  author={Yang, Chengrun and Wang, Xuezhi and Lu, Yifeng and Liu, Hanxiao and Le, Quoc V and Zhou, Denny and Chen, Xinyun},
  booktitle={The Twelfth International Conference on Learning Representations},
  year={2023}
}

@inproceedings{fernando2024promptbreeder,
  title={Promptbreeder: Self-referential self-improvement via prompt evolution},
  author={Fernando, Chrisantha and Banarse, Dylan and Michalewski, Henryk and Osindero, Simon and Rockt{\"a}schel, Tim},
  booktitle={Proceedings of the 41st International Conference on Machine Learning},
  pages={13481--13544},
  year={2024}
}

@inproceedings{kong2025query,
  title={Query-dependent Prompt Optimization via Multi-Loop Offline Reinforcement Learning},
  author={Kong, Yilun and Mao, Hangyu and Zhao, Qi and Zhang, Bin and Ruan, Jingqing and Shen, Li and Chang, Yongzhe and Wang, Xueqian and Zhao, Rui and Tao, Dacheng},
  booktitle={ICLR 2025 Workshop on Navigating and Addressing Data Problems for Foundation Models},
  year={2025}
}

@inproceedings{sun2024querydependent,
title={Query-Dependent Prompt Evaluation and Optimization with Offline Inverse {RL}},
author={Hao Sun and Alihan H{\"u}y{\"u}k and Mihaela van der Schaar},
booktitle={The Twelfth International Conference on Learning Representations},
year={2024},
url={https://openreview.net/forum?id=N6o0ZtPzTg}
}

@article{nica2025trprompt,
  title={{TRPrompt}: Bootstrapping Query-Aware Prompt Optimization from Textual Rewards},
  author={Nica, Andreea and Zakazov, Ivan and Baldwin, Nicolas Mario and Geng, Saibo and West, Robert},
  journal={arXiv preprint arXiv:2507.18618},
  year={2025}
}

@inproceedings{hendrycks2021measuring,
  title={Measuring Mathematical Problem Solving With the MATH Dataset},
  author={Hendrycks, Dan and Burns, Collin and Kadavath, Saurav and Arora, Akul and Basart, Steven and Tang, Eric and Song, Dawn and Steinhardt, Jacob},
  booktitle={Thirty-fifth Conference on Neural Information Processing Systems Datasets and Benchmarks Track (Round 2)},
  year={2021}
}

@article{chen2024viseval,
  title={Viseval: A benchmark for data visualization in the era of large language models},
  author={Chen, Nan and Zhang, Yuge and Xu, Jiahang and Ren, Kan and Yang, Yuqing},
  journal={IEEE Transactions on Visualization and Computer Graphics},
  year={2024},
  publisher={IEEE}
}

@inproceedings{hu2024infiagent,
  title={InfiAgent-DABench: Evaluating agents on data analysis tasks},
  author={Hu, Xueyu and Zhao, Ziyu and Wei, Shuang and Chai, Ziwei and Ma, Qianli and Wang, Guoyin and Wang, Xuwu and Su, Jing and Xu, Jingjing and Zhu, Ming and others},
  booktitle={Proceedings of the 41st International Conference on Machine Learning},
  pages={19544--19572},
  year={2024}
}

@article{shinn2023reflexion,
  title={Reflexion: Language agents with verbal reinforcement learning},
  author={Shinn, Noah and Cassano, Federico and Gopinath, Ashwin and Narasimhan, Karthik and Yao, Shunyu},
  journal={Advances in Neural Information Processing Systems},
  volume={36},
  pages={8634--8652},
  year={2023}
}

@article{yuksekgonul2025optimizing,
  title={Optimizing generative {AI} by backpropagating language model feedback},
  author={Yuksekgonul, Mert and Bianchi, Federico and Boen, Joseph and Liu, Sheng and Lu, Pan and Huang, Zhi and Guestrin, Carlos and Zou, James},
  journal={Nature},
  volume={639},
  number={8055},
  pages={609--616},
  year={2025},
  publisher={Nature Publishing Group}
}

@inproceedings{zhang2024mlcopilot,
  title={MLCopilot: Unleashing the Power of Large Language Models in Solving Machine Learning Tasks},
  author={Zhang, Lei and Zhang, Yuge and Ren, Kan and Li, Dongsheng and Yang, Yuqing},
  booktitle={Proceedings of the 18th Conference of the European Chapter of the Association for Computational Linguistics (Volume 1: Long Papers)},
  pages={2931--2959},
  year={2024}
}

@Misc{smolagents,
  title =        {`smolagents': a smol library to build great agentic systems.},
  author =       {Aymeric Roucher and Albert Villanova del Moral and Thomas Wolf and Leandro von Werra and Erik Kaunismäki},
  howpublished = {\url{https://github.com/huggingface/smolagents}},
  year =         {2025}
}

@article{ramnath2025systematic,
  title={A systematic survey of automatic prompt optimization techniques},
  author={Ramnath, Kiran and Zhou, Kang and Guan, Sheng and Mishra, Soumya Smruti and Qi, Xuan and Shen, Zhengyuan and Wang, Shuai and Woo, Sangmin and Jeoung, Sullam and Wang, Yawei and others},
  journal={arXiv preprint arXiv:2502.16923},
  year={2025}
}

@inproceedings{wangself,
  title={Self-consistency improves chain of thought reasoning in language models},
  author={Wang, Xuezhi and Wei, Jason and Schuurmans, Dale and Le, Quoc and Chi, Ed and Narang, Sharan and Chowdhery, Aakanksha and Zhou, Denny},
  booktitle={The Eleventh International Conference on Learning Representations},
  year={2023}
}

@article{yao2023tree,
  title={Tree of thoughts: Deliberate problem solving with large language models},
  author={Yao, Shunyu and Yu, Dian and Zhao, Jeffrey and Shafran, Izhak and Griffiths, Tom and Cao, Yuan and Narasimhan, Karthik},
  journal={Advances in neural information processing systems},
  volume={36},
  pages={11809--11822},
  year={2023}
}

@article{lewis2020retrieval,
  title={Retrieval-augmented generation for knowledge-intensive nlp tasks},
  author={Lewis, Patrick and Perez, Ethan and Piktus, Aleksandra and Petroni, Fabio and Karpukhin, Vladimir and Goyal, Naman and K{\"u}ttler, Heinrich and Lewis, Mike and Yih, Wen-tau and Rockt{\"a}schel, Tim and others},
  journal={Advances in neural information processing systems},
  volume={33},
  pages={9459--9474},
  year={2020}
}

@article{xu2025causal,
  title={A Causal Approach to Representation Learning for Unstructured Data},
  author={Xu, Sikun and Jiang, Zhenling and Qi, Zhengling and Zhang, Dennis},
  journal={Available at SSRN},
  year={2025}
}

@inproceedings{shi2021invariant,
  title={Invariant representation learning for treatment effect estimation},
  author={Shi, Claudia and Veitch, Victor and Blei, David M},
  booktitle={Uncertainty in artificial intelligence},
  pages={1546--1555},
  year={2021},
  organization={PMLR}
}

@article{ellickson2023estimating,
  title={Estimating marketing component effects: Double machine learning from targeted digital promotions},
  author={Ellickson, Paul B and Kar, Wreetabrata and Reeder III, James C},
  journal={Marketing Science},
  volume={42},
  number={4},
  pages={704--728},
  year={2023},
  publisher={INFORMS}
}

@article{ellickson2024using,
  title={Using contextual embeddings to predict the effectiveness of novel heterogeneous treatments},
  author={Ellickson, Paul B and Kar, Wreetabrata and Reeder III, James C and Zeng, Guang},
  journal={Available at SSRN 4845956},
  year={2024}
}

@article{hunermund2022causal,
  title={Causal machine learning and business decision making},
  author={H{\"u}nermund, Paul and Kaminski, Jermain and Schmitt, Carla},
  journal={Available at SSRN 3867326},
  year={2022}
}

@inproceedings{more2023double,
  title={Double machine learning at scale to predict causal impact of customer actions},
  author={More, Sushant and Kotwal, Priya and Chappidi, Sujith and Mandalapu, Dinesh and Khawand, Chris},
  booktitle={Joint European Conference on Machine Learning and Knowledge Discovery in Databases},
  pages={513--528},
  year={2023},
  organization={Springer}
}

@incollection{geroimenko2025key,
  title={Key Principles of Good Prompt Design},
  author={Geroimenko, Vladimir},
  booktitle={The Essential Guide to Prompt Engineering: Key Principles, Techniques, Challenges, and Security Risks},
  pages={17--36},
  year={2025},
  publisher={Springer}
}

@inproceedings{white2023prompt,
  title={A Prompt Pattern Catalog to Enhance Prompt Engineering with ChatGPT},
  author={White, Jules and Fu, Quchen and Hays, Sam and Sandborn, Michael and Olea, Carlos and Gilbert, Henry and Elnashar, Ashraf and Spencer-Smith, Jesse and Schmidt, Douglas C},
  booktitle={Proceedings of the 30th Conference on Pattern Languages of Programs},
  pages={1--31},
  year={2023}
}

@article{chernozhukov2018double,
  title={Double/debiased machine learning for treatment and structural parameters},
  author={Chernozhukov, Victor and Chetverikov, Denis and Demirer, Mert and Duflo, Esther and Hansen, Christian and Newey, Whitney and Robins, James},
  journal={The Econometrics Journal},
  pages={C1--C68},
  year={2018},
  publisher={JSTOR}
}

@book{agresti2010analysis,
  title={Analysis of ordinal categorical data},
  author={Agresti, Alan},
  year={2010},
  publisher={John Wiley \& Sons}
}

@article{nussbaum2025nomic,
  title={Nomic Embed: Training a Reproducible Long Context Text Embedder},
  author={Nussbaum, Zach and Morris, John Xavier and Mulyar, Andriy and Duderstadt, Brandon},
  journal={Transactions on Machine Learning Research},
  year={2025}
}

@book{imbens2015causal,
  title={Causal inference in statistics, social, and biomedical sciences},
  author={Imbens, Guido W and Rubin, Donald B},
  year={2015},
  publisher={Cambridge university press}
}

@article{kojima2022large,
  title={Large language models are zero-shot reasoners},
  author={Kojima, Takeshi and Gu, Shixiang Shane and Reid, Machel and Matsuo, Yutaka and Iwasawa, Yusuke},
  journal={Advances in neural information processing systems},
  volume={35},
  pages={22199--22213},
  year={2022}
}

@inproceedings{cui2024phaseevo,
  title={PhaseEvo: Towards Unified Long-Context Prompt Optimization for Large Language Models},
  author={Cui, Wendi and Zhang, Jiaxin and Li, Zhuohang and Sun, Hao and Lopez, Damien and Das, Kamalika and Malin, Bradley A and Kumar, Sricharan},
  booktitle={First Workshop on Long-Context Foundation Models@ ICML 2024},
  year={2024}
}

@article{wu2024prompt,
  title={Prompt optimization with EASE? Efficient ordering-aware automated selection of exemplars},
  author={Wu, Zhaoxuan and Lin, Xiaoqiang and Dai, Zhongxiang and Hu, Wenyang and Shu, Yao and Ng, See-Kiong and Jaillet, Patrick and Low, Bryan Kian Hsiang},
  journal={Advances in Neural Information Processing Systems},
  volume={37},
  pages={122706--122740},
  year={2024}
}
